\documentclass[10pt,twocolumn,letterpaper]{article}
\pdfoutput=1

\usepackage{subcaption}

\usepackage{arxiv}
\usepackage{times}
\usepackage{epsfig}
\usepackage{graphicx}
\usepackage{amsmath}
\usepackage{amssymb}

\usepackage{grffile}
\usepackage[group-separator={,}]{siunitx}

\begin{document}

\title{High-Quality Face Capture Using Anatomical Muscles}

\author{
Michael Bao\textsuperscript{1,2} \qquad Matthew Cong\textsuperscript{2,\dag} \qquad St\'ephane Grabli\textsuperscript{2,\dag} \qquad Ronald Fedkiw\textsuperscript{1,2} \\
\textsuperscript{1}Stanford University \qquad \textsuperscript{2}Industrial Light \& Magic \\
{\tt\small \textsuperscript{1}\{mikebao,rfedkiw\}@stanford.edu} \qquad {\tt\small \textsuperscript{\dag}\{mcong,sgrabli\}@ilm.com}
}

\maketitle

\begin{abstract}
    Muscle-based systems have the potential to provide both anatomical accuracy and semantic interpretability as compared to blendshape models; however, a lack of expressivity and differentiability has limited their impact.
    Thus, we propose modifying a recently developed rather expressive muscle-based system in order to make it fully-differentiable; in fact, our proposed modifications allow this physically robust and anatomically accurate muscle model to conveniently be driven by an underlying blendshape basis.
    Our formulation is intuitive, natural, as well as monolithically and fully coupled such that one can differentiate the model from end to end, which makes it viable for both optimization and learning-based approaches for a variety of applications.
    We illustrate this with a number of examples including both shape matching of three-dimensional geometry as as well as the automatic determination of a three-dimensional facial pose from a single two-dimensional RGB image without using markers or depth information.
\end{abstract}

\section{Introduction} \label{sec:intro}
Muscle simulation-based animation systems are attractive due to their ability to preserve important physical properties such as volume conservation as well as their ability to handle contact and collision.
Moreover, utilizing an anatomically motivated set of controls provides a straightforward way of extracting out semantic meaning from the control values.
Unfortunately, even though \cite{sifakis2005automatic} was able to automatically compute muscle activation values given sparse motion capture data, muscle-based animation models have proven to be significantly less expressive and harder to control than their blendshape-based counterparts \cite{lewis2014practice}.

Recently, \cite{cong2016art} introduced a novel method that significantly improved upon the expressiveness of muscle-based animation systems.
They introduced the concept of ``muscle tracks'' to control the deformation of the underlying musculature.
This concept gives the muscle simulation enough expressiveness to target arbitrary shapes, which allowed it be used in high-quality movie productions such as \textit{Kong: Skull Island} where it was used both to aid in the creation of blendshapes and to offer physically-based corrections to artist-created animation sequences \cite{cong2017muscle,lan2017lessons}.
While \cite{cong2016art} alleviates the problems of muscle-based simulation in regards to expressiveness and control, the method is geared towards generative computer graphics problems, and is thus not amenable for estimating a facial pose from a two-dimensional image as is common for markerless performance capture.
One could iterate between solving for a performance using blendshapes and then using a muscle-based solution to correct the blendshapes; however, this iterative method is lossy as the muscle simulation does not have access to the raw data and may thus hallucinate details or erase details of the performance.

In this paper, we extend \cite{cong2016art} by combining the ease of use and differentiability of traditional blendshape models with expressive, physically-plausible muscle track simulations in order to create a differentiable simulation framework that can be used interchangeably with traditional blendshape models for facial performance capture and animation.
Instead of relying on a non-differentiable per-frame volumetric morph to drive the muscle track deformation as in \cite{cong2016art}, we instead create a state-of-the-art blendshape model for each muscle, which is then used to drive its volumetric deformation.
Our model maintains the expressiveness of \cite{cong2016art} while preserving crucial physical properties.
Furthermore, our new formulation is differentiable from end to end, which allows it to be used to target three-dimensional facial poses as well as two-dimensional RGB images.
We demonstrate that our blendshape muscle tracks method shows significant improvements in anatomical plausibility and semantic interpretability when compared to state-of-the-art blendshape-based methods for targeting three-dimensional geometry and two-dimensional RGB images.

\section{Related Work} \label{sec:related_work}

\textbf{Face Models:} 
Although our work does not directly address the modeling part of the pipeline, it relies on having a pre-existing model of the face.
For building a realistic digital double of an actor, multi-view stereo techniques can be used to collect high-quality geometry and texture information in a variety of poses \cite{beeler2010high,beeler2011high,debevec2012light}.
Artists can then use this data to create the final blendshape model.
In state-of-the-art models, the deformation model will include non-linear skinning/enveloping in addition to linear blendshapes to achieve more plausible deformations \cite{lewis2014practice}.
On the other hand, more generalized digital face models would be more useful in cases where the target actor is not known beforehand.
One would generally use a 3D morphable model (3DMM) which can be created using statistical methods from a large database of scanned faces.
Such models include the classic Blanz and Vetter model \cite{blanz1999morphable}, the Basel Face Model (BFM) \cite{luthi2017gaussian,paysan20093d}, FaceWarehouse \cite{cao2014facewarehouse}, and the Large Scale Facial Model (LSFM) \cite{booth20163d}.
Recent models such as the FLAME model \cite{li2017learning} have begun to introduce non-linear deformations by using skinning and corrective blendshapes.
These models tend to be geared towards real-time applications and as a result have a low number of vertices.

\textbf{Face Capture:}
A more comprehensive review of facial performance capture techniques can be found in \cite{zollhofer2018state}.
To date, marker based techniques have been the most popular for capturing facial performances for both real-time applications and feature films.
Helmet mounted cameras (HMCs) are often used to stereo track a sparse set of markers on the face.
These markers are then used as constraints in an optimization to find blendshape weights \cite{bhat2013high}.
In many real-time applications, pre-applied markers are generally not an option so 2D features \cite{cao20133d,chen2013accurate,wu2016anatomically}, depth images \cite{chen2013accurate,kazemi2014real,weise2011realtime}, or low-resolution RGB images \cite{thies2016face2face} are often used instead.
Other methods have focused on using traditional computer vision techniques to track a facial performance with consistent topology \cite{beeler2010high,beeler2011high,fyffe2017multi}.
More recently, methods using neutral networks have been used to reconstruct face geometry \cite{jackson2017large,sela2017unrestricted} and estimate facial control parameters \cite{jourabloo2016large,kim2018inversefacenet}.
Analysis-by-synthesis techniques have also been explored for capturing facial performances \cite{pighin1999resynthesizing}.

\textbf{Face Simulation:}
\cite{sifakis2005automatic} was one of the first to utilize quasistatic simulations to drive the deformation of a 3D face, especially for motion capture.
There has also been interest in using quasistative simulations to drive muscle deformations in the body \cite{irving2004invertible,teran2003finite,teran2005creating}.
However, in general, facial muscle simulations tend to be less expressive than their artist-driven blendshape counterparts.
More recently, significant work has been done to make muscle simulations more expressive \cite{cong2016art,ichim2017phace}.
While these methods can be used to target data in the form of geometry, it is unclear how to cleanly transfer these methods to target non-geometry data such as two-dimensional RGB images.
Other work has been done to try to introduce physical simulations into the blendshape models themselves \cite{barrielle2018realtime,barrielle2016blendforces,ichim2016building,kozlov2017enriching}; however, these works do not focus on the inverse problem.

\section{Blendshape Model} \label{sec:blendshapes}

As discussed in Section \ref{sec:related_work}, there are many different types of blendshape models that exist and we refer interested readers to \cite{lewis2014practice} for a more thorough overview of existing literature.
We focus on the state-of-the-art hybrid blendshape deformation model that is the basis of our method introduced in Section \ref{sec:blendshape_muscle_tracks}.
A hybrid blendshape model refers to a deformation model that uses both linear blendshapes and linear blend skinning to deform the vertices of the mesh.
Our model contains a single 6-DOF joint for the jaw.
We can succinctly write the model given the blendshape parameters $b$ and joint parameters $j$ as
\begin{align}
x(b, j) & = T(j) (n + Bb) \label{eq:blendshapes}
\end{align}
where $n$ is the neutral shape, $B$ is the blendshape deltas matrix, and $T(j)$ contains the linear blend skinning matrix, \ie a transformation matrix due to a change in the jaw joint, for each vertex.
Note that $n + Bb$ is often referred to as the \textit{pre-skinning} shape and $Bb$ as the \textit{pre-skinning} displacements.
More complex animation systems include corrective shapes and intermediate controls and thus we let $w$ denote a broader set of animator controls which we treat as our independent variable rewriting Equation \ref{eq:blendshapes} as
\begin{align}
x(w) & = T(j(w)) (n + Bb(w)) \label{eq:full_blendshapes}
\end{align}
where $j(w)$ and $b(w)$ may include non-linearities such as non-linear corrective blendshapes.

\section{Muscle Model} \label{sec:simulation}

We create an anatomical model of the face consisting of the cranium, jaw, and a tetrahedralized flesh mesh with embedded muscles for a given actor using the method of \cite{cong2015fully}.
Since we desire parity with the facial model used to deform the face surface, we define the jaw joint as a 6-DOF joint equivalent to the one used to skin the face surface in Section \ref{sec:blendshapes}.
Traditionally, face simulation models have been controlled using a vector of muscle activation parameters which we denote as $a$.
We use the same constitutive model for the muscle as \cite{teran2003finite,teran2005creating} which consists of an isotropic Mooney-Rivlin term, a quasi-incompressibility term, and an anisotropic passive/active muscle response term.
The finite-volume method \cite{teran2003finite,teran2005robust} is used to compute the force on each vertex of the tetrahedralized flesh mesh given the current 1st Piola-Kirchoff stress computed using the constitutive model and the current deformation gradient.
Some vertices of the flesh mesh $X^C$ are constrained to kinematically follow along with the cranium/jaw and the steady state position is implicitly defined as the positions of the unconstrained flesh mesh vertices $X^U$ which make the sum of all relevant forces identically $0$, \ie $f(X^C, X^U) = 0$.

One can decompose the forces to be a sum of the finite-volume forces and collision penalty forces
\begin{align}
    f_{\text{fvm}}(X^C, X^U, a) + f_{\text{collisions}}(X^C, X^U) & = 0 \label{eq:full_quasistatics}.
\end{align}
One can further break down the finite-volume forces into the passive force $f_p$ and active force $f_a$.
Then using the fact the the active muscle response is scaled linearly by the muscle activation $a$ \cite{zajac1989muscle}, we can rewrite the finite-volume force as
\begin{align}
    f_{\text{fvm}}(X^C, X^U, a) & = f_{\text{p}}(X^C, X^U) + af_{\text{a}}(X^C, X^U) \label{eq:muscle_forces}.
\end{align}
We refer interested readers to \cite{sifakis2005automatic,teran2003finite,teran2005creating,teran2005robust} for derivations of the aforementioned forces and their associated Jacobians with respect to the flesh mesh vertices.
Given a vector of muscle activations and cranium/jaw parameters, Equation \ref{eq:full_quasistatics} can be solved using the Newton-Raphson method to compute the unconstrained flesh mesh vertex positions $X^U$.

\section{Muscle Tracks} \label{sec:muscle_tracks}

The muscle tracks simulation introduced by \cite{cong2016art} modifies the framework described in Section \ref{sec:simulation} such that the muscle deformations are primarily controlled by a volumetric morph \cite{ali2013anatomy,cong2015fully} rather than directly using muscle activation values.
\cite{cong2016art} first creates a correspondence between the neutral pose $n$ of the blendshape system and the outer boundary surface of the tetrahedral mesh $X^b$.
Then, given a blendshape target expression $x^*(b, j)$ with surface mesh displacements $x^* - n$, \cite{cong2016art} creates target displacements for the outer boundary of the tetrahedral mesh $\delta X^b$.
Using $\delta X^b$ as Dirichlet boundary conditions, \cite{cong2016art} solves a Poisson equation for the displacements $\delta X = X - X_0$, \ie $\nabla^2 \delta X = 0$, where $X_0$ are the rest-state vertex positions consistent with the neutral pose $n$.
Neumann boundary conditions are used on the inner boundary of the tetrahedral mesh.
Afterwards, zero-length springs are attached between the tetrahedralized flesh mesh vertices interior to each muscle and their corresponding target locations resulting from the Poisson equation.
The muscle track force resulting from the zero-length springs for each muscle $m$ has the form
\begin{align}
   f_{\text{tracks},m} & = K_m (M_m - I_m X^U)\label{eq:muscle_track_force}
\end{align}
where $K_m$ is the per-muscle spring stiffness matrix, $I_m$ is the selector matrix for the flesh mesh vertices interior to the muscle, and $M_m$ are the target locations resulting from the volumetric morph.
Thus the expanded quasistatics equation can be written as
\begin{align}
f_{\text{fvm}} + f_{\text{collisions}} + f_{\text{tracks}} & = 0 \label{eq:expanded_quasistatics}
\end{align}
where $f_{\text{tracks}}$ includes Equation \ref{eq:muscle_track_force} for every muscle.
Since the activation values $a$ are no longer specified manually, they must be computed automatically given the final post-morph shape of a muscle to reintroduce the effects of muscle tension into the simulation.
\cite{cong2016art} barycentrically embeds a piecewise linear curve into each muscle and uses the length of that curve to determine an appropriate activation value.

\section{Blendshape-Driven Muscle Tracks} \label{sec:blendshape_muscle_tracks}

The morph from Section \ref{sec:muscle_tracks} was designed in the spirit of the computer graphics pipeline, and as such, does not allow for the sort of full end-to-end coupling that facilitates differentiability, inversion, and other typical inverse problem methodologies.
Thus, our key contribution is to replace the morphing step with a blendshape deformation in the form of Equation \ref{eq:blendshapes} to drive the muscle volumes and their center-line curves thereby creating a direct functional connection between the animator controls $w$ and the muscle tracks target locations $M_m$ and activation values $a$.

For each muscle, we create a tetrahedralized volume $M_m^0$ and piecewise linear center-line curve $C_m^0$ in the neutral pose.
Furthermore, for each blendshape in the face surface model, we use the morph from \cite{cong2016art} to create a corresponding shape for each muscle's tetrahedralized volume $M_m^k$ and center-line curve $C_m^k$, where $k$ is used to denote the $k$th blendshape.
Alternatively, one could morph and subsequently simulate as in Section \ref{sec:muscle_tracks} using tracks in order to create $M_m^k$ and $C_m^k$.
In addition, we assign skinning weights to each vertex in $M_m^0$ and $C_m^0$ and assemble them into linear blend skinning transformation matrices $T^M_m$ and $T^C_m$.
This allows us to write
\begin{align}
    M_m(b, j) & =  T^M_m(j) \left( M_m^0 + \sum_kM_m^k b_k \right) \label{eq:blendshape_muscles} \\
    C_m(b, j) & =  T^C_m(j) \left( C_m^0 + \sum_kC_m^k b_k \right) \label{eq:blendshape_curves}
\end{align}
which parallel Equation \ref{eq:blendshapes}.
Notably, we are able to obtain Equations \ref{eq:blendshape_muscles} and \ref{eq:blendshape_curves} in part because we solve the Poisson equation on the pre-skinning neutral as compared to \cite{cong2016art} which uses the post-skinning neutral.
In addition, this better prevents linearized rotation artifacts from diffusing into the tetrahedralized flesh mesh.
Finally, we can write the length of each center-line curve as
\begin{align}
    L(C_m(b,j)) & = \sum_{i} \left|\left| C_{m,i}(b,j) - C_{m,i-1}(b,j) \right|\right|_2 \label{eq:length_center_line_curve}
\end{align}
where $C_{m,i}(b,j)$ is the $i$th vertex of the piecewise linear center-line curve for the $m$th muscle.

To justify our approach, we can write the linear system to solve the Poisson equation as $A^U(X_0) \delta X = A^C(X_0) Bb$ where $A^U(X_0)$ is the portion of the Laplacian matrix discretized on the tetrahedralized volume at rest using the method of \cite{zheng2015new} for the unconstrained vertices.
Similarly, $A^C(X_0)$ is the portion for the constrained vertices post-multiplied by the linear correspondence between the neutral pose $n$ of the blendshape system and the outer boundary of the tetrahedral mesh $X^b$.
Equivalently, we may write
\begin{align}
    A^U(X_0)\delta X = \sum_k A^C(X_0) Be_k b_k \label{eq:expanded_poisson_linear}
\end{align}
(where $e_k$ are the standard basis vectors) which is equivalent to doing $k$ solves of the form
\begin{align}
A^U(X_0) \delta X_k = A^C(X_0)Be_k \label{eq:separate_poisson_linear}
\end{align}
and then subsequently summing both sides to obtain $\delta X = \sum_k \delta X_k b_k$.
That is, the linearity of the Poisson equation allows us to precompute its action for each blendshape and subsequently obtain the exact result on any combination of blendshapes by simply summing the results obtained on the individual blendshapes.

In summary, for each of the $k$ blendshapes, we solve a Poisson equation (Equation \ref{eq:separate_poisson_linear}) to precompute $M_m^k$ and $C_m^k$, and then given animator controls $w$ which yield $b$ and $j$, we obtain $M_m$ and $C_m$ via Equations \ref{eq:blendshape_muscles} and \ref{eq:blendshape_curves}.
This replaces the morphing step allowing us to proceed with the quasistatic muscle simulation using tracks driven entirely by the animator parameters $w$.

\section{End-to-End Differentiability} \label{sec:differentiable_system}

In this section, we outline the derivative of the simulated tetrahedral mesh vertex positions with respect to the blendshape parameters $b$ and jaw controls $j$ that parameterize the simulation results as per Section \ref{sec:blendshape_muscle_tracks}.
The derivative of $b$ and $j$ with respect to the animator controls $w$ depend on the specific controls and can be post-multiplied.
If one cares about the resulting vertices of a rendered mesh embedded in or constrained to the tetrahedral mesh, then this embedding, typically linear, can be pre-multiplied.

Although the constrained nodes $X^C$ typically only depend on the joint parameters, one may wish, at times, to simulate only a subset of the tetrahedral flesh mesh.
In such instances, the constrained nodes can appear on the unsimulated boundary which in turn can be driven by the blendshape parameters $b$; thus, we write $X^C(b,j)$ and concatenate it with $X^U(b,j)$ to obtain $X(b,j)$ for the purposes of this section.
The collision forces only depend on the nodal positions, and we may write $f_\text{collisions}(X(b,j))$.
The finite volume force depends on both the nodal positions and activations, and the activations are determined from an activation-length curve where the length is given in Equation \ref{eq:length_center_line_curve}.
Our precomputation makes $C_m$ only a function of $b$ and $j$ and notably independent of $X$, and so we may write $a_m(L_m(C_m(b,j)))$ combining the activation length curve with Equations \ref{eq:blendshape_curves} and \ref{eq:length_center_line_curve}.
We stress that the activations are independent of the positions, $X$.
Thus, we may write $f_\text{fvm}(X(b,j),C(b,j))$.
Similarly, we may write $f_\text{tracks}(X(b,j), M(b,j))$.
Therefore, all the forces in Equation \ref{eq:expanded_quasistatics} are a function of $X$, $C$, and $M$ which are in turn a function of $b$ and $j$.

Using the aforementioned dependencies, we can take the total derivative of the forces $f_T = f_\text{fvm} + f_\text{collisions}$ in Equation \ref{eq:full_quasistatics} with respect to a single blendshape parameter $b_k$ to obtain $(\partial f_T / \partial X) (\partial X / \partial b_k) + (\partial f_T / \partial C)(\partial C / \partial b_k) = 0$ which is equivalent to $(\partial f_T / \partial X) (\partial X / \partial b_k) + (\partial f_\text{fvm} / \partial C)(\partial C / \partial b_k) = 0$ since $f_\text{collisions}$ is independent of $C$.
Since our activations are still independent of $X$ just as they were in \cite{sifakis2005automatic}, $\partial f_T / \partial X$ here is identical to that discussed in \cite{sifakis2005automatic}, and thus their quasistatic solve can be used to determine $\partial X / \partial b_k$ by solving $(\partial f_T / \partial X) (\partial X / \partial b_k) = - (\partial f_\text{fvm} / \partial C) (\partial C / \partial b_k)$.
To compute the right hand side, note that $\partial C / \partial b_k$ can be obtained from Equation \ref{eq:blendshape_curves}.
To obtain $\partial f_\text{fvm} / \partial C$, we compute $\partial f_\text{fvm} / \partial C = (\partial f_\text{fvm} / \partial a) (\partial a / \partial L) (\partial L / \partial C)$.
$\partial f_\text{fvm} / \partial a$ are simply the active forces $f_a$ in Equation \ref{eq:muscle_forces}, $\partial a / \partial L$ is the local slope of the activation length curve, and $\partial L / \partial C$ is readily computed from Equation \ref{eq:length_center_line_curve}.
The $\partial X / \partial j_k$ are determined similarly.

One may take a similar approach to Equation \ref{eq:expanded_quasistatics}, obtaining $\partial X / \partial b_k$ by solving $(\partial f_T / \partial X) (\partial X / \partial b_k) = - (\partial f_\text{fvm} / \partial C) (\partial C / \partial b_k) - (\partial f_\text{tracks} / \partial M) (\partial M / \partial b_k)$.
We stress that the coefficient matrix $\partial f_T / \partial X$ of the quasistatic solve is now augmented by $\partial f_\text{tracks} / \partial X$ (see Equation \ref{eq:muscle_track_force}) and is the same quasistatic coefficient matrix in \cite{cong2016art}.
$\partial f_\text{tracks} / \partial M$ and $\partial M / \partial b_k$ are obtained from Equations \ref{eq:muscle_track_force} and \ref{eq:blendshape_muscles} respectively.
Again, the $\partial X / \partial j_k$ are found similarly.
In summary, finding $\partial X / \partial b_k$ and $\partial X / \partial j_k$ involves solving the same quasistatics problem of \cite{sifakis2005automatic} with the slight augmention to the coefficient matrix from \cite{cong2016art} merely with different right hand sides.
Although this requires a quasistatic solve for each $b_k$ and $j_k$, they are all independent and can thus be done in parallel.

\section{Experiments} \label{sec:experiments}

We use the Dogleg optimization algorithm \cite{lourakis2005levenberg} as implemented by the Chumpy autodifferentation library \cite{loperchumpy} in order to target our face model to both three-dimensional geometry and two-dimensional RGB images to demonstrate the efficacy of our end-to-end fully differentiable formulation.
Other optimization algorithms and/or applications may similarly be pursued.
Our nonlinear least squares optimization problems generally have the form
\begin{align}
\text{min}_w ||F^* - F(x_R(w))||_2^2 + \lambda||w||_2^2 \label{eq:nllq_problem}
\end{align}
where $w$ are the animator controls that deform the face, $x(w)$ are the positions of the vertices on the surface of the face deformed using the full blendshape-driven muscle simulation system as described in Section \ref{sec:blendshape_muscle_tracks}, $F(x_R(w))$ is a function of those vertex positions, and $F^*$ is the desired output of that function.
$R(\theta)$ and $t$ are an additional rigid rotation and translation, respectively where $\theta$ represents Euler angles, \ie $x_R(w) = R(\theta) x(w) + t$ .
We use a standard L2 norm regularization on the animator controls $||w||_2^2$, where $\lambda$ is set experimentally to avoid overfitting.

\subsection{Model Creation} \label{sec:model_creation}

\begin{figure}[t]
\centering
\begin{subfigure}[b]{0.24\linewidth}
    \includegraphics[width=\linewidth]{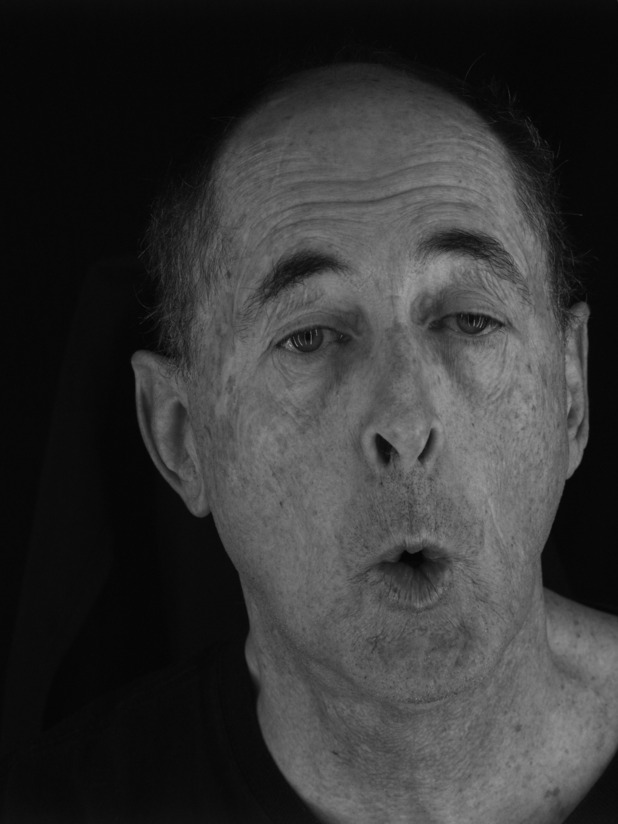}
\end{subfigure}
\begin{subfigure}[b]{0.24\linewidth}
    \includegraphics[width=\linewidth]{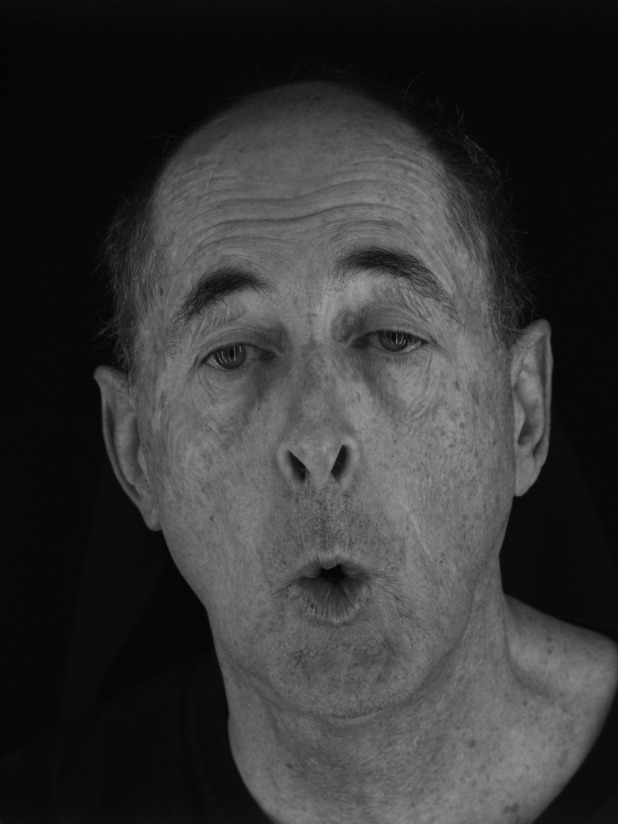}
\end{subfigure}
\begin{subfigure}[b]{0.24\linewidth}
    \includegraphics[width=\linewidth]{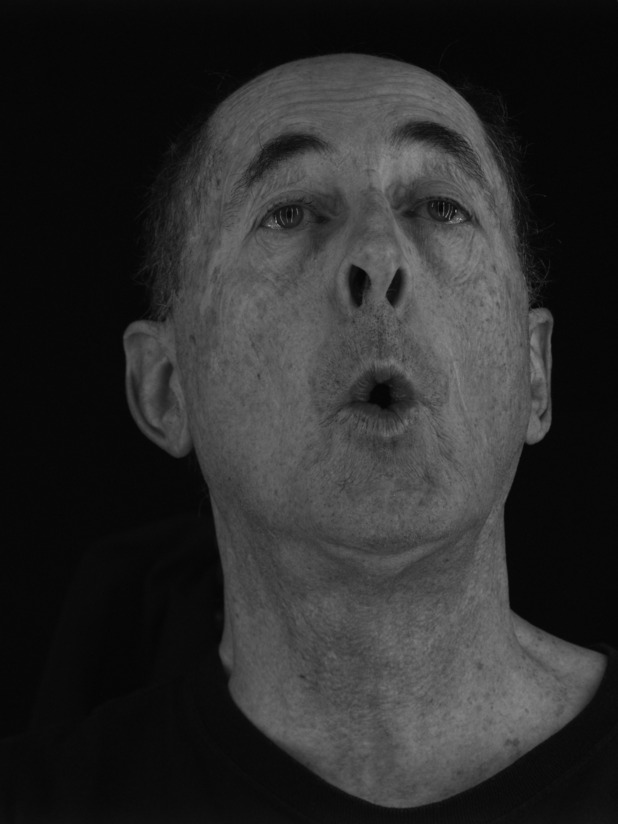}
\end{subfigure}
\begin{subfigure}[b]{0.24\linewidth}
    \includegraphics[width=\linewidth]{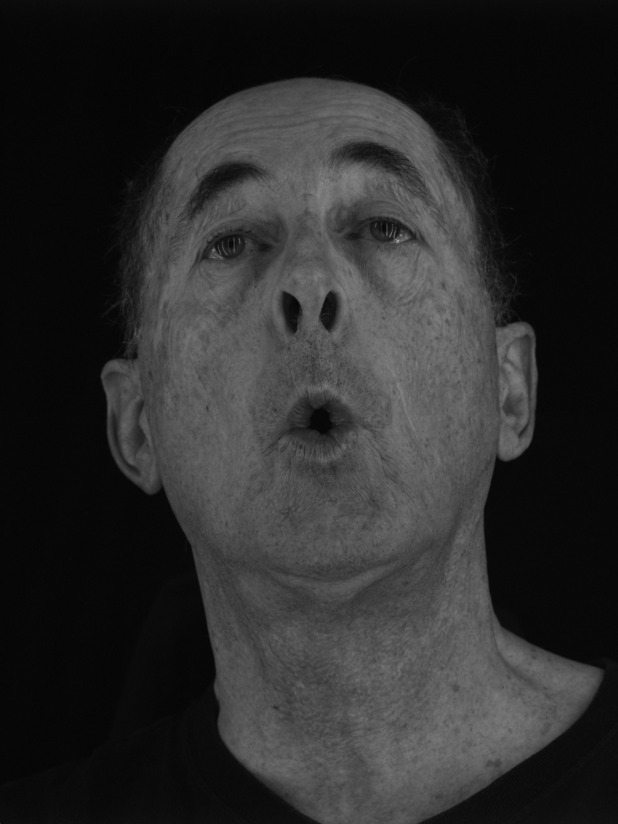}
\end{subfigure}\hfill
\begin{subfigure}[b]{0.24\linewidth}
    \includegraphics[width=\linewidth]{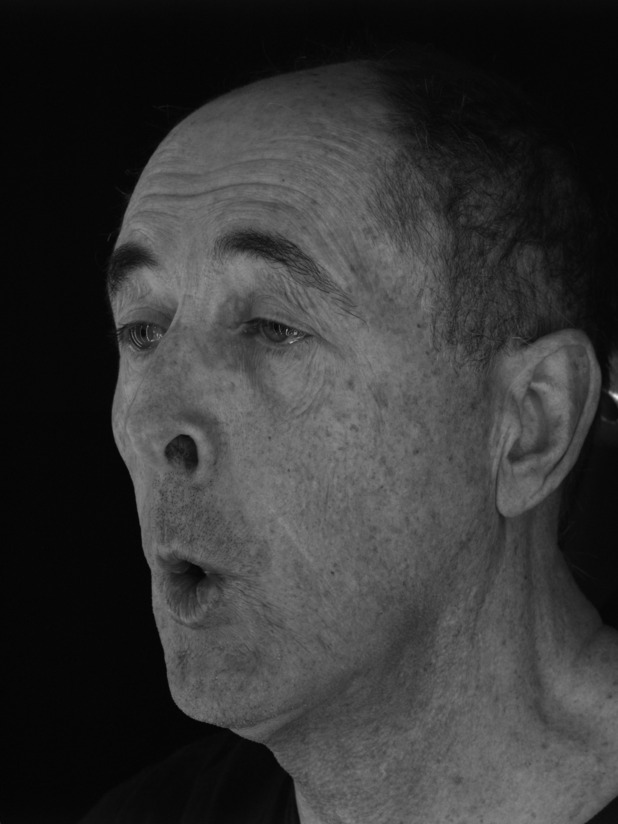}
\end{subfigure}
\begin{subfigure}[b]{0.24\linewidth}
    \includegraphics[width=\linewidth]{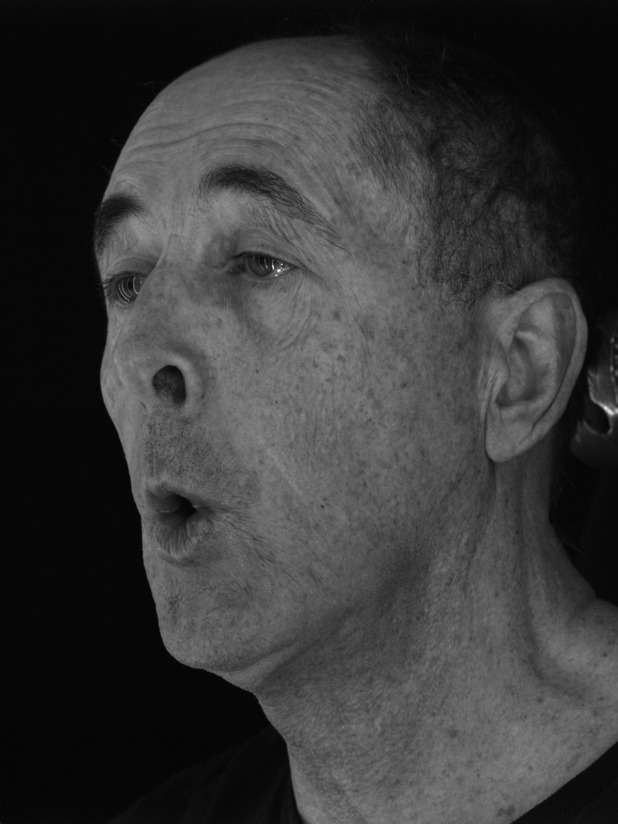}
\end{subfigure}
\begin{subfigure}[b]{0.24\linewidth}
    \includegraphics[width=\linewidth]{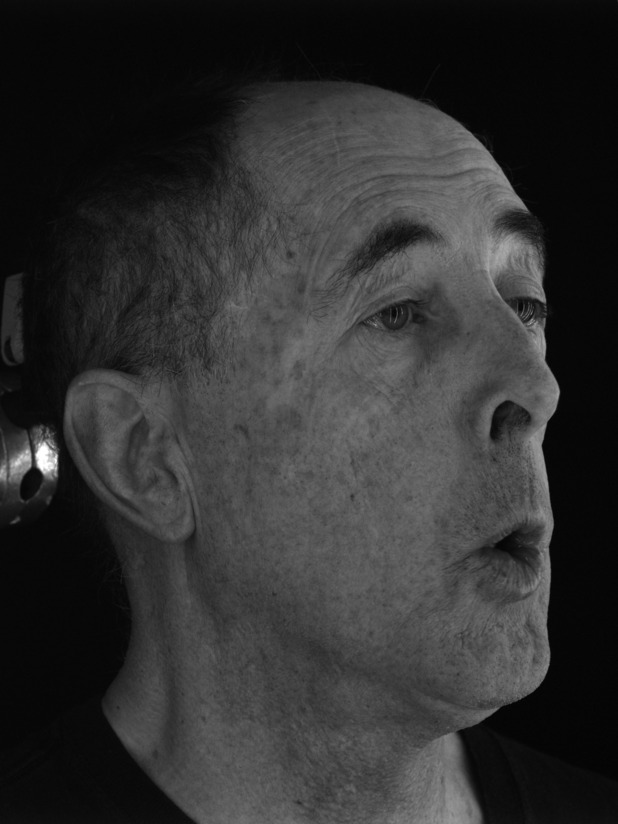}
\end{subfigure}
\begin{subfigure}[b]{0.24\linewidth}
    \includegraphics[width=\linewidth]{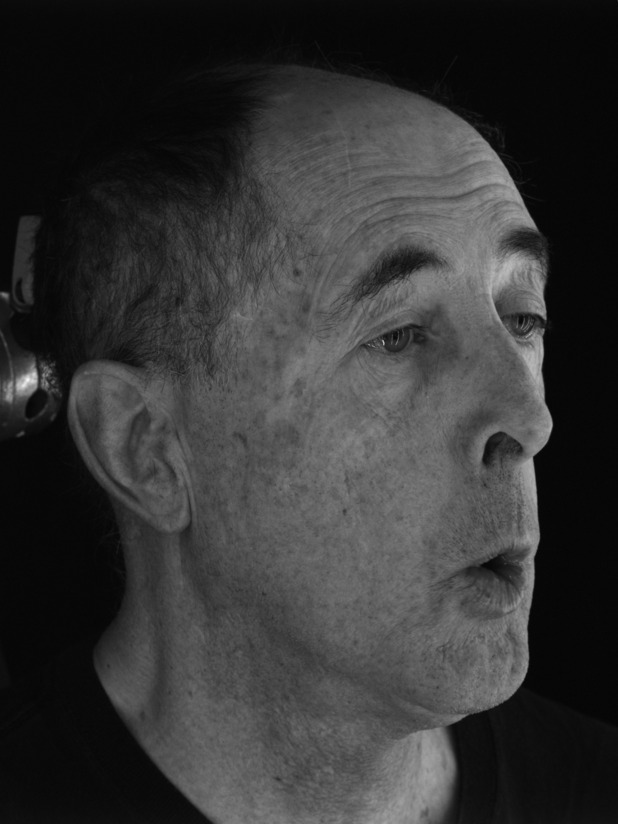}
\end{subfigure}
\caption{The eight viewpoints used to reconstruct the facial geometry for a particular pose.}
\label{fig:medusa_plates}
\end{figure}

\begin{figure}[b]
\centering
\begin{subfigure}[b]{0.32\linewidth}
    \includegraphics[width=\linewidth]{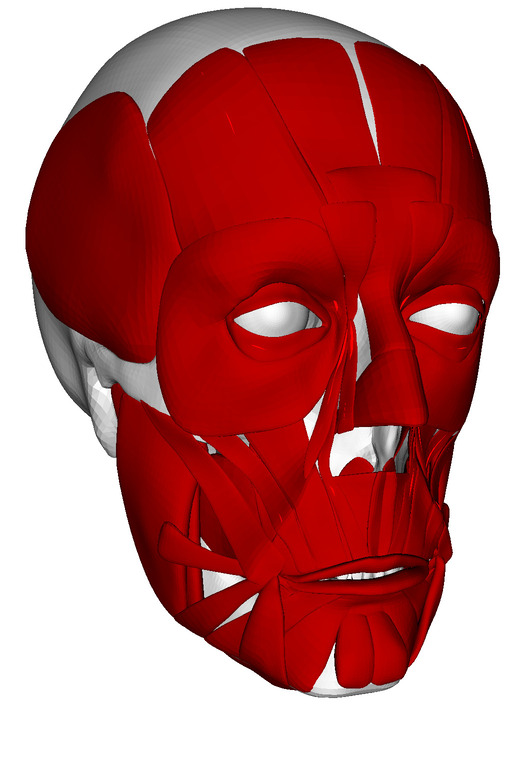}
    \caption{Neutral}
\end{subfigure}
\begin{subfigure}[b]{0.32\linewidth}
    \includegraphics[width=\linewidth]{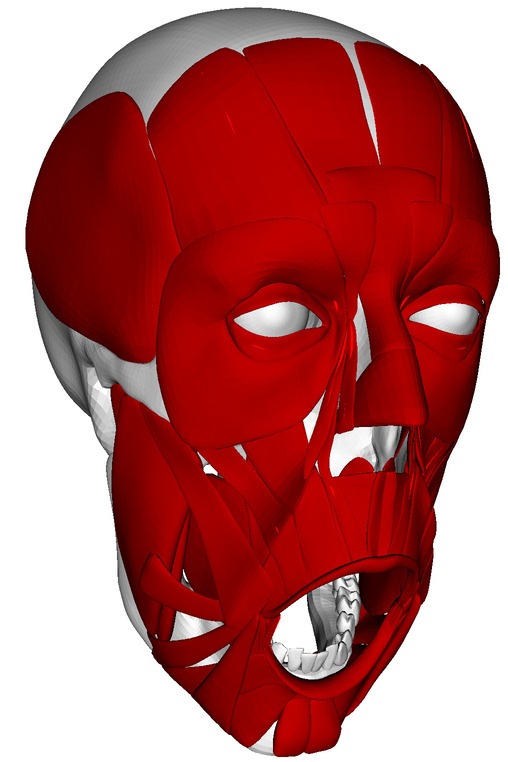}
    \caption{Jaw Open}
\end{subfigure}
\hfill
\caption{The underlying anatomical model of the face in the neutral pose as well as the jaw open pose using linear blend skinning.}
\label{fig:neutral_pose_jaw_muscle_model}
\end{figure}

\begin{figure}[t]
\centering
\begin{subfigure}[b]{0.24\linewidth}
    \includegraphics[width=\linewidth]{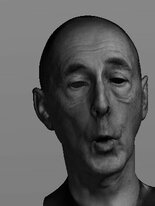}
\end{subfigure}
\begin{subfigure}[b]{0.24\linewidth}
    \includegraphics[width=\linewidth]{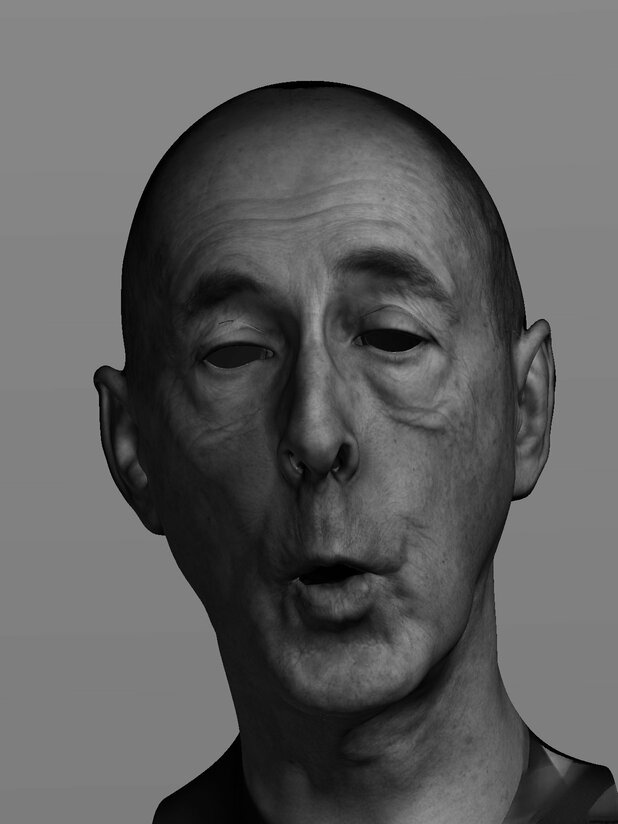}
\end{subfigure}
\begin{subfigure}[b]{0.24\linewidth}
    \includegraphics[width=\linewidth]{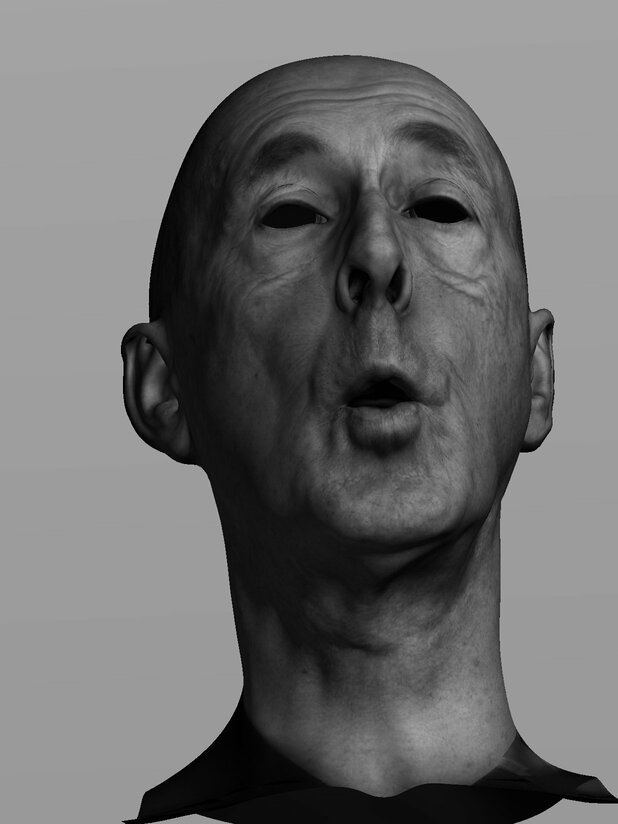}
\end{subfigure}
\begin{subfigure}[b]{0.24\linewidth}
    \includegraphics[width=\linewidth]{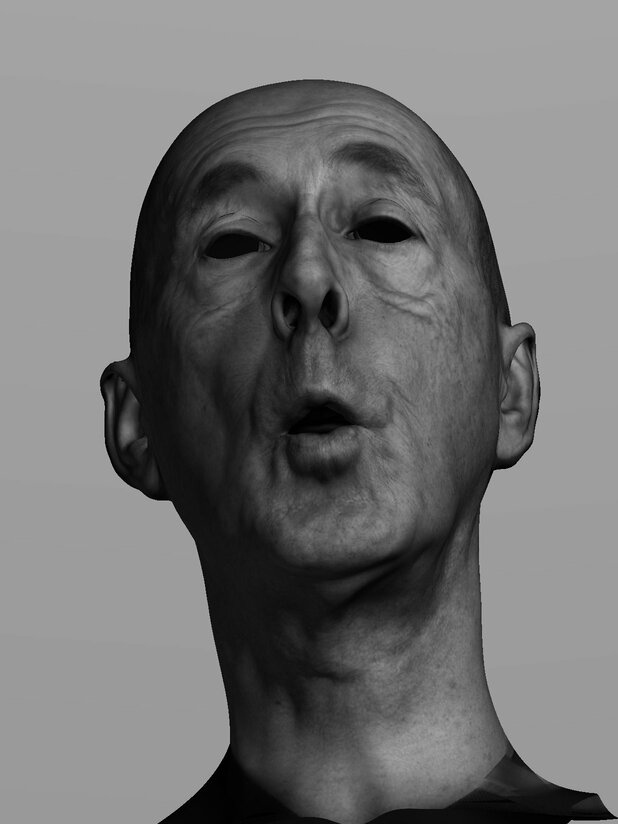}
\end{subfigure}\hfill
\begin{subfigure}[b]{0.24\linewidth}
    \includegraphics[width=\linewidth]{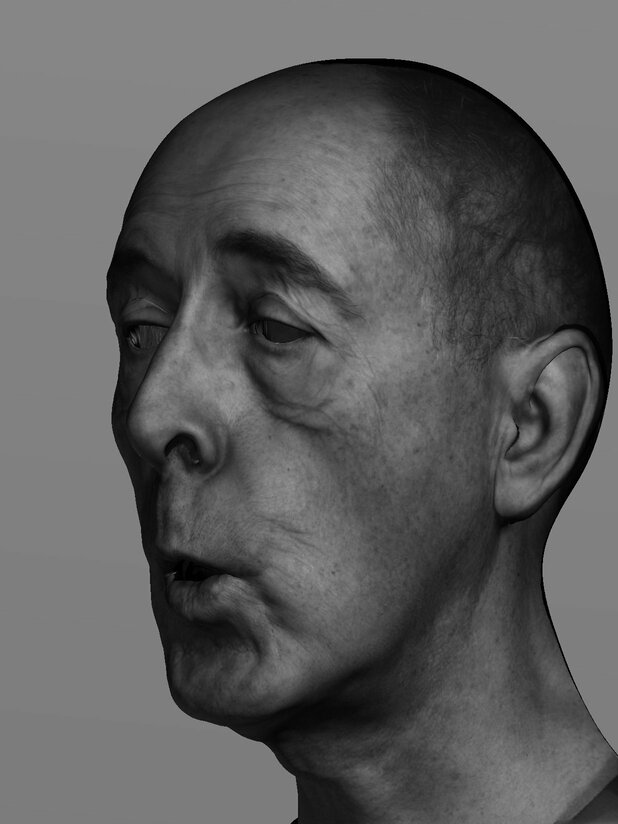}
\end{subfigure}
\begin{subfigure}[b]{0.24\linewidth}
    \includegraphics[width=\linewidth]{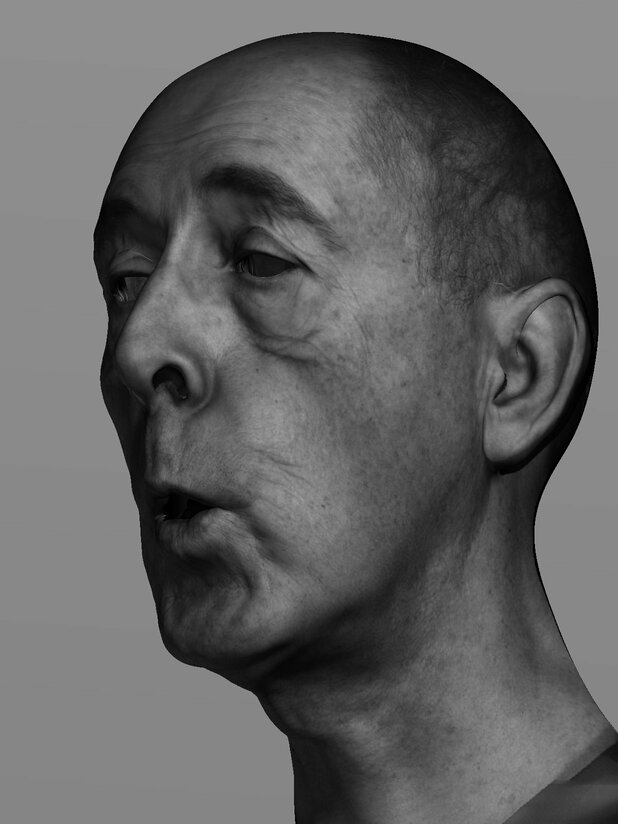}
\end{subfigure}
\begin{subfigure}[b]{0.24\linewidth}
    \includegraphics[width=\linewidth]{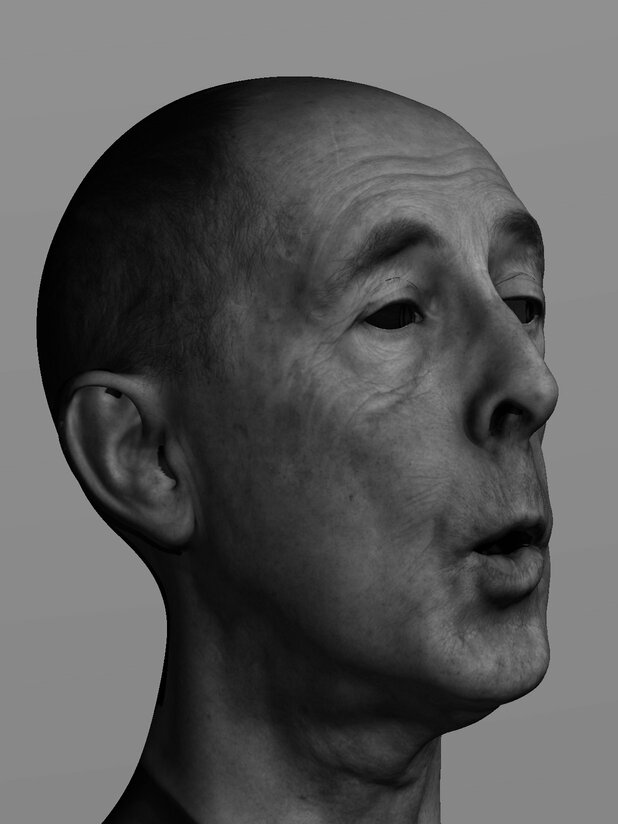}
\end{subfigure}
\begin{subfigure}[b]{0.24\linewidth}
    \includegraphics[width=\linewidth]{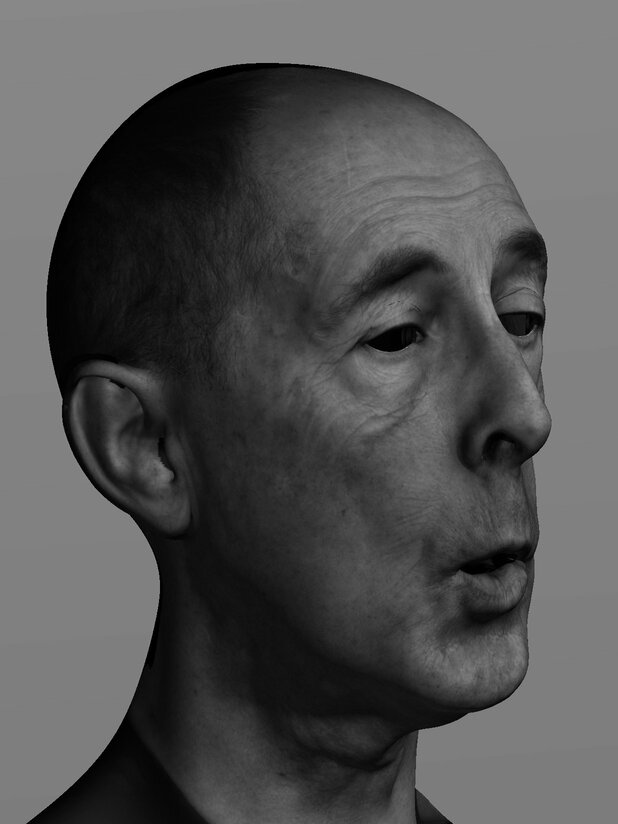}
\end{subfigure}
\caption{The geometry reconstructed by applying the multi-view stereo algorithm described in \cite{beeler2010high,beeler2011high} to the input images shown in Figure \ref{fig:medusa_plates}.}
\label{fig:medusa_raw_geometry}
\vspace{-4.0mm}
\end{figure}

The blendshape system is created from the neutral pose $n$ as well as FACS-based expressions \cite{ekman2002facial} using the methods of \cite{beeler2010high,beeler2011high}.
Eight black and white cameras from varying viewpoints (see Figure \ref{fig:medusa_plates}) are used to reconstruct the geometry of the actor.
Artists clean up these scans and use them as inspiration for a blendshape model and to calibrate the linear blend skinning matrices for the face surface (see Figure \ref{fig:medusa_raw_geometry}).
Of course, any reasonable method could be used to create the blendshape system.
A subset of the full face surface model with $\num{52228}$ surface vertices is used in the optimization.

We use the neutral pose of the blendshape system and the method of \cite{cong2015fully} to create the tetrahedral flesh mesh $X_0$, tetrahedral muscle volumes $M_m^0$, and muscle center line curves $C_m^0$ by morphing them from a template asset.
Our simulation mesh has $\num{302235}$ vertices and $\num{1470102}$ tetrahedra.
We use $\num{60}$ muscles with a total of $\num{50710}$ vertices and $\num{146965}$ tetrahedra (some tetrahedra are duplicated between muscles due to overlap).
The linear blend skinning weights used to form $T_j$ on the face surface are propagated to the surface of the tetrahedral mesh and used as boundary conditions in a Poisson equation solve again as in \cite{ali2013anatomy,cong2015fully} to obtain linear blend skinning weights throughout the volumetric tetrahedral mesh as well as for the muscles and center-line curves, thus defining skinning transformation matrices $T_m^M$ and $T_m^C$.
Figure \ref{fig:neutral_pose_jaw_muscle_model} shows the muscles in the neutral pose $M_m^0$ as well as the result after skinning with the jaw open, \ie Equation \ref{eq:blendshape_muscles} with all $b_k$ identically $0$.

\begin{figure}[b]
\centering
\begin{subfigure}[b]{0.32\linewidth}
    \includegraphics[width=\linewidth]{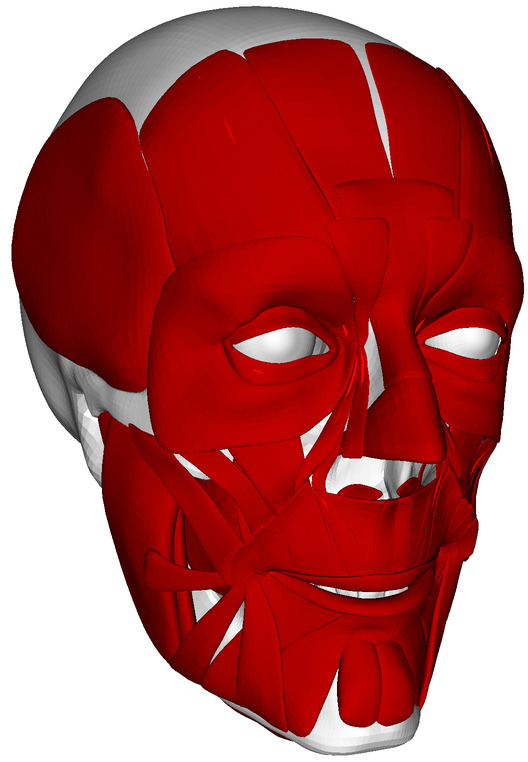}
    \caption{Smile}
\end{subfigure}
\begin{subfigure}[b]{0.32\linewidth}
    \includegraphics[width=\linewidth]{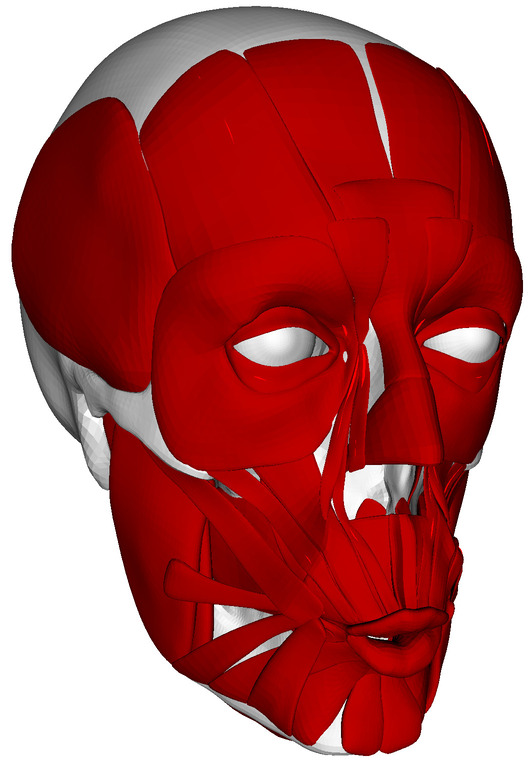}
    \caption{Pucker}
\end{subfigure}
\begin{subfigure}[b]{0.32\linewidth}
    \includegraphics[width=\linewidth]{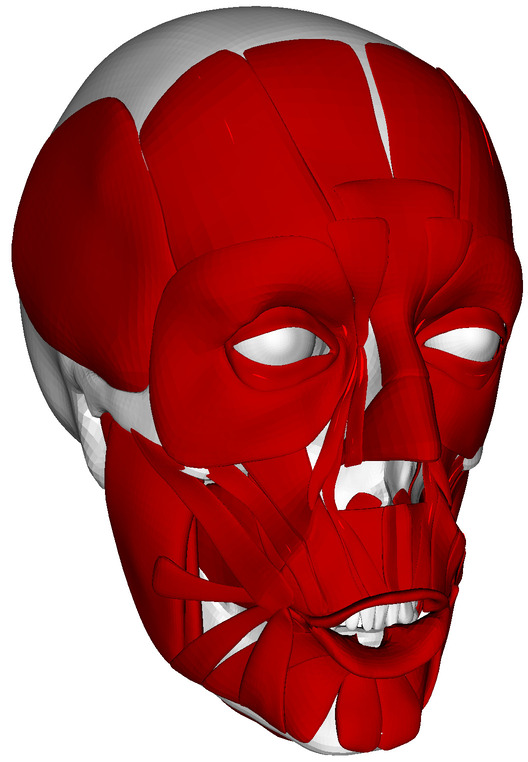}
    \caption{Funneler}
\end{subfigure}
\hfill
\caption{The anatomical model of the face peforming a variety of expressions using only the blendshape deformation from Equation \ref{eq:blendshape_muscles}.}
\label{fig:blendshape_muscle_model}
\end{figure}

Finally, for each shape in the blendshape system, we solve a Poisson equation (Equation \ref{eq:separate_poisson_linear}) for the vertex displacements $\delta X_k$ which are then transferred to the muscle volumes and center-line curves to obtain $M_m^k$ and $C_m^k$.
This allows us full use of Equations \ref{eq:blendshape_muscles} and \ref{eq:blendshape_curves} parameterized by the blendshapes $b_k$.
Figure \ref{fig:blendshape_muscle_model} shows some examples of the muscles evaluated using Equation \ref{eq:blendshape_muscles} for a variety of expressions.

\subsection{Targeting 3D Geometry} \label{sec:target_3d_geom}

\begin{figure}[b]
\centering
\begin{subfigure}[b]{0.32\linewidth}
    \includegraphics[width=\linewidth]{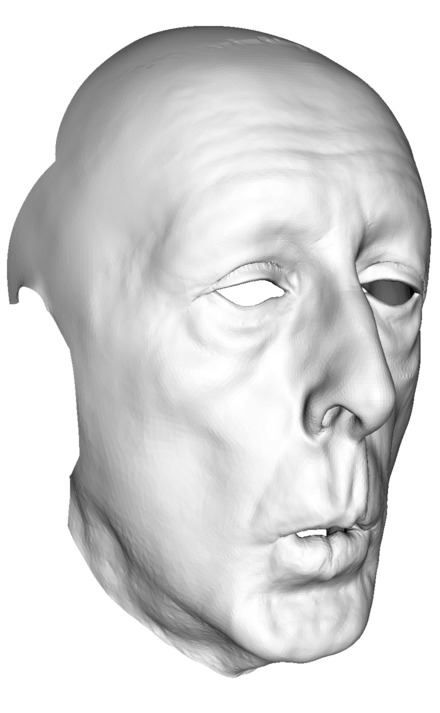}
    \caption{Blendshape}
\end{subfigure}
\begin{subfigure}[b]{0.32\linewidth}
    \includegraphics[width=\linewidth]{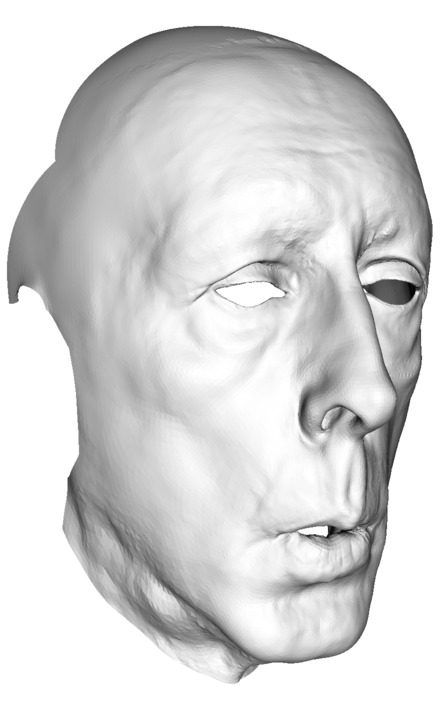}
    \caption{Simulation}
\end{subfigure}
\begin{subfigure}[b]{0.32\linewidth}
    \includegraphics[width=\linewidth]{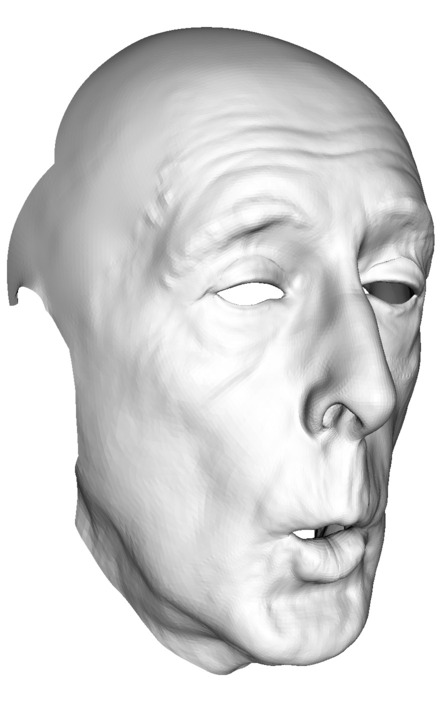}
    \caption{Target}
\end{subfigure}
\hfill
\begin{subfigure}[b]{0.32\linewidth}
    \includegraphics[width=\linewidth]{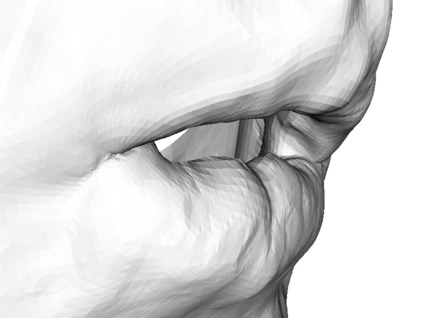}
\end{subfigure}
\begin{subfigure}[b]{0.32\linewidth}
    \includegraphics[width=\linewidth]{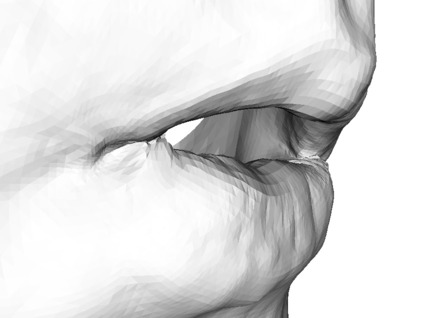}
\end{subfigure}
\begin{subfigure}[b]{0.32\linewidth}
    \includegraphics[width=\linewidth]{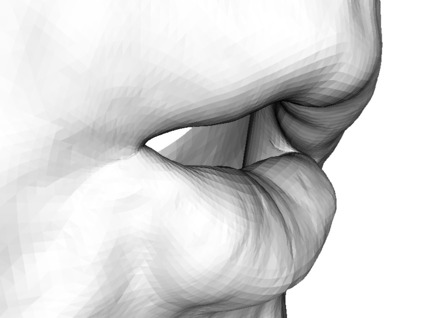}
\end{subfigure}
\hfill
\caption{We target the geometry shown in (c) using purely blendshapes shown in (a) versus the blendshape driven muscle simulation model shown in (b).
While neither method exactly matches target geometry, in general, we found that the simulation results preserve key physical properties such as volume preservation around the lips.
A close-up of the lips is shown in the bottom row where it is more apparent how the pure blendshape inversion has significant volume loss around the lips.}
\label{fig:results_3d_geometry}
\end{figure}

Oftentimes, one has captured a facial pose in the form of three-dimensional mesh; however, this data is generally noisy, and it is desirable to convert this data into a lower dimensional representation.
Using a lower dimensional representation facilitates editing, extracting semantic information, and performing statistical analysis.
In our case, the lower dimensional representation is the parameter space of the blendshape or simulation model.

In general, extracting a lower dimensional representation from an arbitrary mesh requires extracting a set of correspondences between the mesh and the face model.
However, for simplicity, we assume that the correspondence problem has been solved beforehand and that each vertex of the incoming mesh captured by a system using the methods of \cite{beeler2010high,beeler2011high} has a corresponding vertex on our face surface.
We can thus use an optimization problem in the form of Equation \ref{eq:nllq_problem} to solve for the model parameters where $F^*$ are the vertex positions of the target geometry, and $F(x) = x$ is the identity function.

While a rigid alignment between the $F^*$ and the neutral mesh $n$, \ie $R(\theta)$ and $t$, is created as a result of \cite{beeler2010high,beeler2011high}, we generally found it to be inaccurate.
As a result, we also allow the optimization to solve for $\theta$ and $t$ as well.
Our optimization problem for targeting three-dimensional geometry thus has the form 
\begin{align}
\text{min}_{w,\theta,t} ||F^* - x_R(w) ||_2^2 + \lambda ||w||_2^2
\label{eq:invert_geometry_nllq}
\end{align}
where $\lambda = \num{1e-6}$ is set experimentally.

\begin{figure}[b]
\centering
\begin{subfigure}[b]{0.49\linewidth}
    \includegraphics[width=\linewidth]{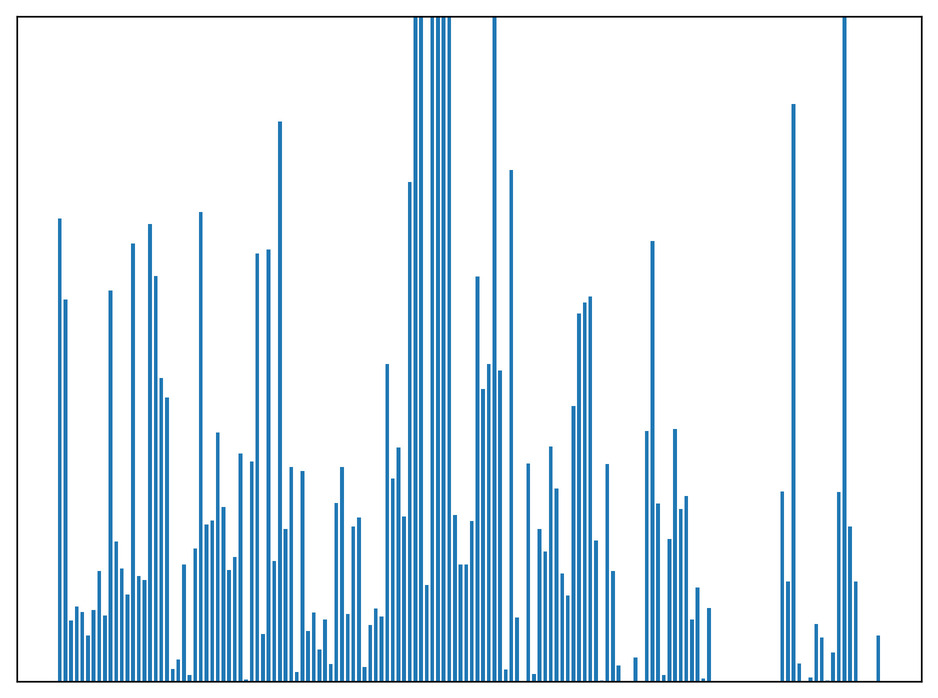}
    \caption{Blendshape Weights}
\end{subfigure}
\begin{subfigure}[b]{0.49\linewidth}
    \includegraphics[width=\linewidth]{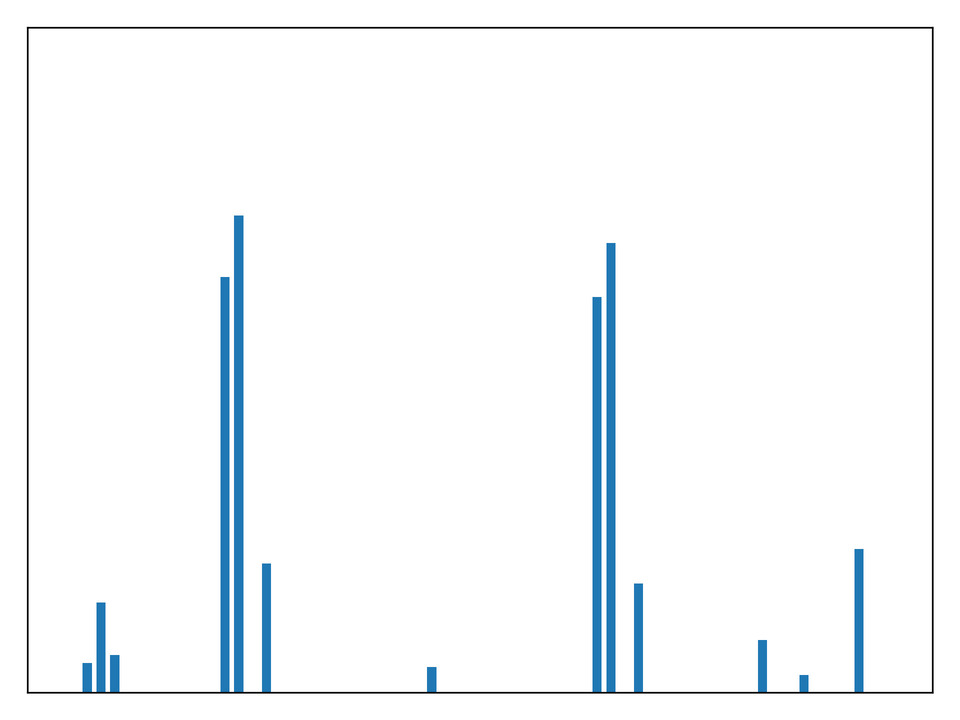}
    \caption{Muscle Activations}
\end{subfigure}
\hfill
\caption{The blendshape solve results in blendshape weights that are dense, overdialed, and hard to decipher.
The largest weights are related to closing the mouth (with magnitudes ranging from \num{6.5} to \num{2.77}, \ie three to six times taller than what is shown in the figure).
It is not until the $11$th most dialed in shape that we see a blendshape related to the pucker.
Whereas all \num{129} (of \num{146}; shapes for the neck, etc.~were not used) blendshapes used have non-zero values, only \num{13} of the available \num{60} muscles have non-zero activation values.
The top four most activated muscles are related to the frontalis indicating that the eyebrows are raised \cite{standring2015gray}.
The activations of the incisivus labii superioris and orbicularis oris muscles are also among the top activated muscles properly indicating a compression of the lips \cite{hur2018anatomical,standring2015gray}.
These muscle activations succintly describe the performance of the actor in this frame.}
\label{fig:bs_vs_activations_geometry}
\end{figure}

We demonstrate the efficacy of our method on a pose where the actor has his mouth slightly open and is making a pucker shape.
We compare the results of targeting three-dimensional geometry when it is driven using simulation via the blendshape muscle tracks as described in Section \ref{sec:blendshape_muscle_tracks} versus when it is driven using the pure blendshape model described in Section \ref{sec:blendshapes}.
Traditionally, pucker shapes have been difficult for activation-muscle based simulations to hit.
See Figure \ref{fig:results_3d_geometry}.
Although neither inversion quite captures the tightness of the mouth's pucker, the muscle simulation results demonstrate how the simulation's volume preservation property significantly improves upon the blendshape results where the top and bottom lips seem to shrink.
This property is also useful in preserving the general shape of the philtrum; the blendshape models's inversion causes the part of the philtrum near the nose to incorrectly bulge significantly.
Furthermore, the resulting muscle activation values are easier to draw semantic meaning from due to their sparsity and anatomical meaning as seen in Figure \ref{fig:bs_vs_activations_geometry}.

Note that errors in the method of \cite{beeler2010high,beeler2011high} in performing multi-view reconstruction will cause the vertices of the target geometry to contain noise and potentially be in physically implausible locations.
Additionally, errors in finding correspondences between the target geometry and the face surface will result in an inaccurate objective function.
Furthermore, there is no guarantee that our deformation model $x(w)$ is able to hit all physically attainable poses even when the capture and correspondence are perfect.
This demonstrates the efficacy of introducing physically-based priors into the optimization.
Additional comparisons and results are shown in the supplementary material and video.

\subsection{Targeting Monocular RGB Images} \label{sec:target_images}

\begin{figure}[b]
\centering
\begin{subfigure}[b]{0.35\linewidth}
    \includegraphics[width=\linewidth]{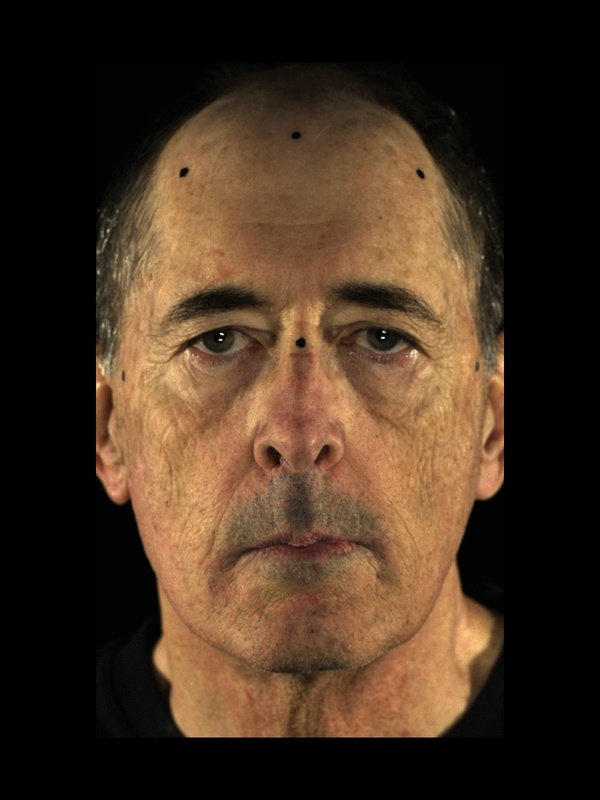}
    \caption{Plate}
\end{subfigure}
\begin{subfigure}[b]{0.35\linewidth}
    \includegraphics[width=\linewidth]{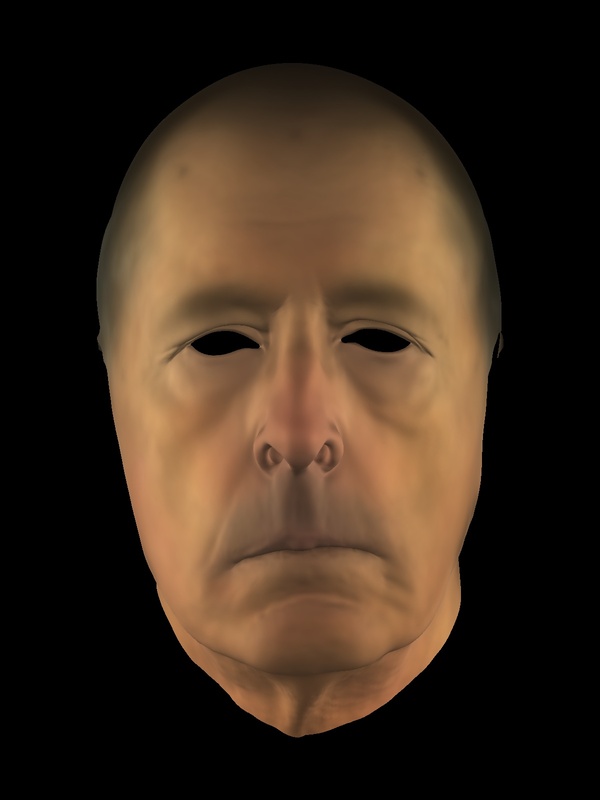}
    \caption{Lighting/Albedo}
\end{subfigure}
\hfill
\caption{Before estimating the facial pose, we first estimate lighting and albedo on a neutral or close to neutral pose.}
\label{fig:result_albedo_lighting}
\end{figure}

To further demonstrate the efficacy of our approach, we consider facial reconstruction from monocular RGB images.
The images were captured using an $100$mm lens attached to an ARRI Alexa XT Studio running at 24 frames-per-second with an 180 degree shutter angle at ISO $800$.
We refer to images captured by the camera as the ``plates.''
The original plates have a resolution of $2880 \times 2160$, but we downsample them to $720 \times 540$.
The camera was calibrated using the method of \cite{heikkila1997four} and the resulting distortion parameters are used to undistort the plate to obtain $F^*$.

$F(x)$ renders the face geometry in its current pose with a set of camera, lighting, and material parameters.
We use a simple pinhole camera with extrinsic parameters determined by the camera calibration step.
The rigid transformation of the face is determined by manually tracking features on the face in the plate.
The face model is lit with a single spherical harmonics light with $9$ coefficients $\gamma$, see \cite{ramamoorthi2001efficient}, and is shaded with Lambertian diffuse shading.
Each vertex $i$ also has an RGB color $c_i$ associated with it.
We solve for $\gamma$ and all $c_i$ using a non-linear least squares optimization of the form
\begin{align}
\text{min}_{\gamma,c} ||F^* - F(x_R(0), \gamma, c)||_2^2 + \lambda ||S(c)||_2^2
\label{eq:albedo_light_nllq}
\end{align}
where the per-vertex colors is regularized using $S(c) = \sum_i \sum_{j \in N(i)} c_i - c_j$ where $N(i)$ are the neighboring vertices of vertex $i$.
This lighting and albedo solve is done as a preprocess on a neutral or close to neutral pose with $\lambda = 2500$ set experimentally.
OpenDR \cite{loper2014opendr} is used to differentiate $F(x)$ to solve Equation \ref{eq:albedo_light_nllq}; however, any other differentiable renderer (\eg \cite{li2018differentiable}) can be used instead.
Then we assume that $\gamma$ and $c$ stay constant throughout the performance.
See Figure \ref{fig:result_albedo_lighting}.

\begin{figure}[b]
\centering
\begin{subfigure}[b]{0.32\linewidth}
    \includegraphics[width=\linewidth]{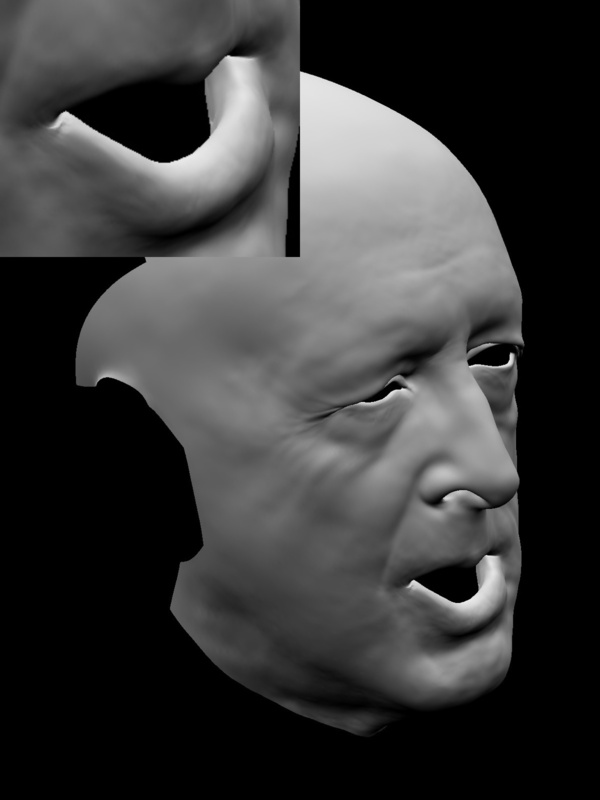}
    \caption{Blendshapes}
\end{subfigure}
\begin{subfigure}[b]{0.32\linewidth}
    \includegraphics[width=\linewidth]{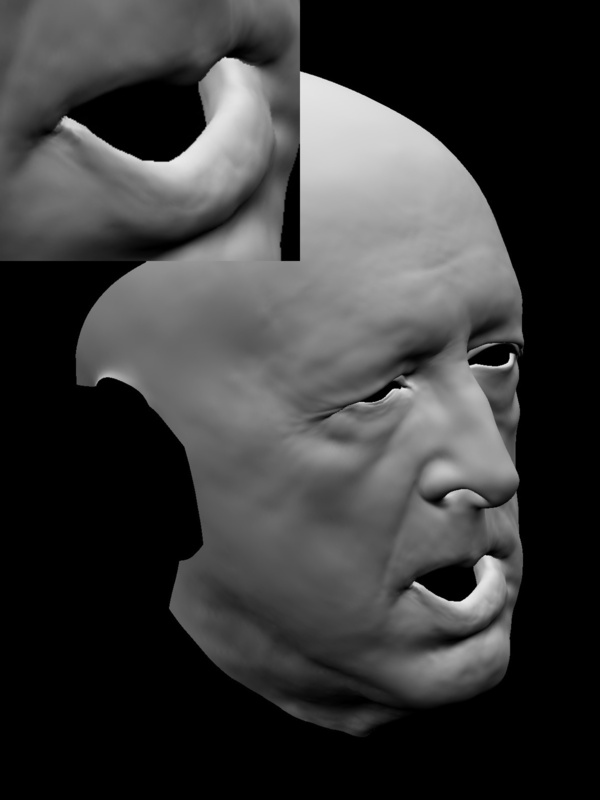}
    \caption{Simulation}
\end{subfigure}
\begin{subfigure}[b]{0.32\linewidth}
    \includegraphics[width=\linewidth]{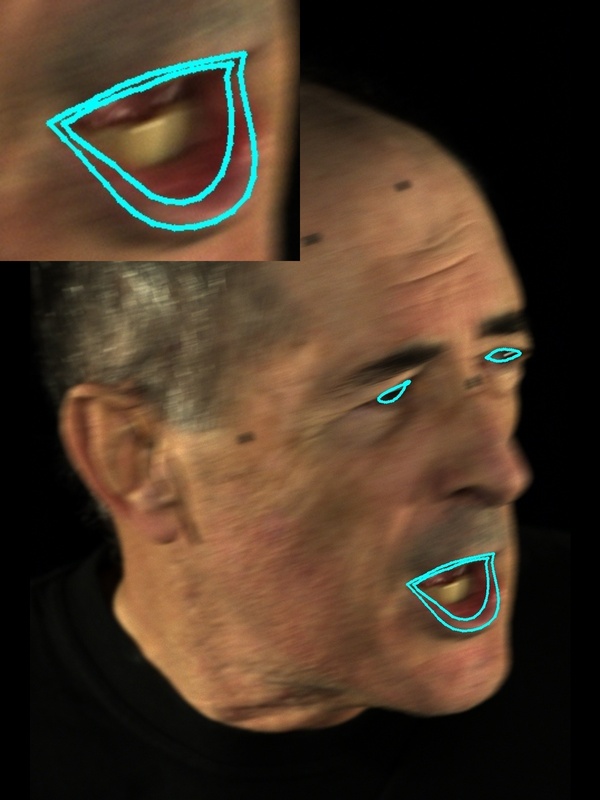}
    \caption{Roto Curves}
\end{subfigure}
\hfill
\caption{We use rotoscoped curves on the plate to solve for an initial estimate of the face pose.}
\label{fig:result_roto}
\end{figure}

We solve for the parameters $w$ in two steps.
Given curves around the eyes and lips on the three-dimensional neutral face mesh, a rotoscope artist draws corresponding curves on the two-dimensional film plate.
Then, we solve for an initial guess $\hat{w}$ by solving an optimization problem of the form
\begin{align}
\text{min}_{\hat{w}} ||E_1(\hat{w})||_2^2 + \lambda_1 ||\hat{w}||_2^2
\label{eq:roto_nllq}
\end{align}
where $\lambda_1 = 3600$ is set experimentally.
$E_1(\hat{w})$ is the two-dimensional Euclidean distance between the points on the rotoscoped curves  on the plate and the corresponding points on the face surface $x(w)$ projected into the image plane.
See Figure \ref{fig:result_roto}.
We then use $\hat{w}$ to initialize a shape from shading solve
\begin{align}
\text{min}_{w} ||E_2(w)||_2^2 + \lambda_1 ||E_1(w)||_2^2 + \lambda_2 ||w - \hat{w}||_2^2
\label{eq:deformation_nllq}
\end{align}
to determine the final parameters $w$ where $\lambda_1 = \num{1e-4}$ and $\lambda_2 = \num{1}$ are set experimentally.
Here, $E_2 = G(F^* - F(x_R(w), \gamma, c))$ is a three-level Gaussian pyramid of the per-pixel differences between the plate and the synthetic render.

\begin{figure}[t]
\centering
\begin{subfigure}[b]{\dimexpr0.31\linewidth+10pt\relax}
    \makebox[10pt]{\raisebox{40pt}{\rotatebox[origin=c]{90}{1112}}}%
    \includegraphics[width=\dimexpr\linewidth-10pt\relax]{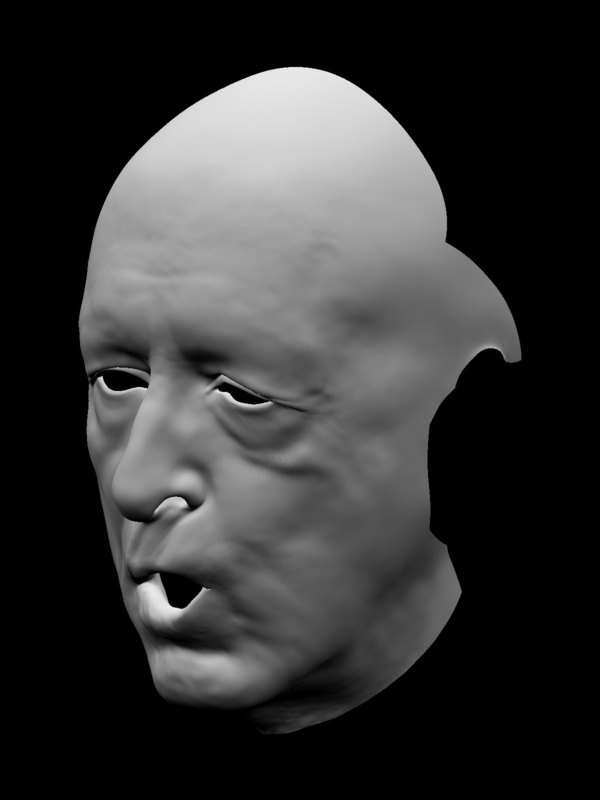}
    \makebox[10pt]{\raisebox{40pt}{\rotatebox[origin=c]{90}{1134}}}%
    \includegraphics[width=\dimexpr\linewidth-10pt\relax]{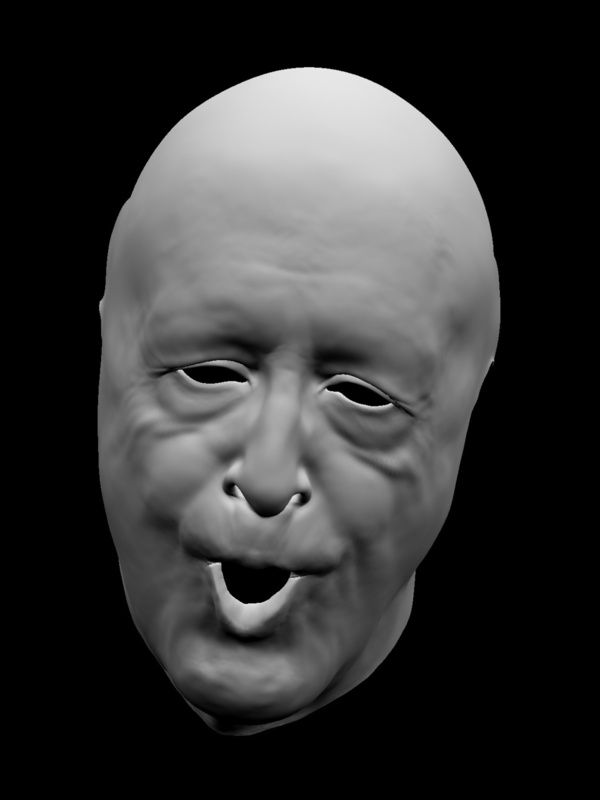}
    \makebox[10pt]{\raisebox{40pt}{\rotatebox[origin=c]{90}{1160}}}%
    \includegraphics[width=\dimexpr\linewidth-10pt\relax]{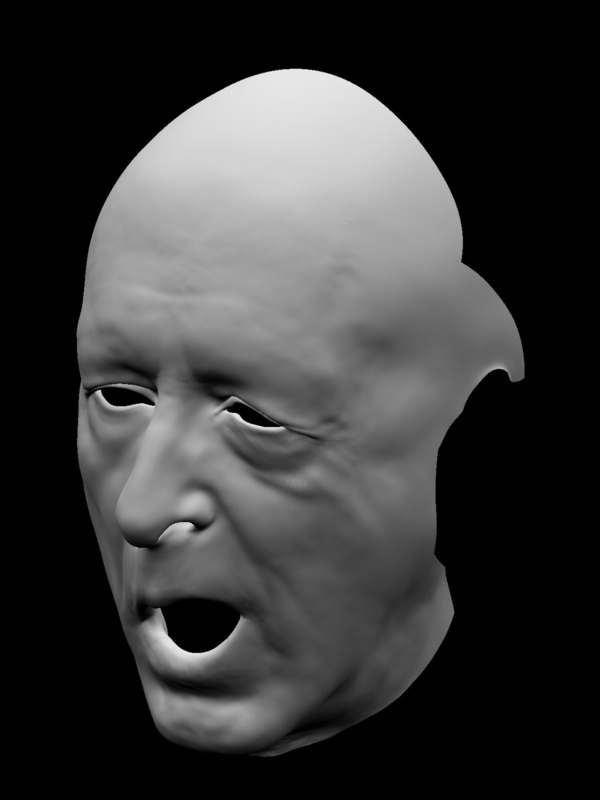}
    \makebox[10pt]{\raisebox{40pt}{\rotatebox[origin=c]{90}{1170}}}%
    \includegraphics[width=\dimexpr\linewidth-10pt\relax]{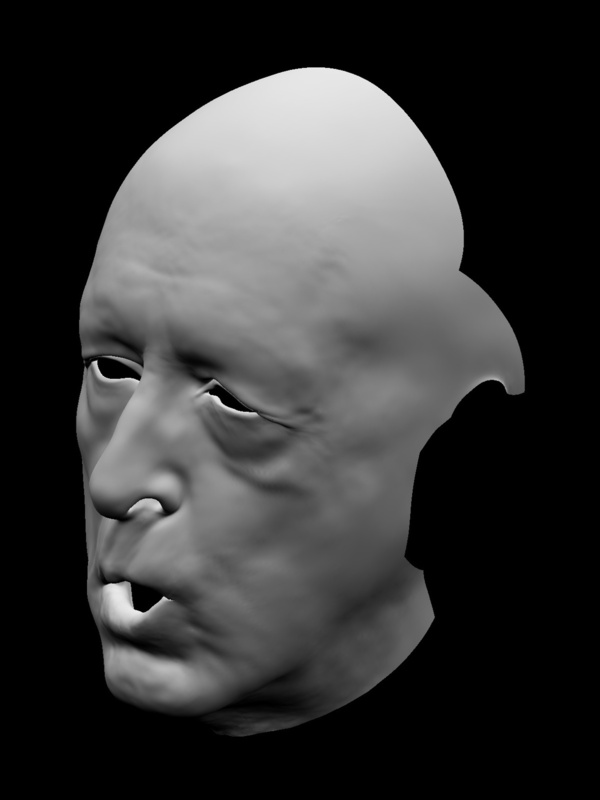}
    \caption{Blendshapes}
\end{subfigure}
\begin{subfigure}[b]{0.31\linewidth}
    \includegraphics[width=\linewidth]{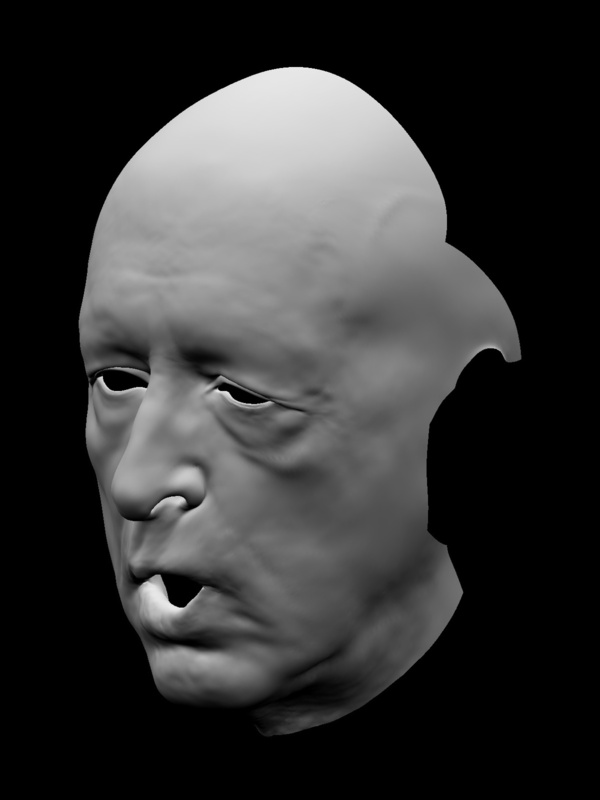}
    \includegraphics[width=\linewidth]{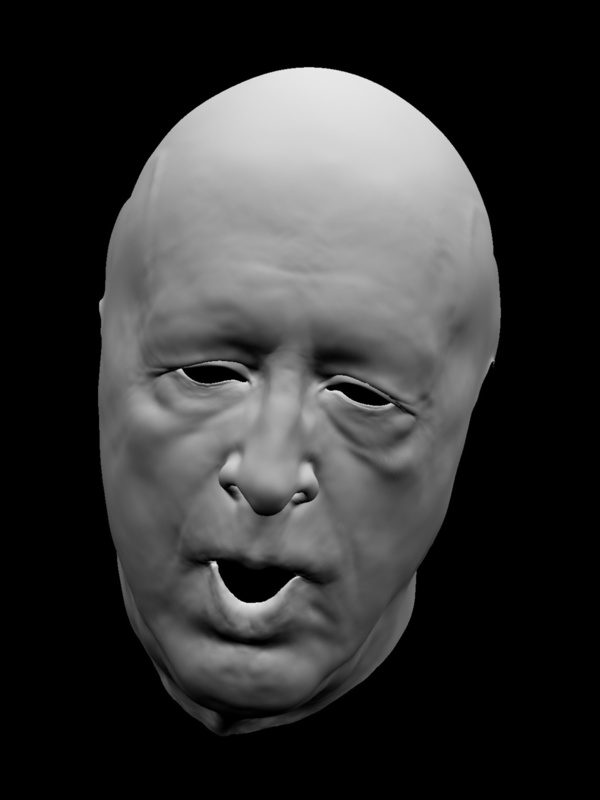}
    \includegraphics[width=\linewidth]{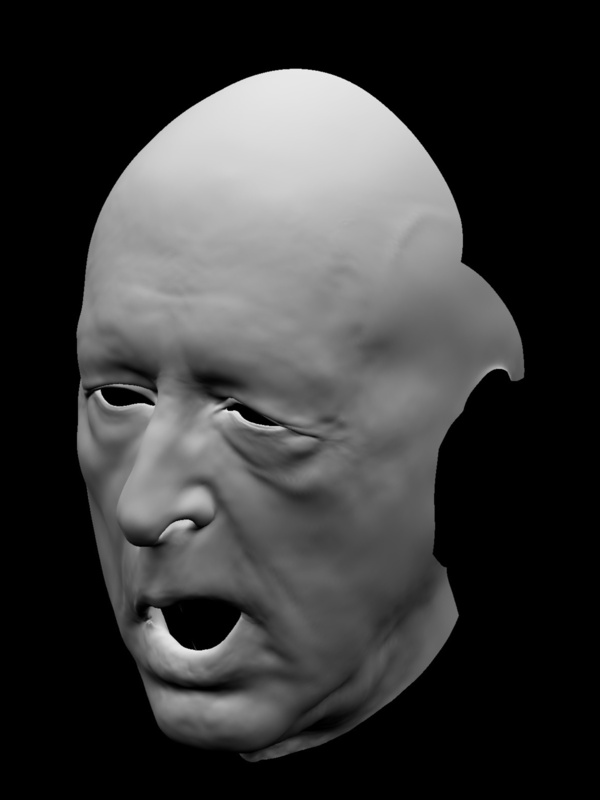}
    \includegraphics[width=\linewidth]{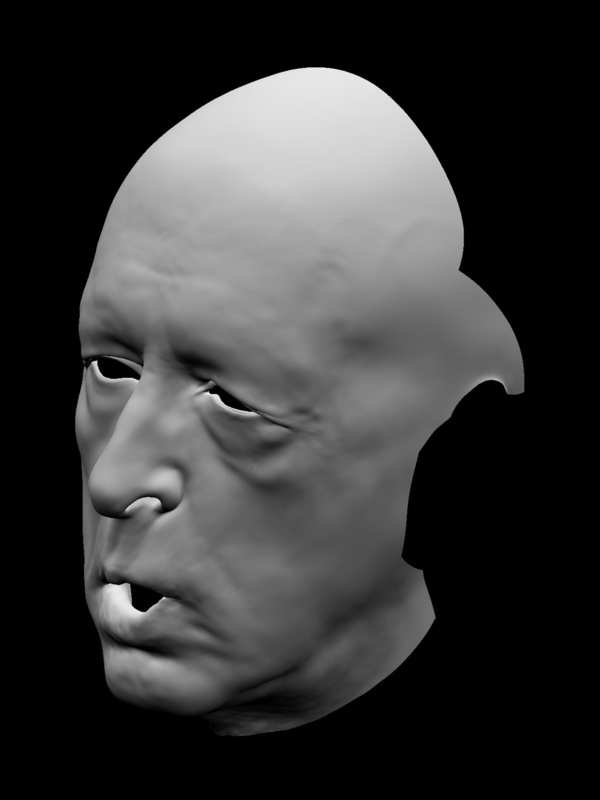}
    \caption{Simulation}
\end{subfigure}
\begin{subfigure}[b]{0.31\linewidth}
    \includegraphics[width=\linewidth]{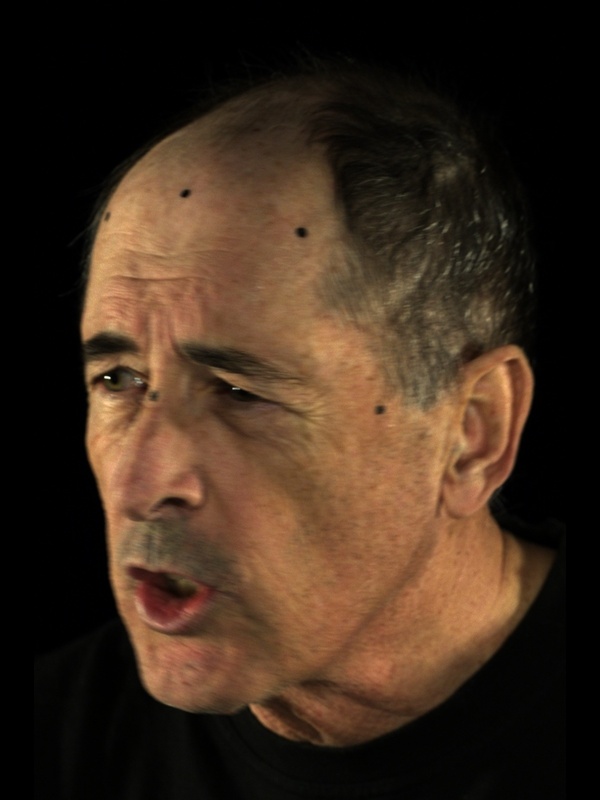}
    \includegraphics[width=\linewidth]{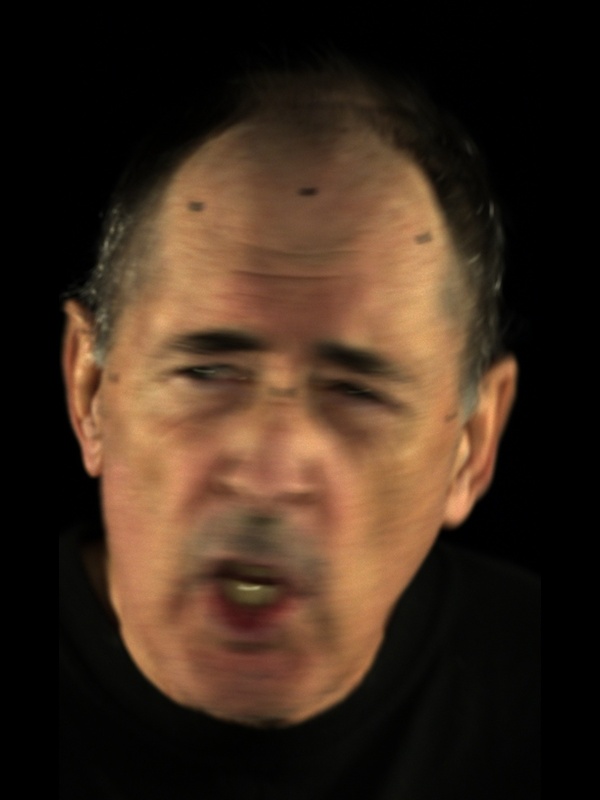}
    \includegraphics[width=\linewidth]{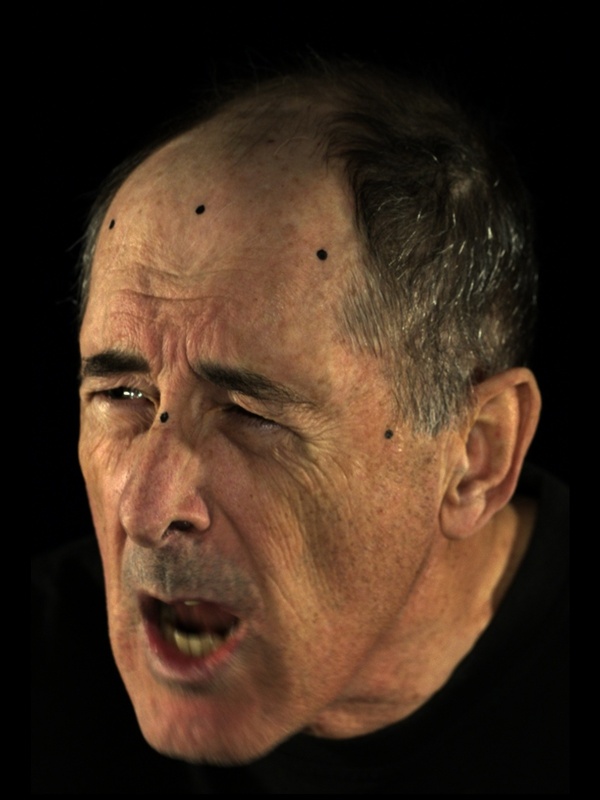}
    \includegraphics[width=\linewidth]{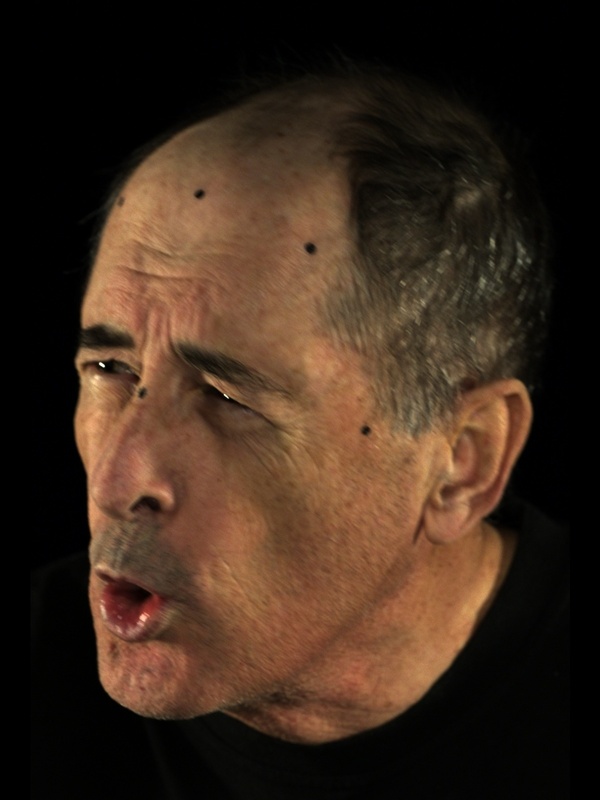}
    \caption{Plate}
\end{subfigure}
\hfill
\caption{We target the raw image data using our face model $x(w)$ using both simulation and blendshapes on a number of frames of an actor's performance.
Both sets of results suffer from some depth ambiguity due to only using monocular two-dimensional data in the optimization.}
\label{fig:result_inverse_plate}
\vspace{-2.5mm}
\end{figure}

We demonstrate the efficacy of our approach on \num{66} frames of a facial performance.
As in Section \ref{sec:target_3d_geom}, we compare the results of solving Equations \ref{eq:roto_nllq} and \ref{eq:deformation_nllq} using $x(w)$ driven by a simulation model versus a blendshape model.
In particular, we choose four frames with particularly challenging facial expressions (frames \num{1112}, \num{1160}, \num{1170}) as well as capture conditions such as motion blur (frame \num{1134}).
We note that a significant portion of the facial expression is captured using the rotoscoped curves and the shape-from-shading step primarily helps to refine the expression and the contours of the face.
Both $E_1$ and $E_2$ (Equations \ref{eq:roto_nllq} and \ref{eq:deformation_nllq}) require end-to-end differentiability through our blendshape driven method.
See Figure \ref{fig:result_inverse_plate}.
While the general expressions are similar, we note that the simulation's surface geometry tends to be more physically plausible due the simulation's ability to preserve volume, especially around the lips.
This regularization is especially prominent on frame \num{1134}.
As shown in supplementary material, the resulting muscle activation values are also comparatively sparser which leads to an increased ability to extract semantic meaning out of the performance.
Additional comparisons and results are shown in the supplementary material and video.

\section{Conclusion and Future Work} \label{sec:conclusion}

Although promising anatomically based muscle simulation systems have existed for some time and have had the ability to target data as in \cite{sifakis2005automatic}, they have lacked the high-end efficacy required to produce compelling results.
Although the recently proposed \cite{cong2016art} does produce quite compelling results, it requires a full face shape as input and is not differentiable.
In this paper, we alleviated both of the aforementioned difficulties, extending \cite{cong2016art} with end-to-end differentiability and a morphing system driven by blendshape parameters.
This blendshape-driven morph removes the need for a full face surface mesh as a pre-existing target.
We demonstrate the efficacy of our approach by targeting three-dimensional geometry and two-dimensional RGB images.
To the best of our knowledge, we are the first to use quasistatic simulation of a muscle model to target RGB images.
We note that methods such as \cite{sifakis2005automatic} could be used in the optimizations presented in this paper (as outlined in the second to last paragraph of Section \ref{sec:differentiable_system}); however, the resulting simulation results would be less expressive and would not be able to effectively reproduce the desired expressions.

Although the computer vision community expends great efforts in regards to identifying faces in images, segmenting them cleanly from their surroundings, and even identifying their shape, semantic understanding of what such faces are doing or intend to do or feel is still in its infancy consisting mostly of preliminary image labeing and annotation.
The ability to express a facial pose or image using a muscle activation basis provides an anatomically-motivated way to extract semantic information.
Even without extensive model calibration, our anatomical model's muscle activations have shown to be useful for extracting anatomically-based smenatic information.
This is a promising avenue for future work.
Additionally, muscle activations could also be used as a basis for statistical/deep learning instead of semantically meaningless combinations of blendshape weights.

Finally, one of the more philosophical questions in deep learning seems to revolve around what should or should not be considered a ``learning crime'' (drawing similarities to variational crimes \cite{strang1972variational}).
For example, in \cite{bailey2018fast}, the authors learn a perturbation of linear blend skinning as opposed to the whole shape, assuming that the perturbation is lower-dimensional, spatially correlated, and/or easier to learn.
The authors in \cite{feng2018joint,ranjan2018generating} use spatially correlated networks for spatially correlated information under the assumption, once again, that this leads to a network that is easier to train and generalizes better.
It seems that adding strong priors, domain knowledge, informed procedural methods, etc.~to generate as much of a function as possible before training a network to learn the rest is often considered prudent.
Our anatomically-based physical simulation system incorporates physical properties such as volume preservation, contact, and collision so that a network would not need to learn or explain them; instead the network only needs to learn what further perturbations are required to match the data.

\section*{Acknowledgements}
Research supported in part by ONR N00014-13-1-0346, ONR N00014-17-1-2174, ARL AHPCRC W911NF-07-0027, and generous gifts from Amazon and Toyota.
In addition, we would like to thank both Reza and Behzad at ONR for supporting our efforts into computer vision and machine learning, as well as Cary Phillips, Kiran Bhat, and Industrial Light \& Magic for supporting our efforts into facial performance capture.
M. Bao was supported in part by The VMWare Fellowship in Honor of Ole Agesen.
We would also like to thank Paul Huston for his acting and Jane Wu for her help in preparing the supplementary video.

\section*{Appendices}
\appendix
\section{Targeting 3D Geometry - Additional Results}

\begin{figure*}[t]
\centering
\begin{subfigure}[b]{\dimexpr0.18\linewidth+10pt\relax}
    \makebox[10pt]{\raisebox{55pt}{\rotatebox[origin=c]{90}{Blendshapes}}}%
    \includegraphics[width=\dimexpr\linewidth-10pt\relax]{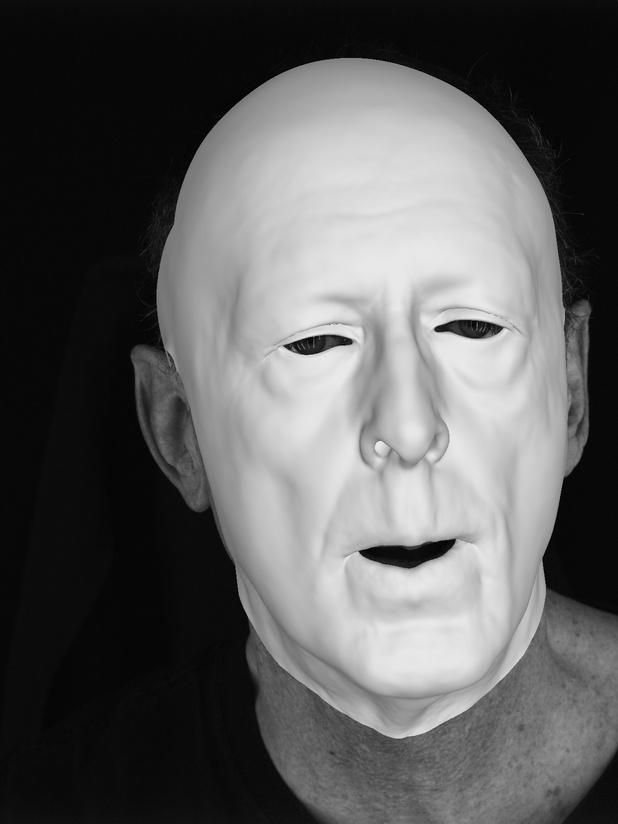}
    \makebox[10pt]{\raisebox{55pt}{\rotatebox[origin=c]{90}{Simulation}}}%
    \includegraphics[width=\dimexpr\linewidth-10pt\relax]{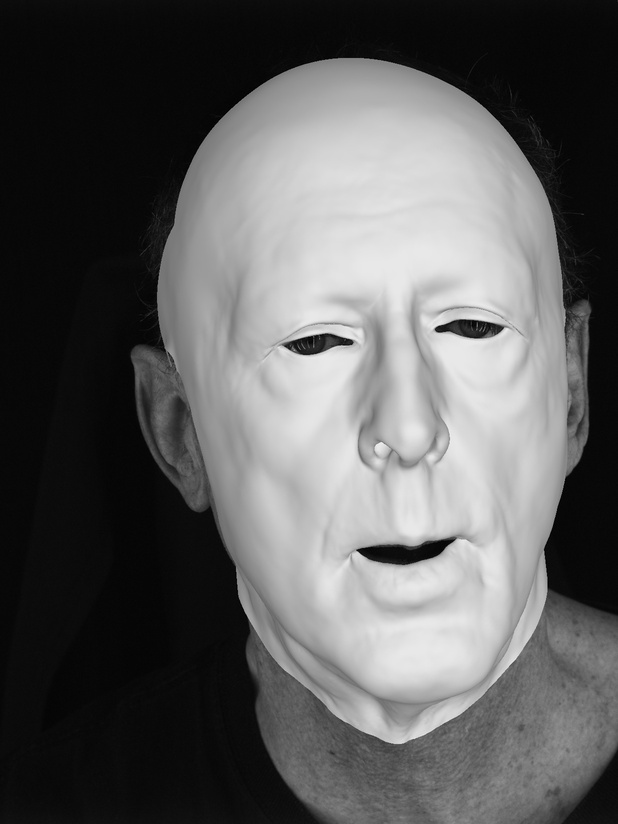}
    \makebox[10pt]{\raisebox{55pt}{\rotatebox[origin=c]{90}{Target}}}%
    \includegraphics[width=\dimexpr\linewidth-10pt\relax]{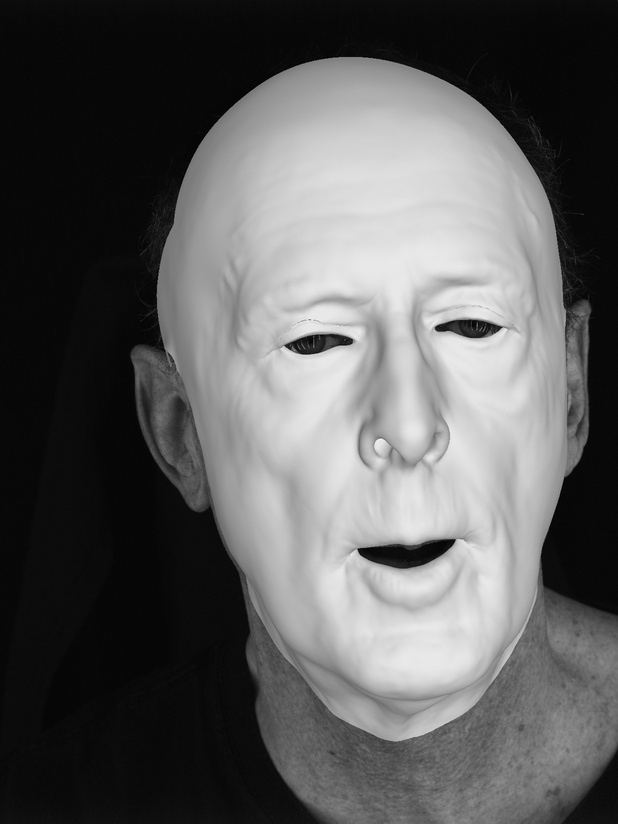}
    \caption{2536}
\end{subfigure}
\begin{subfigure}[b]{0.18\linewidth}
    \includegraphics[width=\linewidth]{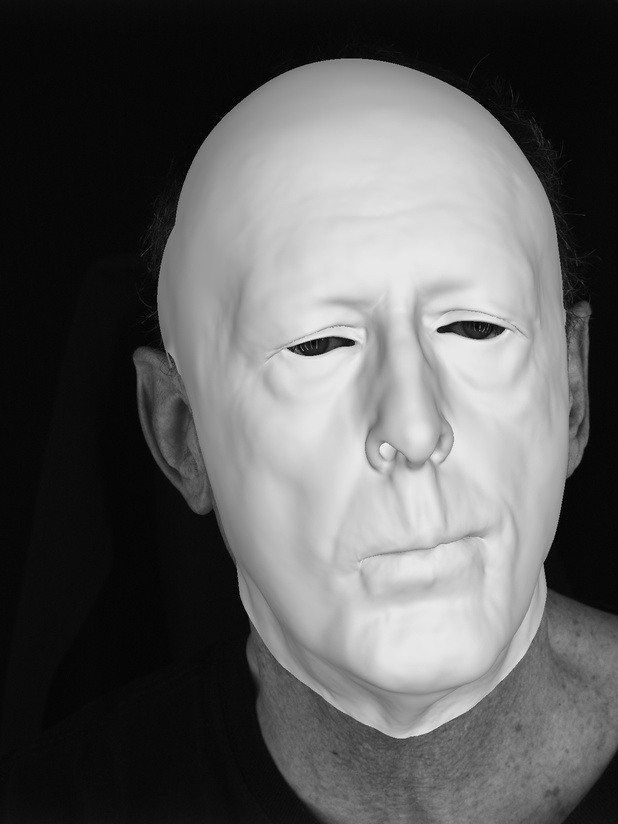}
    \includegraphics[width=\linewidth]{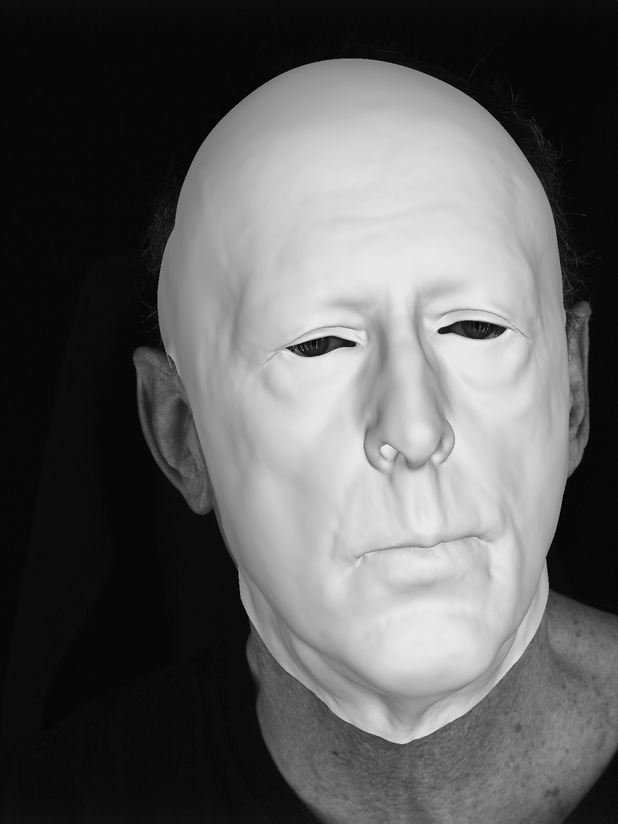}
    \includegraphics[width=\linewidth]{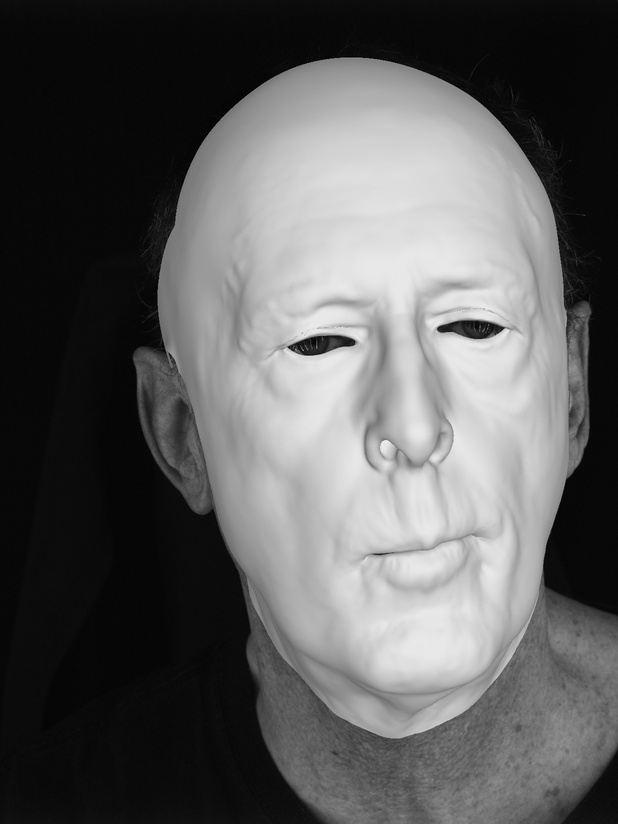}
    \caption{2540}
\end{subfigure}
\begin{subfigure}[b]{0.18\linewidth}
    \includegraphics[width=\linewidth]{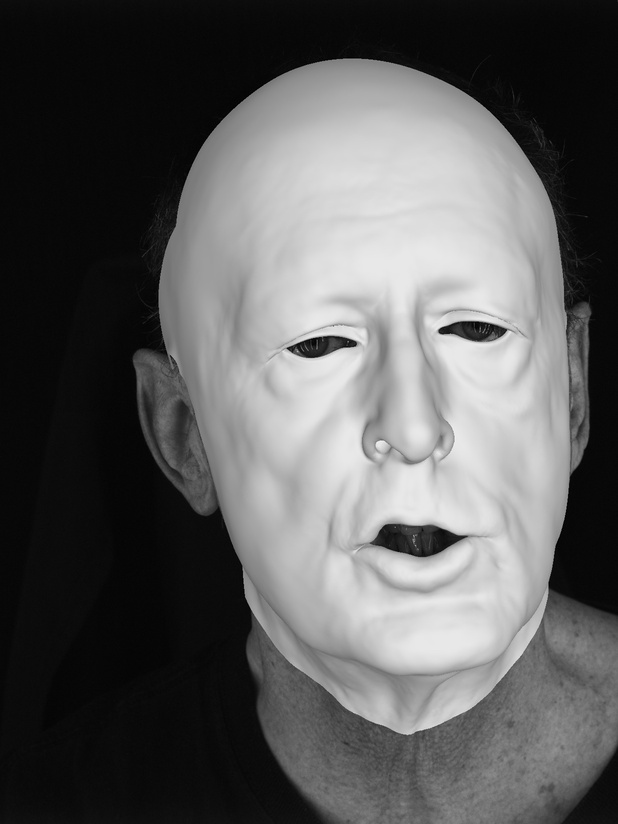}
    \includegraphics[width=\linewidth]{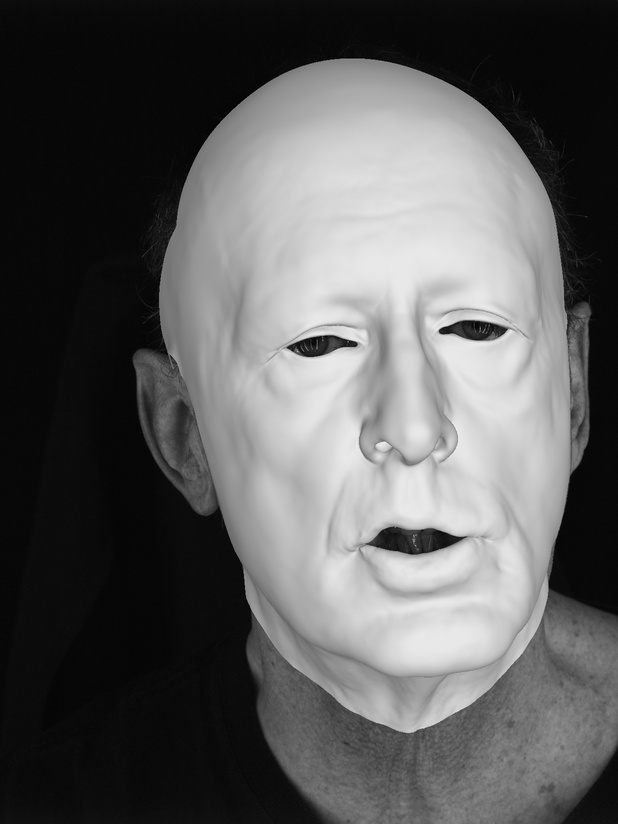}
    \includegraphics[width=\linewidth]{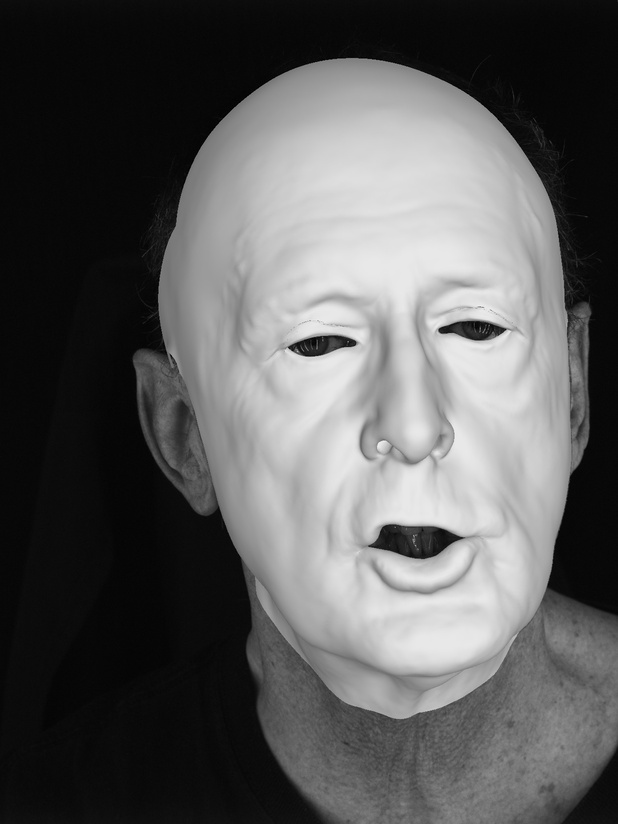}
    \caption{2560}
\end{subfigure}
\begin{subfigure}[b]{0.18\linewidth}
    \includegraphics[width=\linewidth]{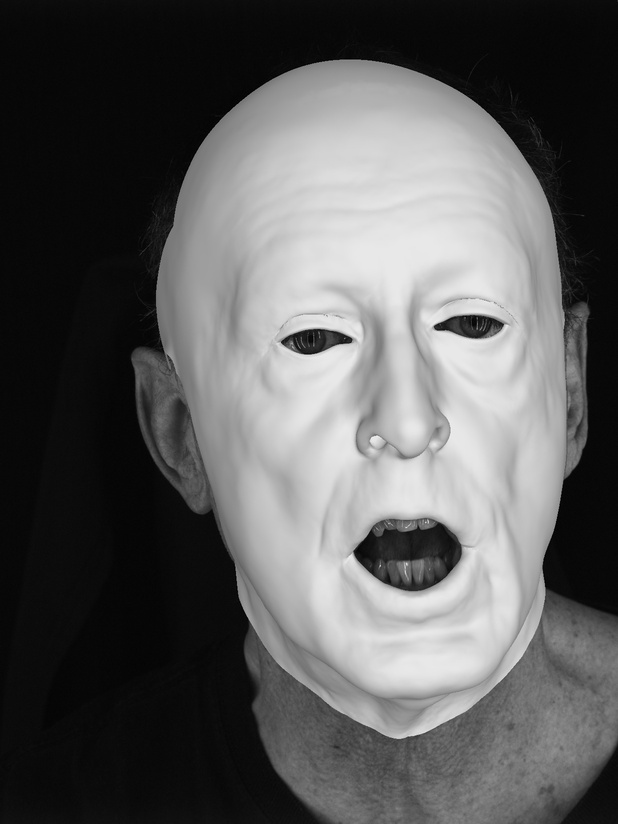}
    \includegraphics[width=\linewidth]{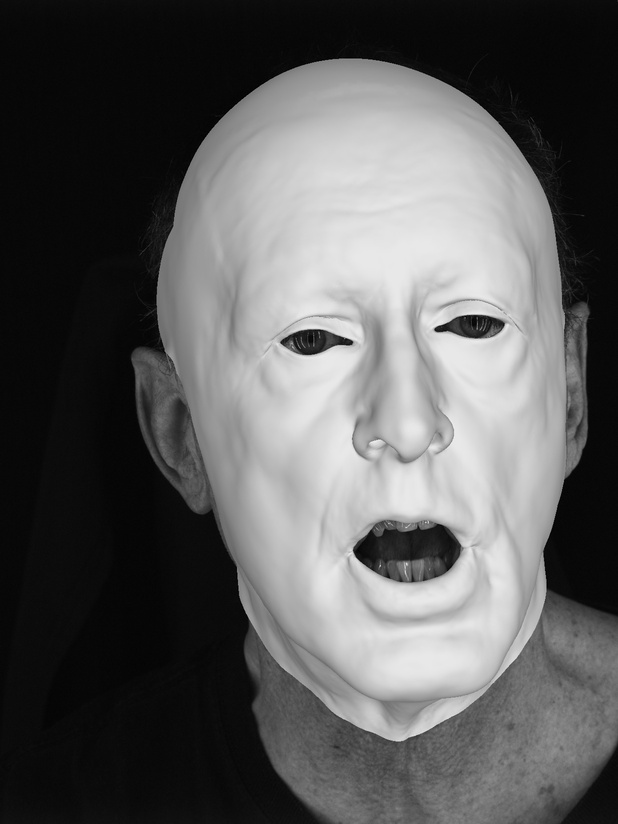}
    \includegraphics[width=\linewidth]{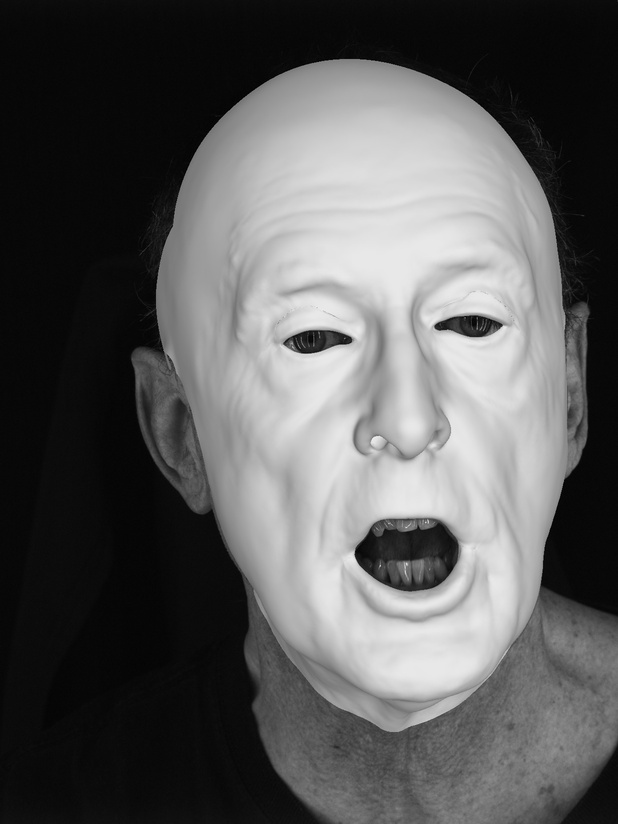}
    \caption{2573}
\end{subfigure}
\begin{subfigure}[b]{0.18\linewidth}
    \includegraphics[width=\linewidth]{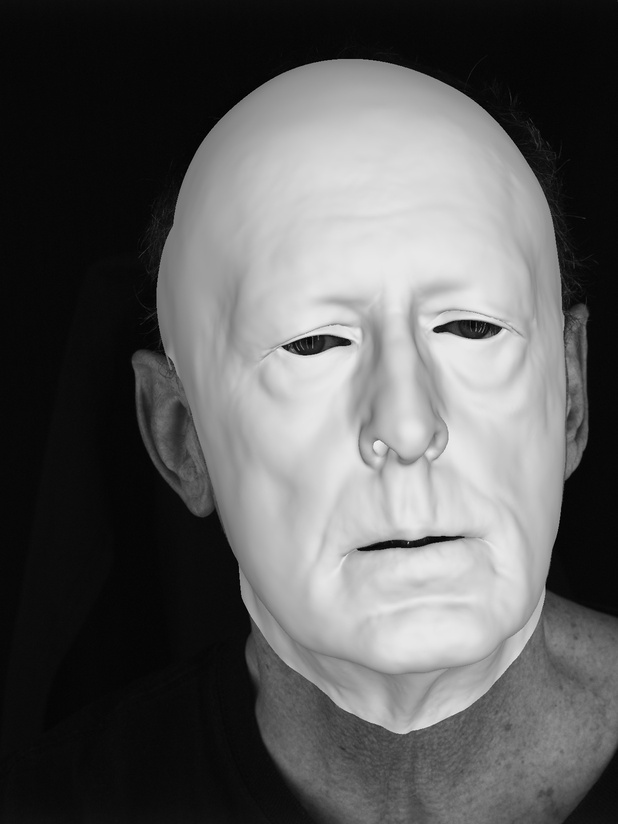}
    \includegraphics[width=\linewidth]{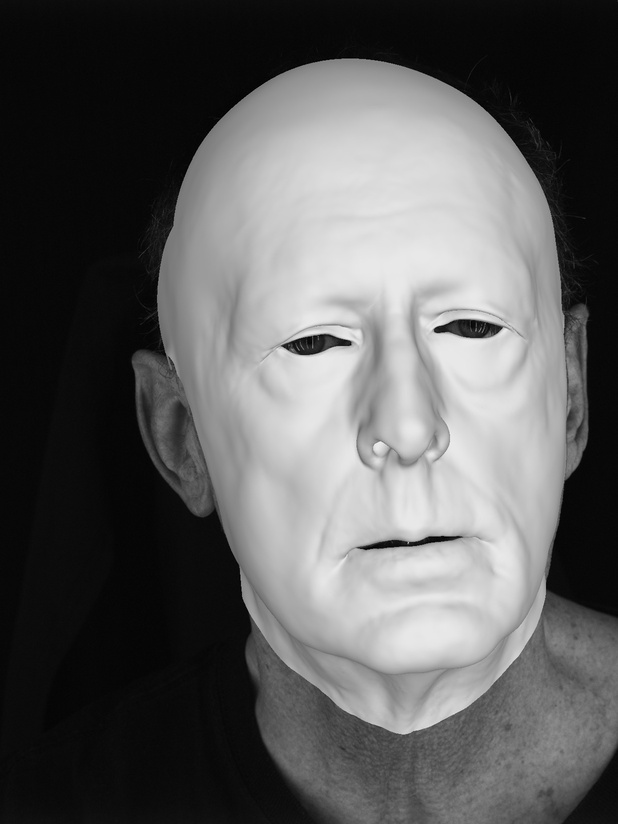}
    \includegraphics[width=\linewidth]{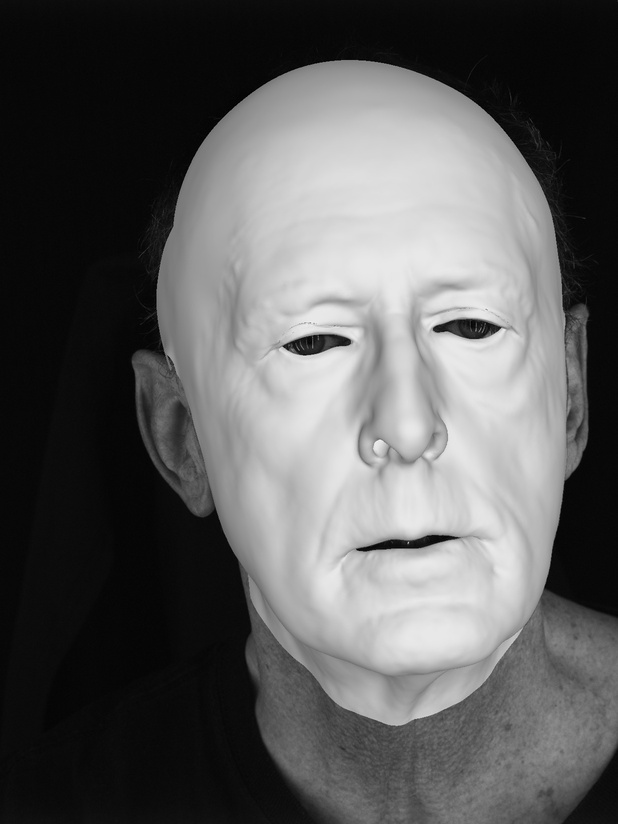}
    \caption{2590}
\end{subfigure}
\hfill
\caption{Additional comparisons when targeting geometry viewed from one of the original camera viewpoints.}
\label{fig:medusa_geometry_appendix}
\end{figure*}

\begin{figure*}[t]
\centering
\begin{subfigure}[b]{\dimexpr0.18\linewidth+10pt\relax}
    \makebox[10pt]{\raisebox{30pt}{\rotatebox[origin=c]{90}{Blendshapes}}}%
    \includegraphics[width=\dimexpr\linewidth-10pt\relax]{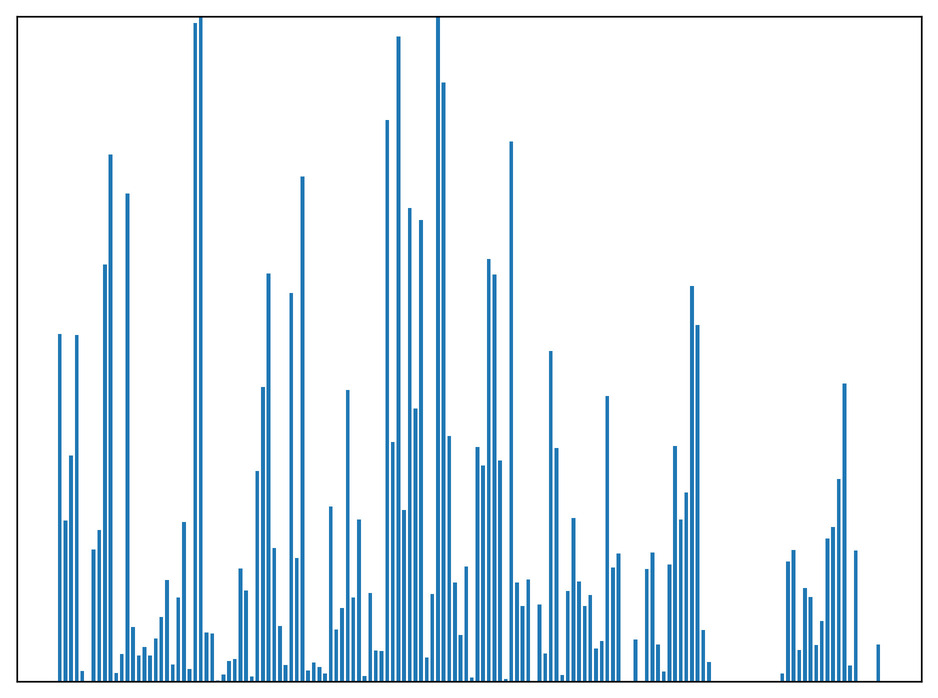}
    \makebox[10pt]{\raisebox{30pt}{\rotatebox[origin=c]{90}{Simulation}}}%
    \includegraphics[width=\dimexpr\linewidth-10pt\relax]{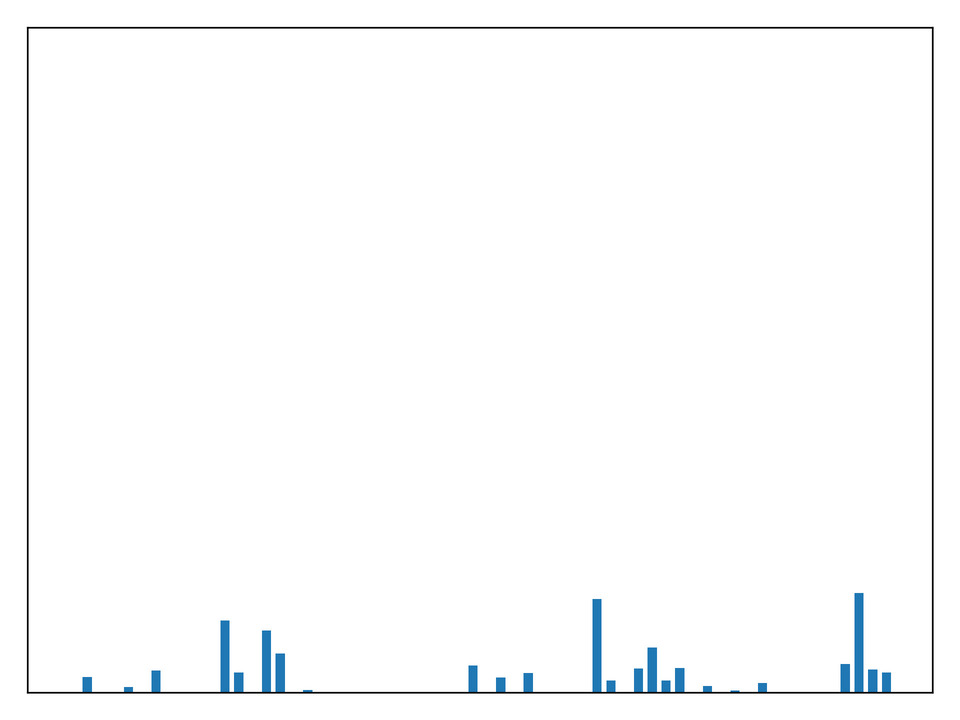}
    \caption{2536}
\end{subfigure}
\begin{subfigure}[b]{0.18\linewidth}
    \includegraphics[width=\linewidth]{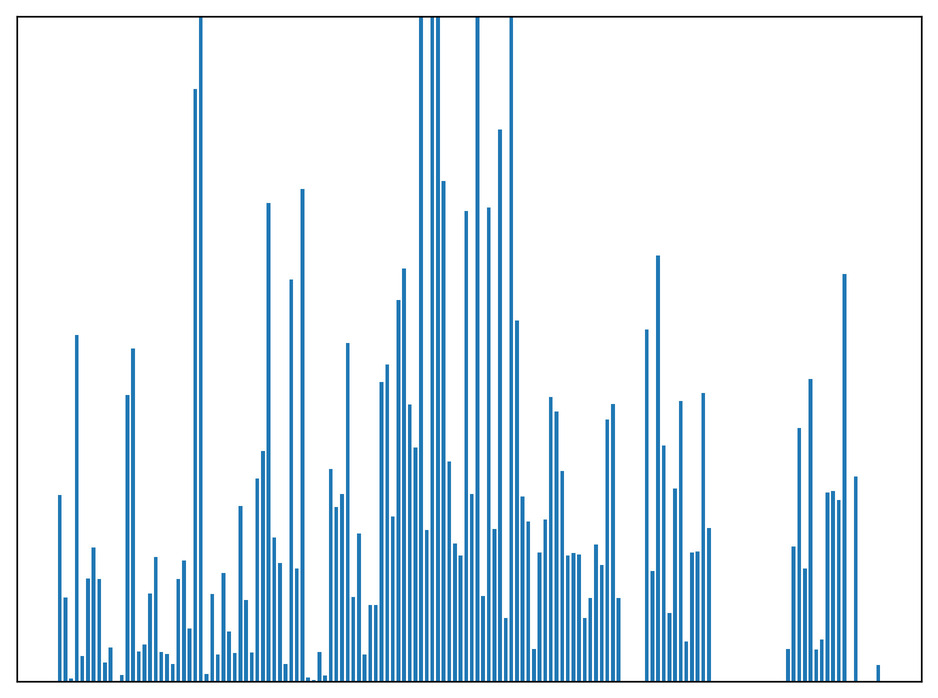}
    \includegraphics[width=\linewidth]{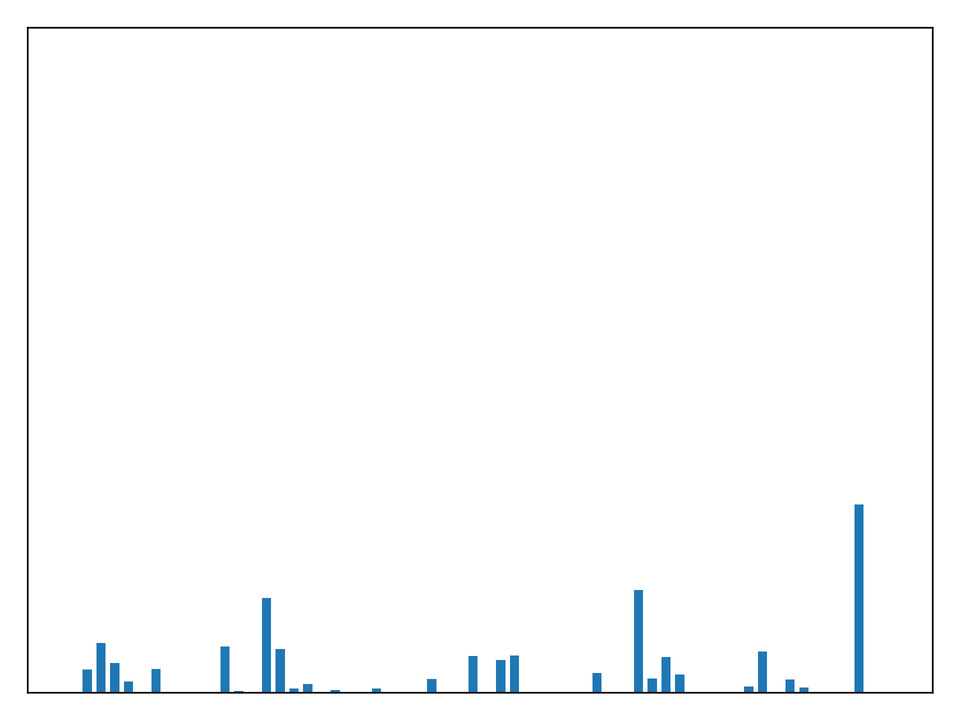}
    \caption{2540}
\end{subfigure}
\begin{subfigure}[b]{0.18\linewidth}
    \includegraphics[width=\linewidth]{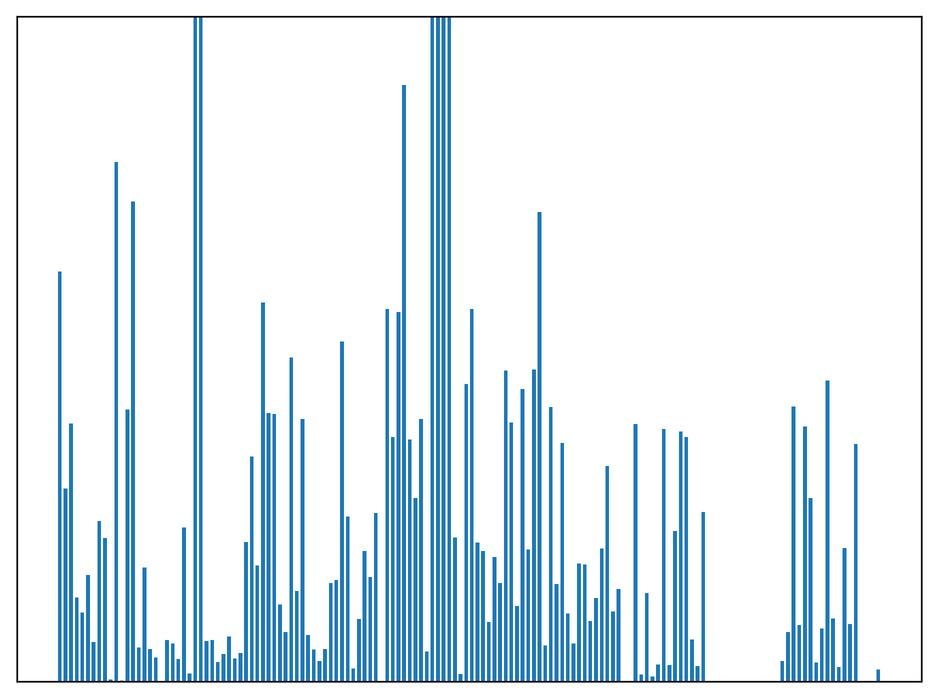}
    \includegraphics[width=\linewidth]{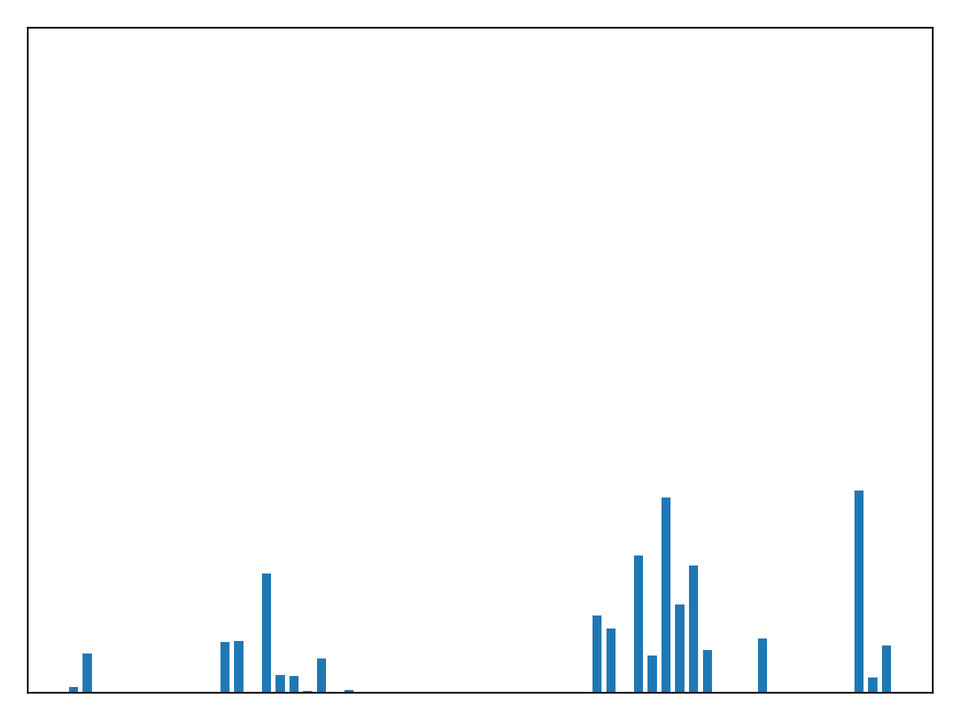}
    \caption{2560}
\end{subfigure}
\begin{subfigure}[b]{0.18\linewidth}
    \includegraphics[width=\linewidth]{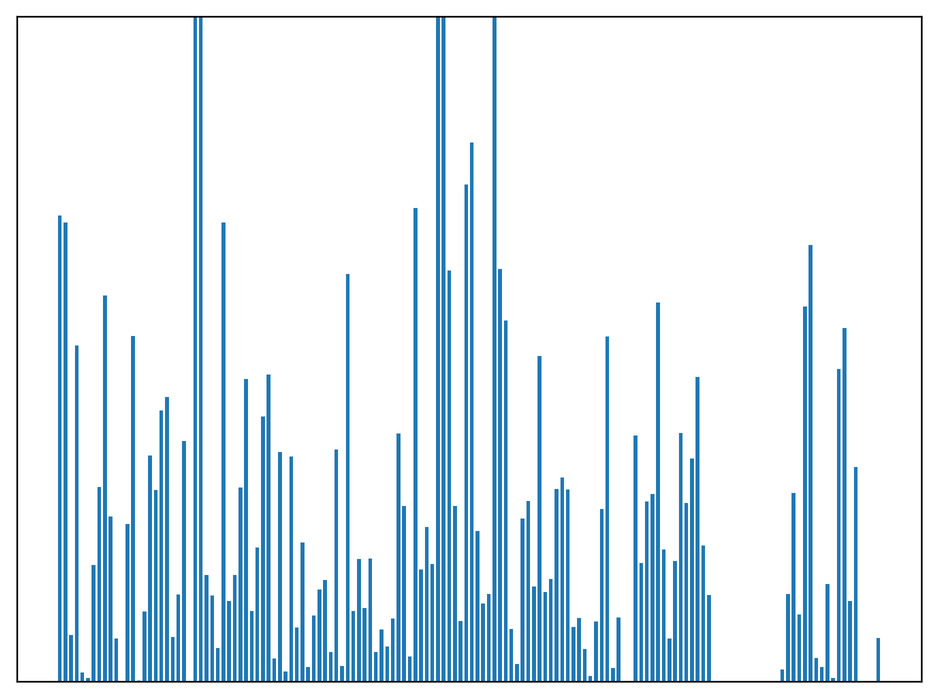}
    \includegraphics[width=\linewidth]{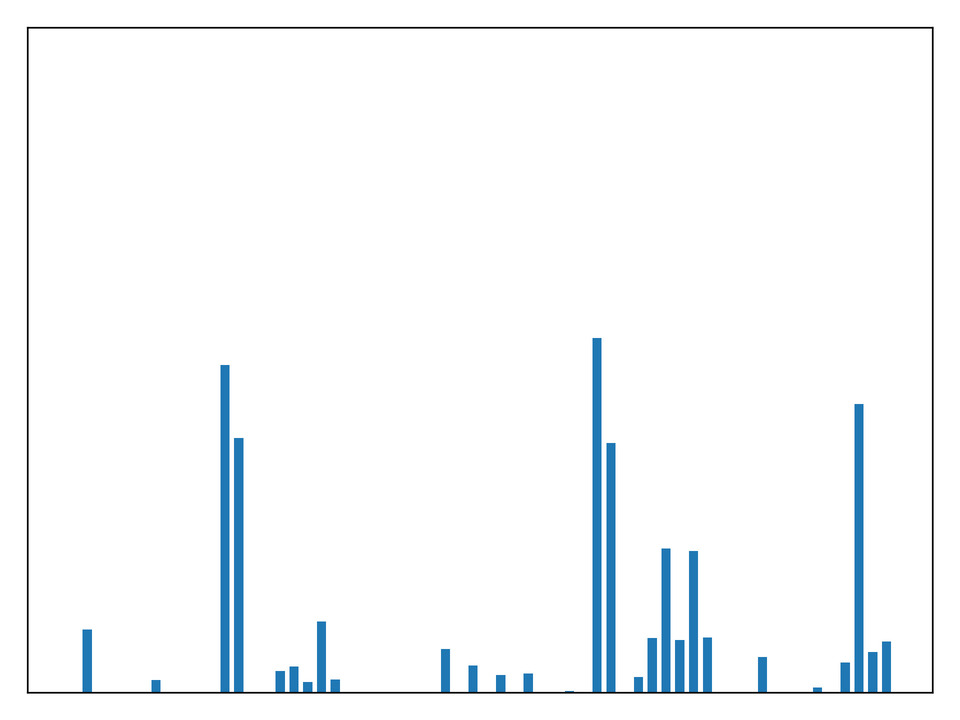}
    \caption{2573}
\end{subfigure}
\begin{subfigure}[b]{0.18\linewidth}
    \includegraphics[width=\linewidth]{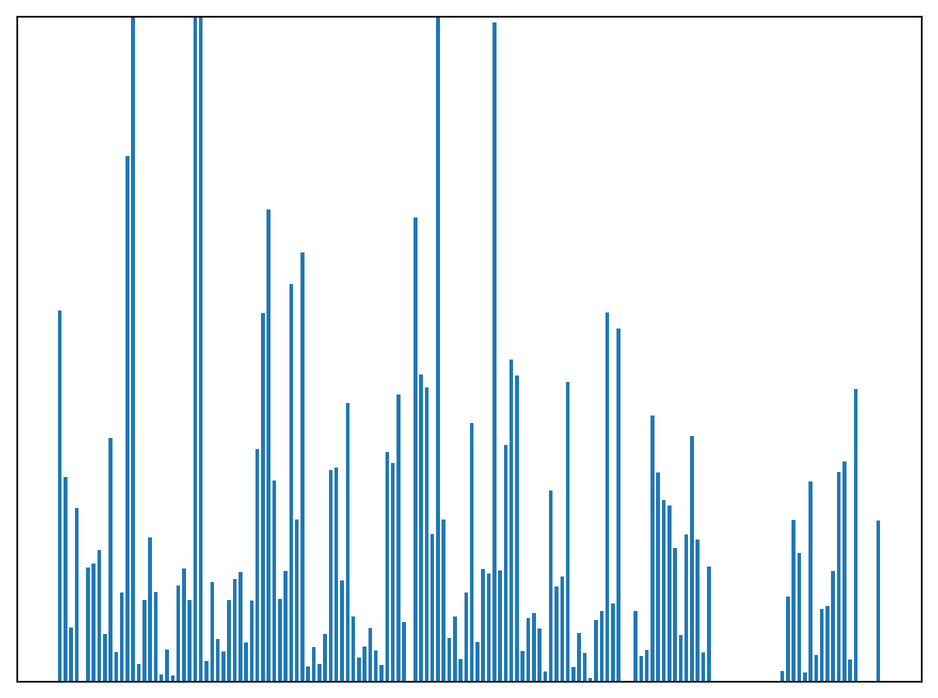}
    \includegraphics[width=\linewidth]{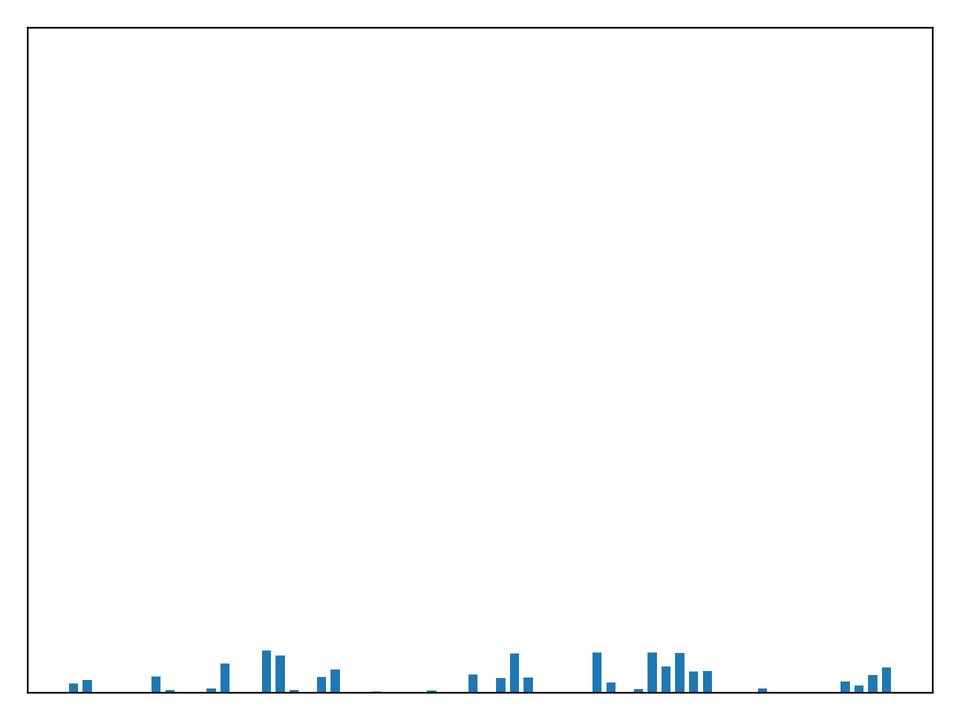}
    \caption{2590}
\end{subfigure}
\hfill
\caption{Additional comparisons between the resulting blendshape weights and muscle activations when targeting geometry.}
\label{fig:medusa_weights_appendix}
\end{figure*}

\begin{figure*}[t]
\centering
\begin{subfigure}[b]{\dimexpr0.18\linewidth}
    \includegraphics[width=\linewidth]{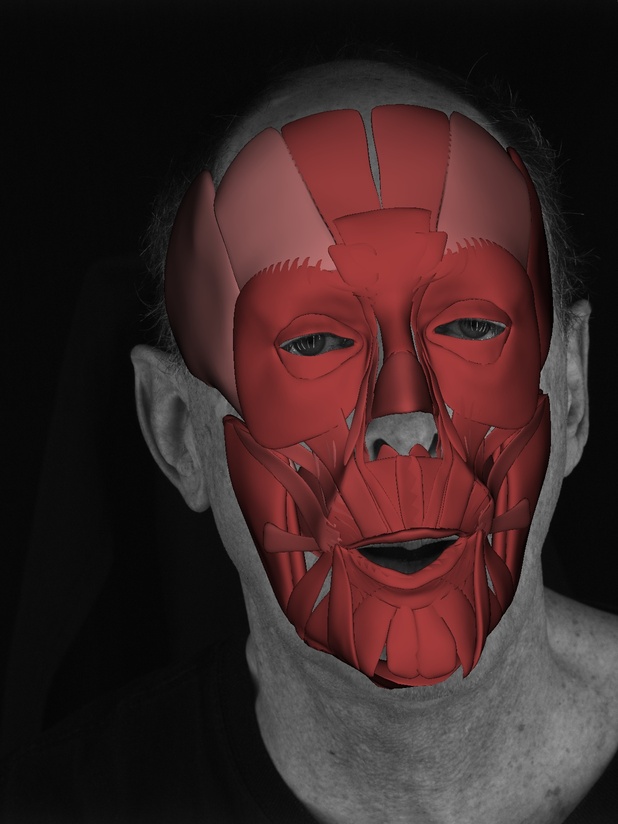}
    \caption{2536}
\end{subfigure}
\begin{subfigure}[b]{0.18\linewidth}
    \includegraphics[width=\linewidth]{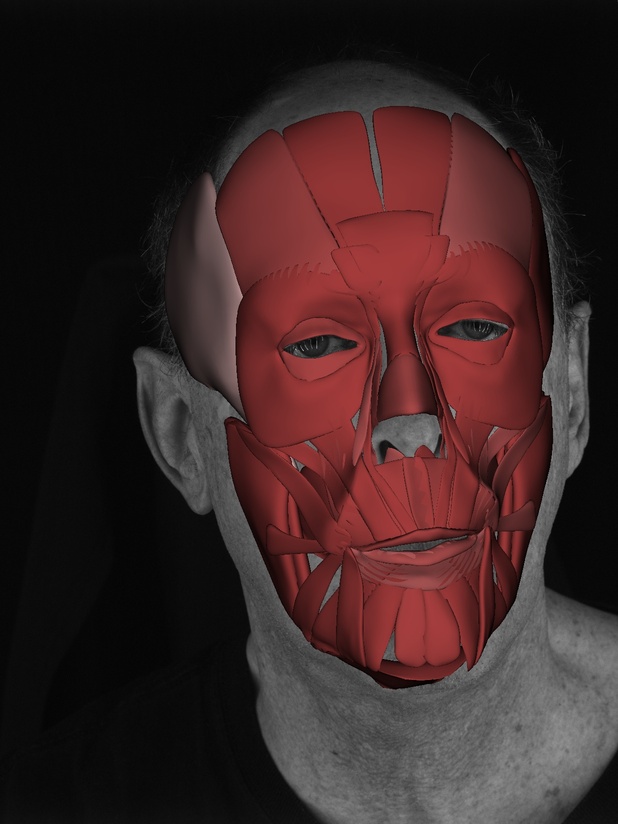}
    \caption{2540}
\end{subfigure}
\begin{subfigure}[b]{0.18\linewidth}
    \includegraphics[width=\linewidth]{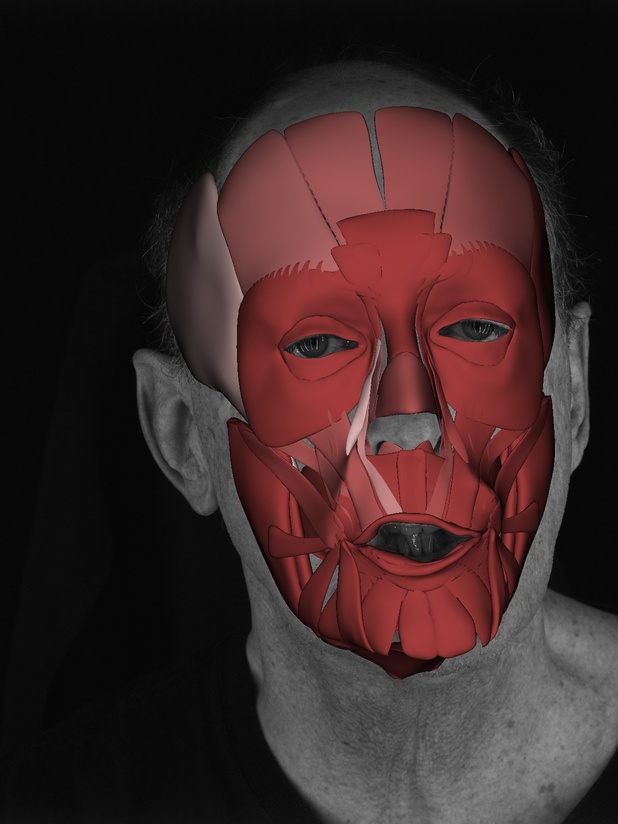}
    \caption{2560}
\end{subfigure}
\begin{subfigure}[b]{0.18\linewidth}
    \includegraphics[width=\linewidth]{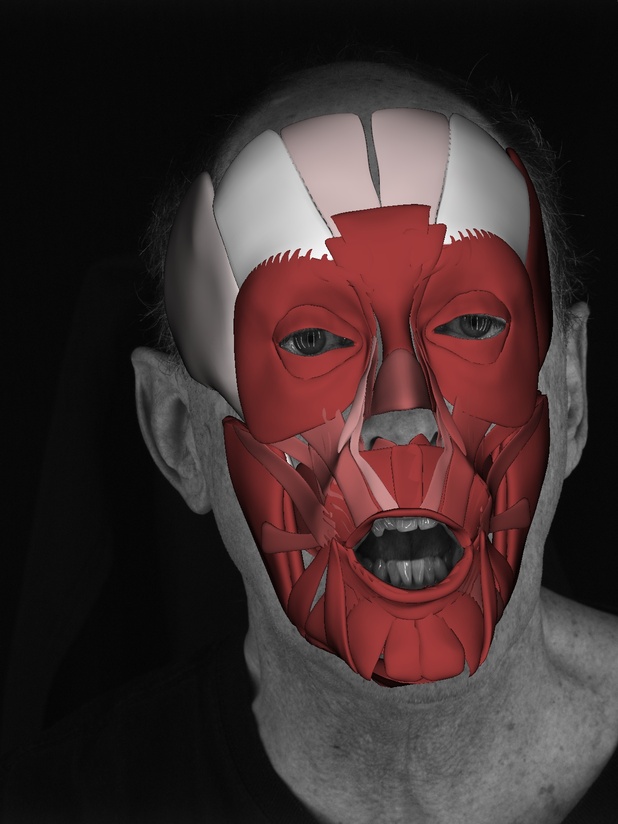}
    \caption{2573}
\end{subfigure}
\begin{subfigure}[b]{0.18\linewidth}
    \includegraphics[width=\linewidth]{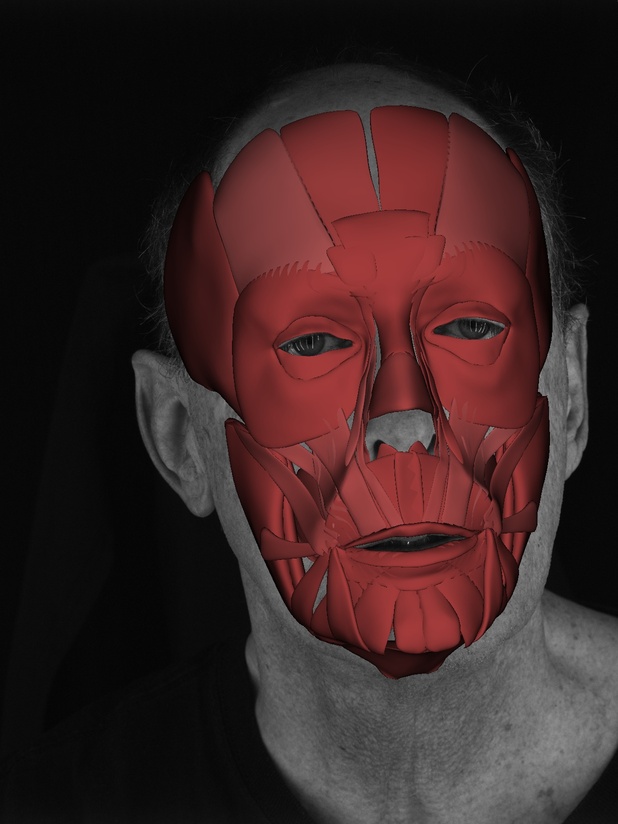}
    \caption{2590}
\end{subfigure}
\hfill
\caption{Muscle activations from Figure \ref{fig:medusa_weights_appendix} visualized where activations greater than \num{0.5} are colored white and activations at \num{0} are colored red.}
\label{fig:medusa_weights_geometry}
\end{figure*}

We present additional comparisons between using blendshapes and simulations for targeting three-dimensional geometry in Figure \ref{fig:medusa_geometry_appendix}. 
Our approach using muscle simulation results in facial expressions similar to that obtained via blendshapes, but also introduces physical properties such as volume preservation.
Our results can be improved by further calibrating and refining the anatomical model.
As seen in Figure \ref{fig:medusa_weights_appendix}, the resulting muscle activation weights are sparser and less overdialed than their blendshape counterparts.
In particular, note how the muscle activations generally track the magnitude of the expression.
This is especially evident in frame \num{2590} where the face is in a close to neutral pose; while the muscle activations are close to all $0$, the blendshape weights are still dialed in heavily to match the expression.
The overdialing of blendshape weights could be alleviated by increasing the L2 regularization of the weights; however, this will also cause the captured performance to become less representative of the original performance.
Figure \ref{fig:medusa_weights_geometry} shows that muscle activations result in anatomically and semantically meaningful information.
Note that further calibration of the anatomical model will also lead to more accurate muscle activation weights.

\section{Targeting RGB Images - Additional Results}

\begin{figure}[t]
\centering
\begin{subfigure}[b]{\dimexpr0.31\linewidth+10pt\relax}
    \makebox[10pt]{\raisebox{40pt}{\rotatebox[origin=c]{90}{1115}}}%
    \includegraphics[width=\dimexpr\linewidth-10pt\relax]{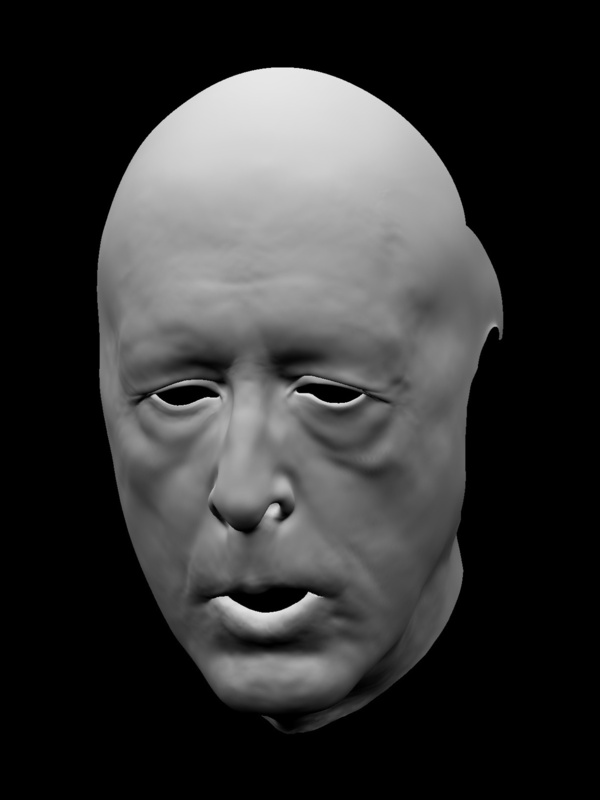}
    \makebox[10pt]{\raisebox{40pt}{\rotatebox[origin=c]{90}{1120}}}%
    \includegraphics[width=\dimexpr\linewidth-10pt\relax]{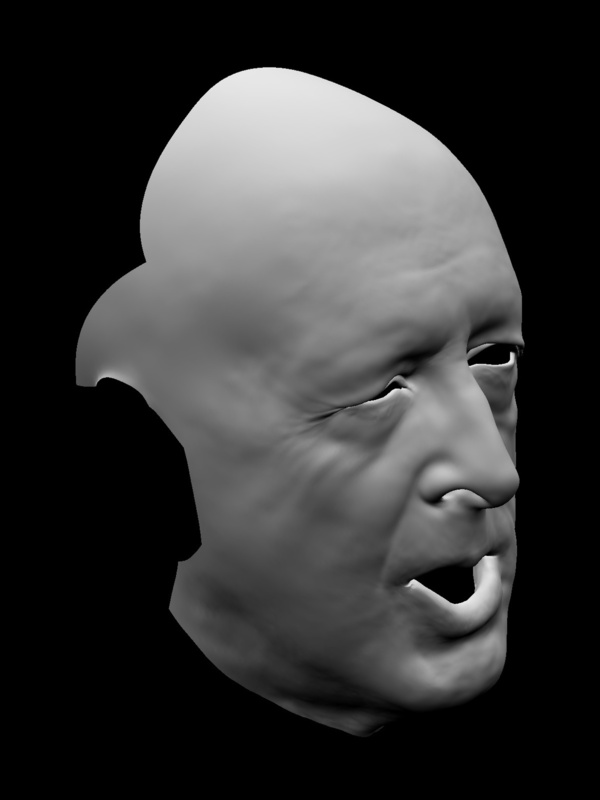}
    \makebox[10pt]{\raisebox{40pt}{\rotatebox[origin=c]{90}{1130}}}%
    \includegraphics[width=\dimexpr\linewidth-10pt\relax]{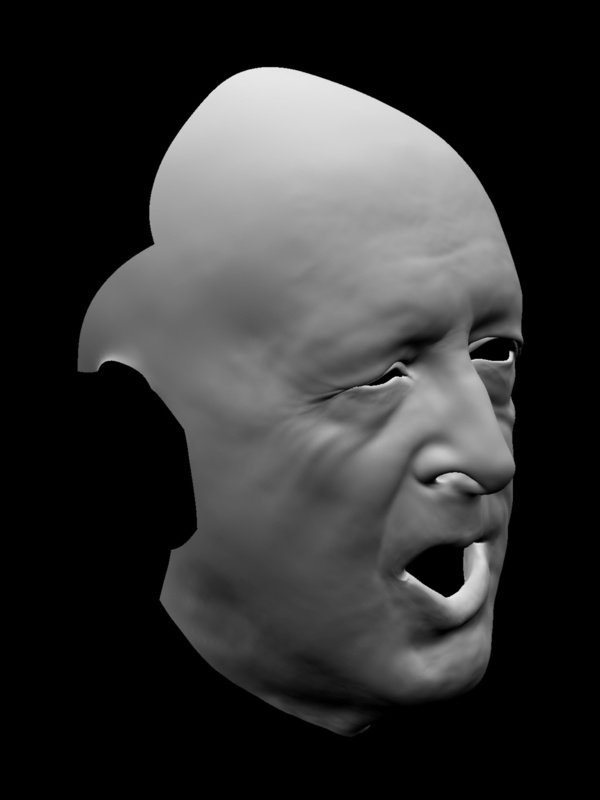}
    \makebox[10pt]{\raisebox{40pt}{\rotatebox[origin=c]{90}{1155}}}%
    \includegraphics[width=\dimexpr\linewidth-10pt\relax]{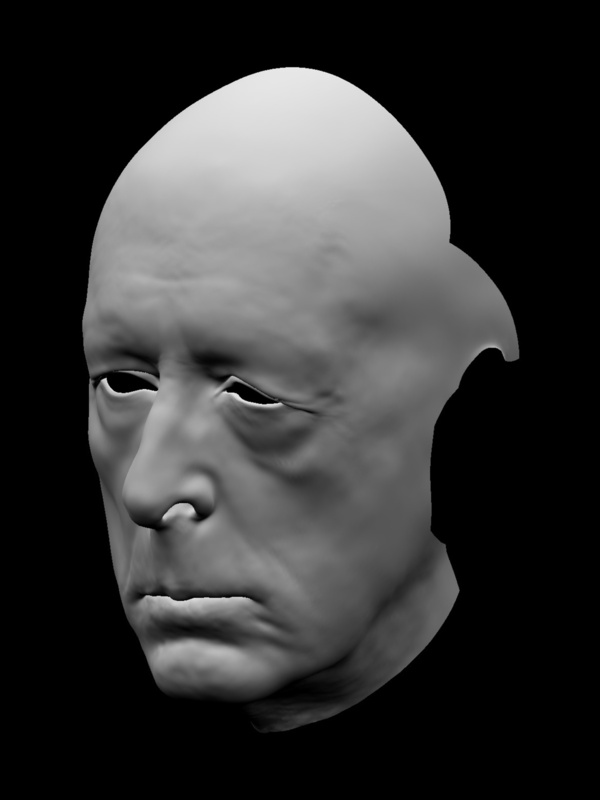}
    \caption{Blendshapes}
\end{subfigure}
\begin{subfigure}[b]{0.31\linewidth}
    \includegraphics[width=\linewidth]{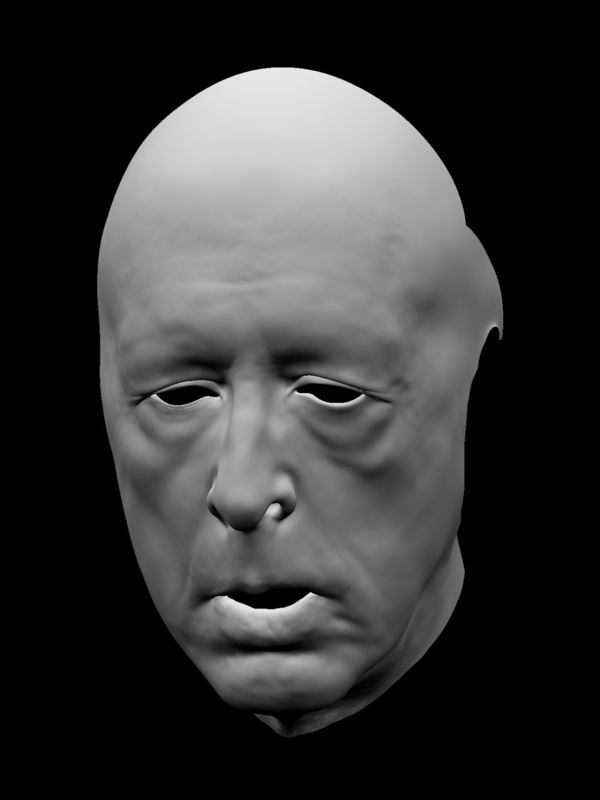}
    \includegraphics[width=\linewidth]{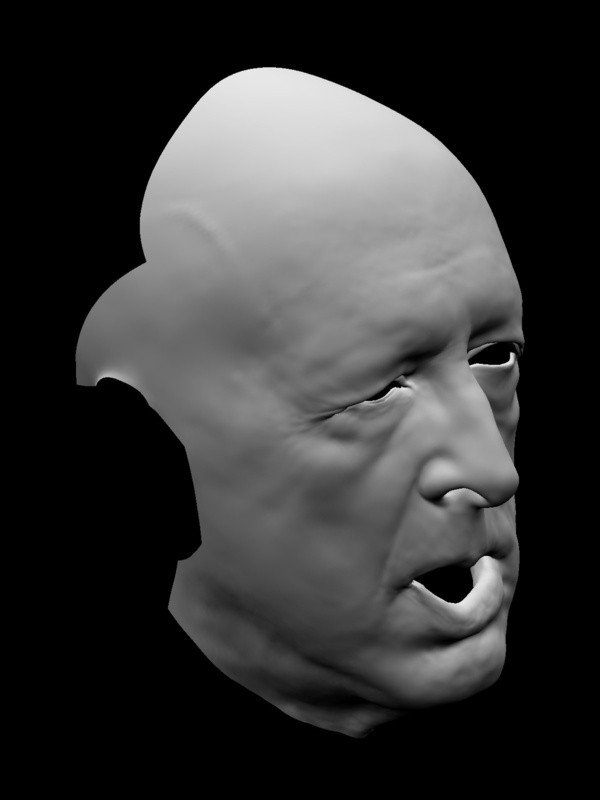}
    \includegraphics[width=\linewidth]{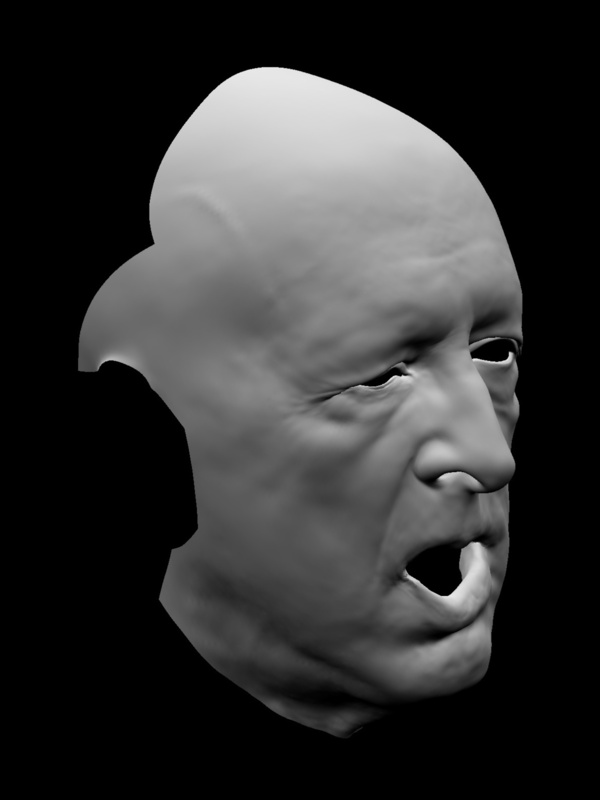}
    \includegraphics[width=\linewidth]{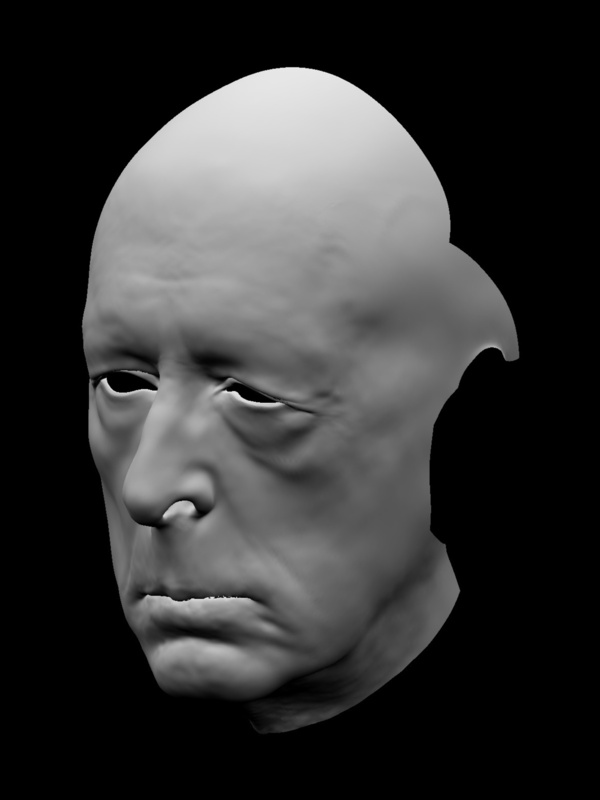}
    \caption{Simulation}
\end{subfigure}
\begin{subfigure}[b]{0.31\linewidth}
    \includegraphics[width=\linewidth]{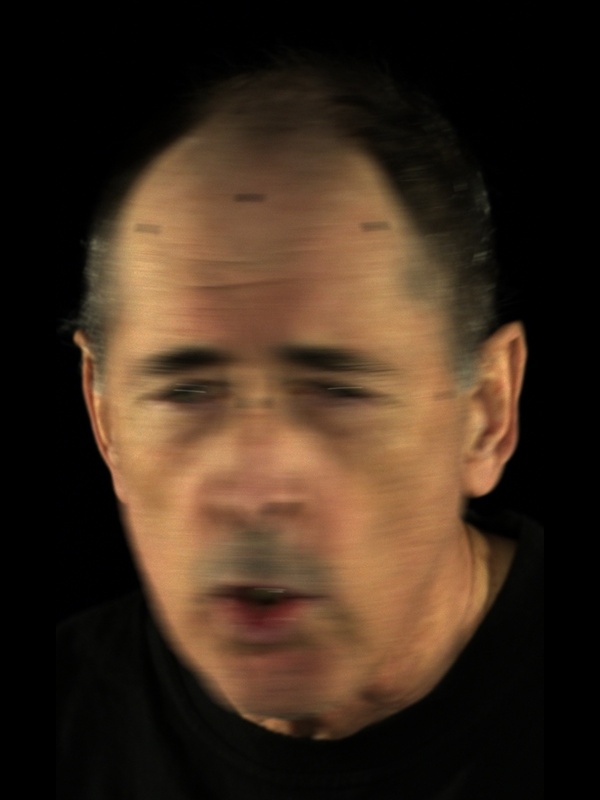}
    \includegraphics[width=\linewidth]{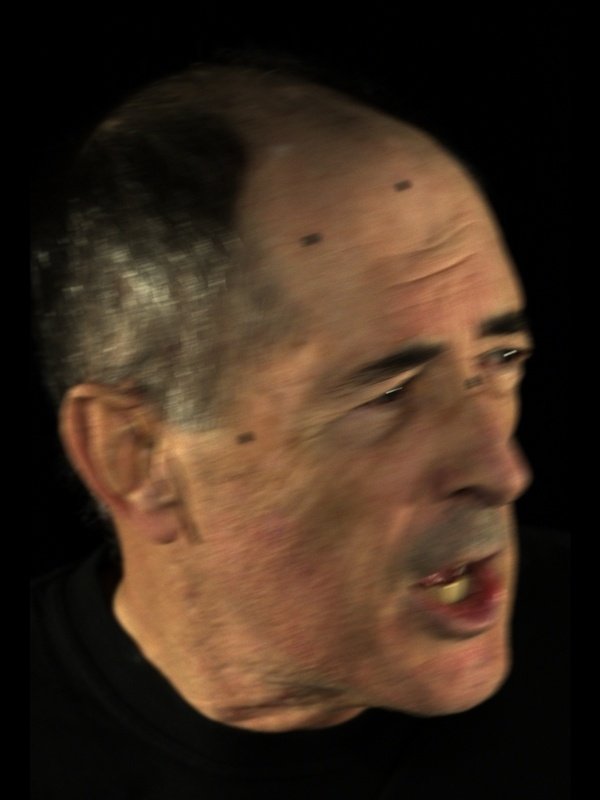}
    \includegraphics[width=\linewidth]{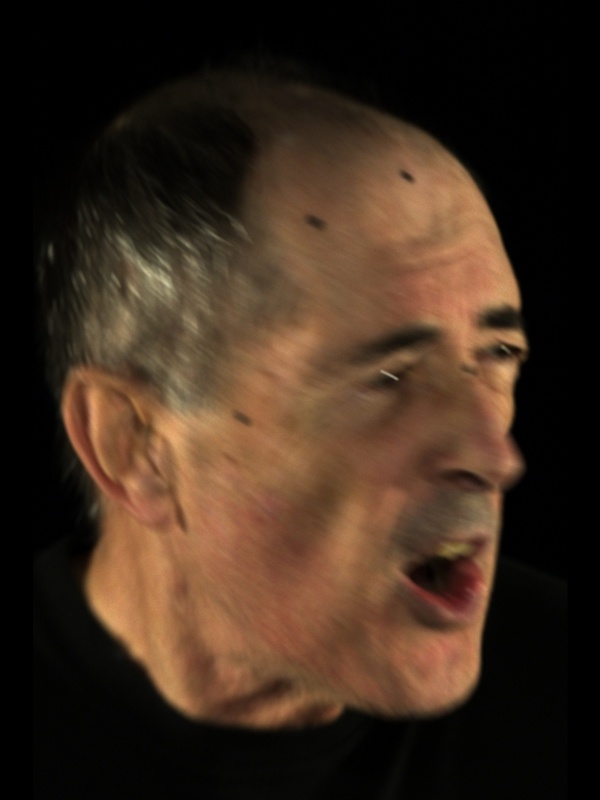}
    \includegraphics[width=\linewidth]{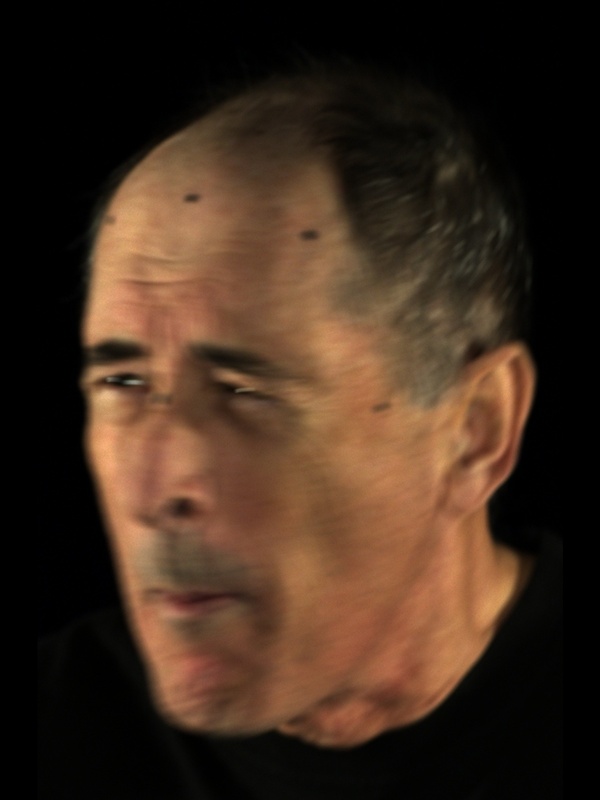}
    \caption{Plate}
\end{subfigure}
\hfill
\caption{Targeting the monocular RGB image using shape-from-shading and rotoscope curves with blendshapes and simulation from the main camera's perspective.}
\label{fig:plate_inversion_camB_appendix}
\end{figure}

\begin{figure}[t]
\centering
\begin{subfigure}[b]{\dimexpr0.31\linewidth+10pt\relax}
    \makebox[10pt]{\raisebox{40pt}{\rotatebox[origin=c]{90}{1115}}}%
    \includegraphics[width=\dimexpr\linewidth-10pt\relax]{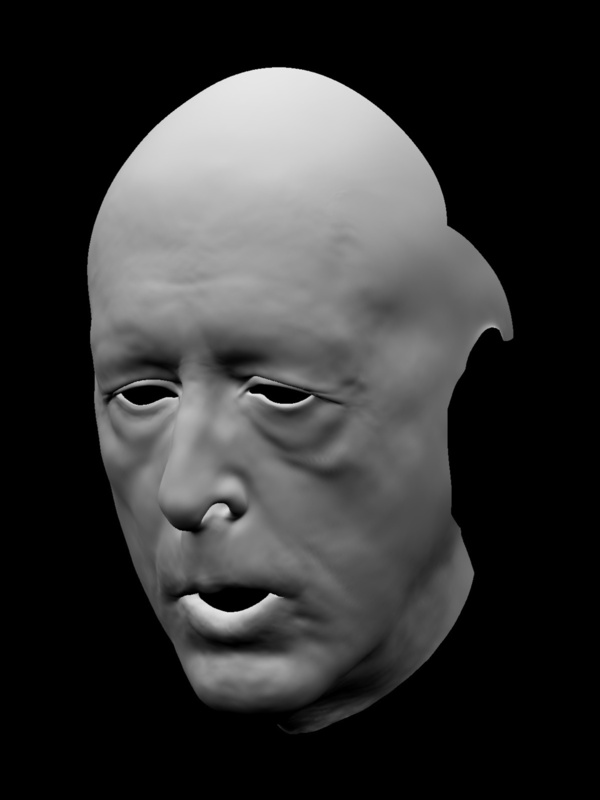}
    \makebox[10pt]{\raisebox{40pt}{\rotatebox[origin=c]{90}{1120}}}%
    \includegraphics[width=\dimexpr\linewidth-10pt\relax]{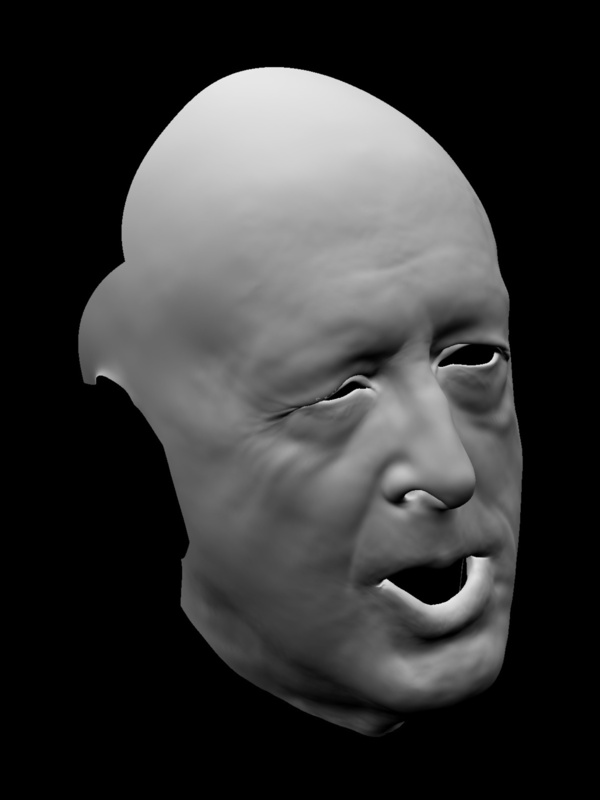}
    \makebox[10pt]{\raisebox{40pt}{\rotatebox[origin=c]{90}{1130}}}%
    \includegraphics[width=\dimexpr\linewidth-10pt\relax]{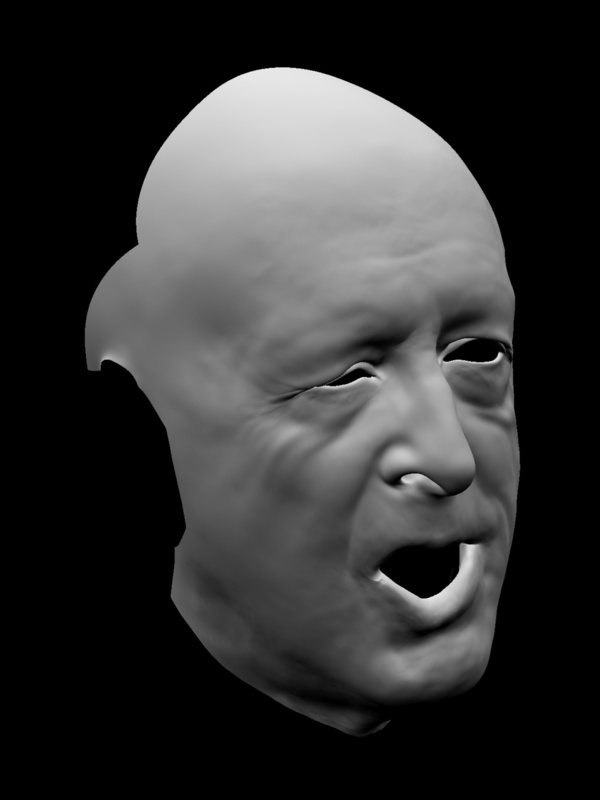}
    \makebox[10pt]{\raisebox{40pt}{\rotatebox[origin=c]{90}{1155}}}%
    \includegraphics[width=\dimexpr\linewidth-10pt\relax]{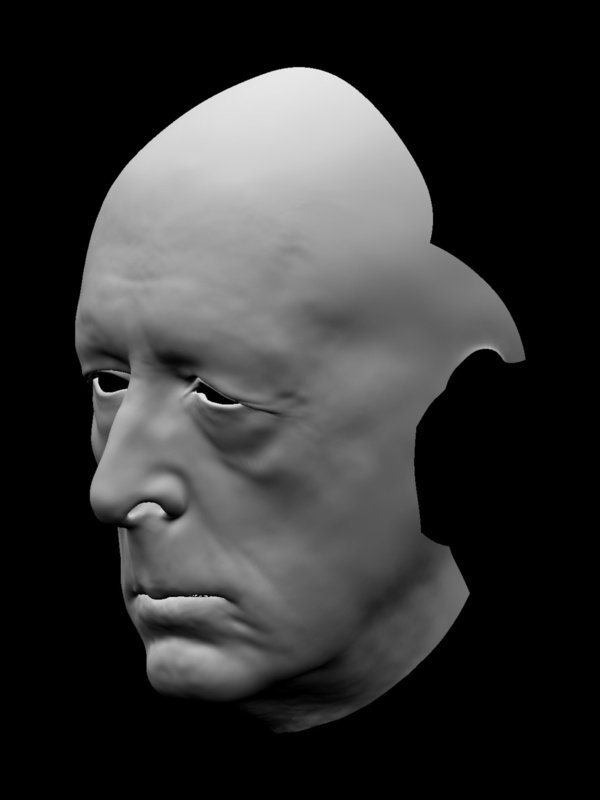}
    \caption{Blendshapes}
\end{subfigure}
\begin{subfigure}[b]{0.31\linewidth}
    \includegraphics[width=\linewidth]{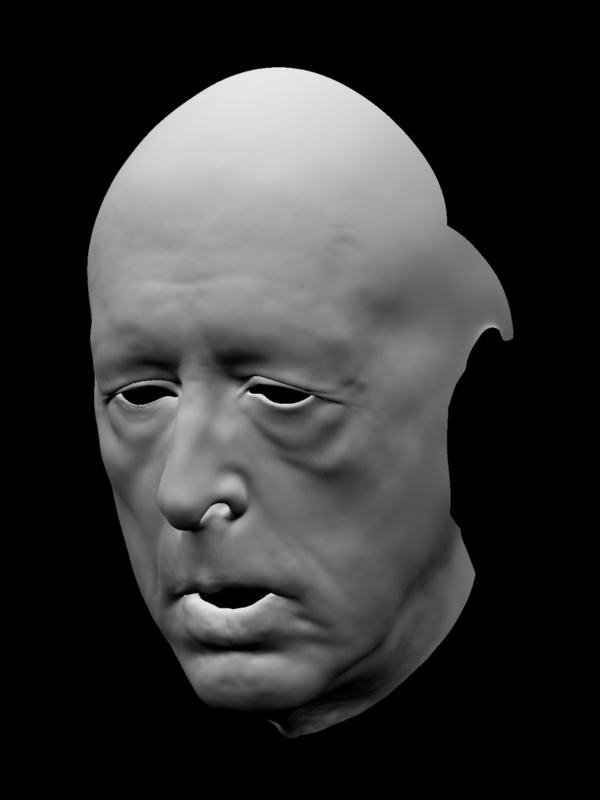}
    \includegraphics[width=\linewidth]{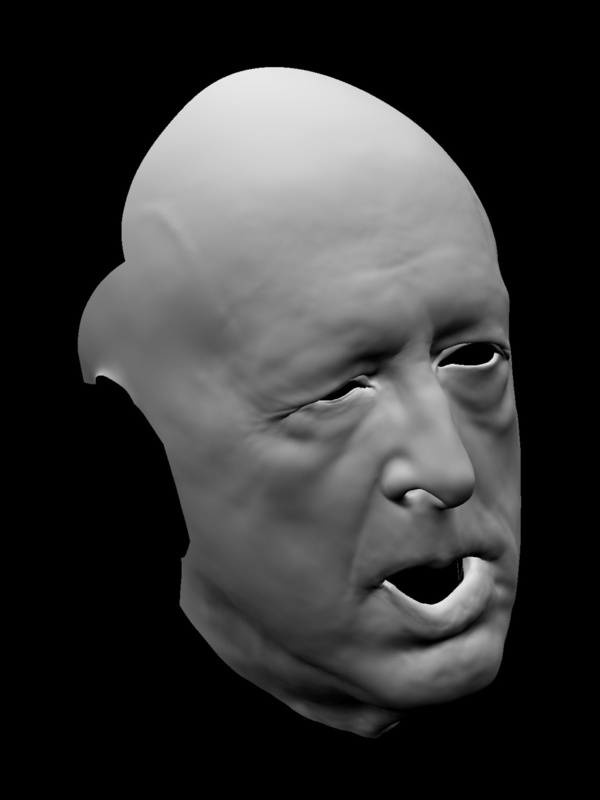}
    \includegraphics[width=\linewidth]{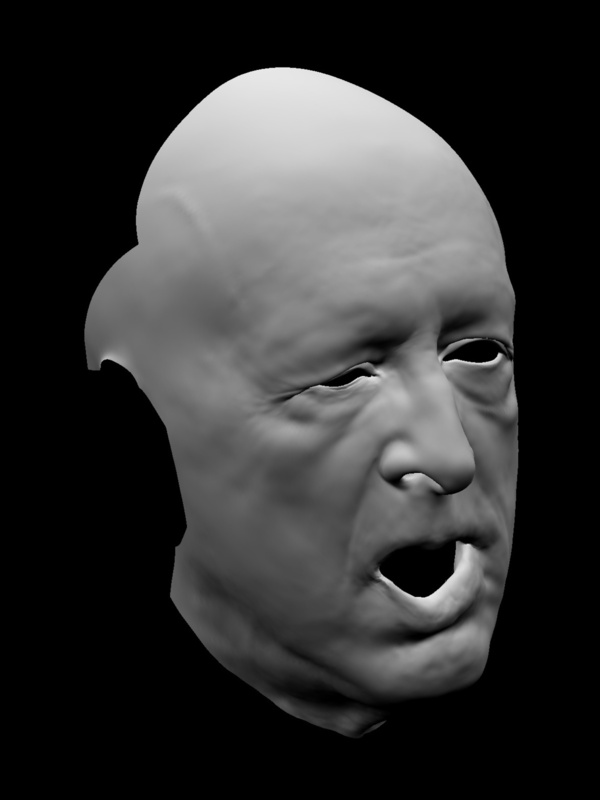}
    \includegraphics[width=\linewidth]{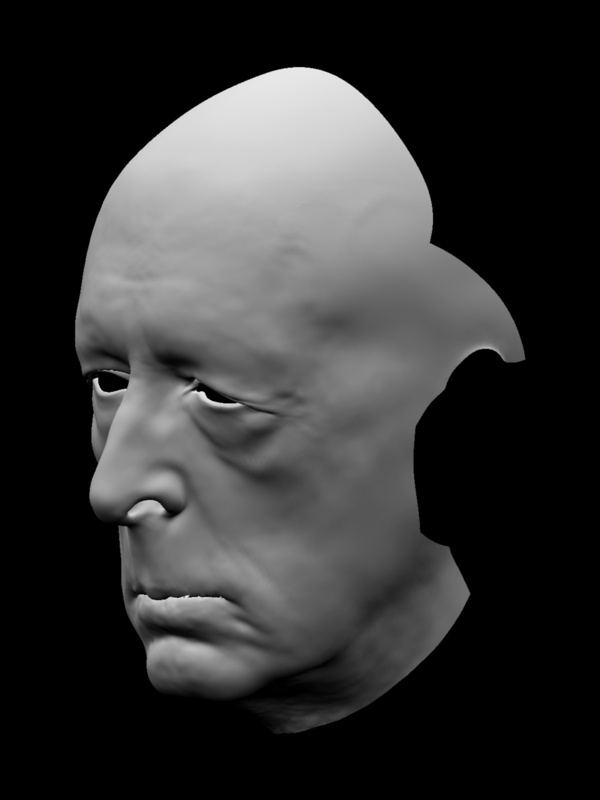}
    \caption{Simulation}
\end{subfigure}
\begin{subfigure}[b]{0.31\linewidth}
    \includegraphics[width=\linewidth]{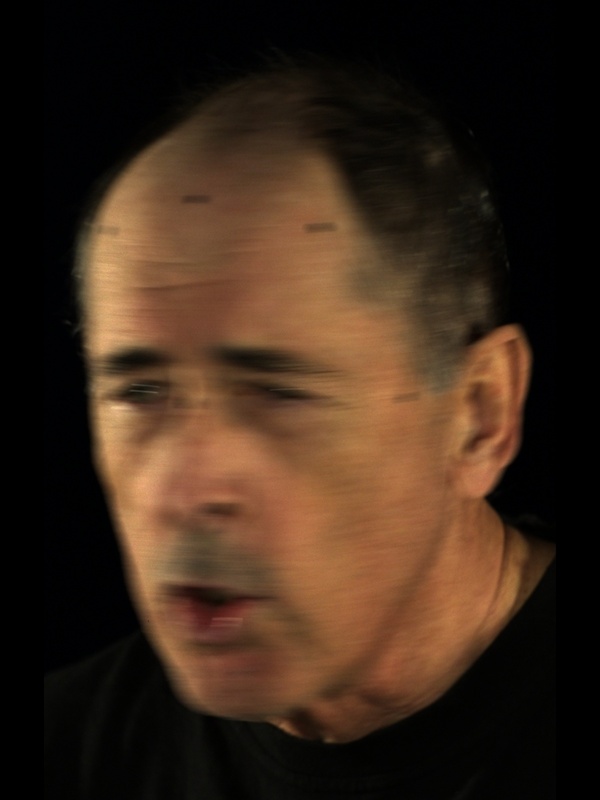}
    \includegraphics[width=\linewidth]{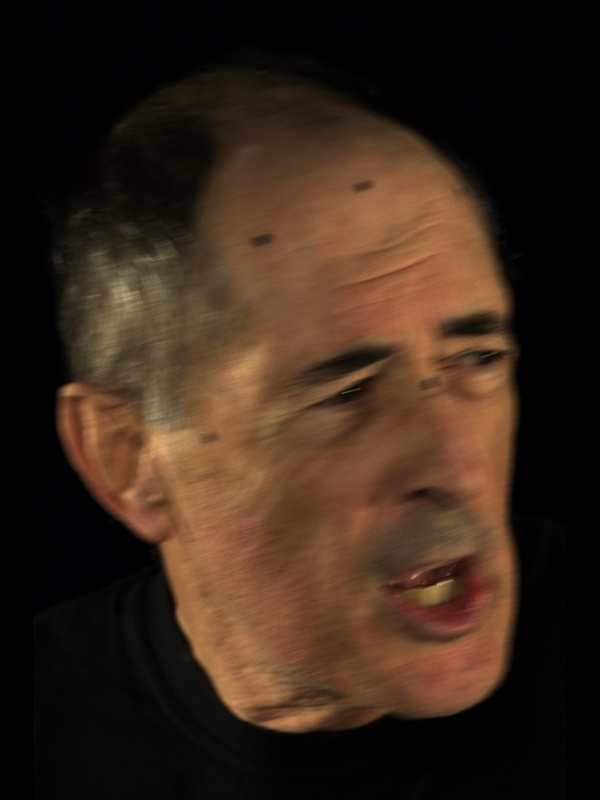}
    \includegraphics[width=\linewidth]{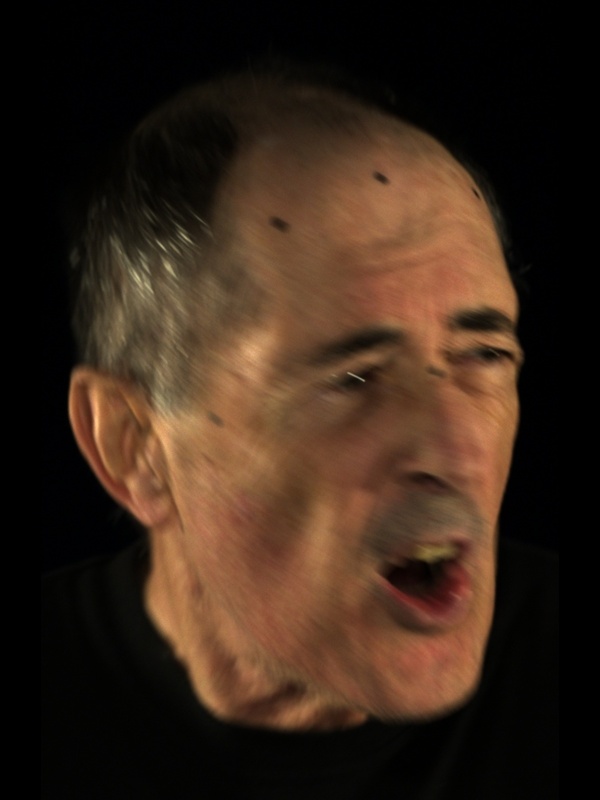}
    \includegraphics[width=\linewidth]{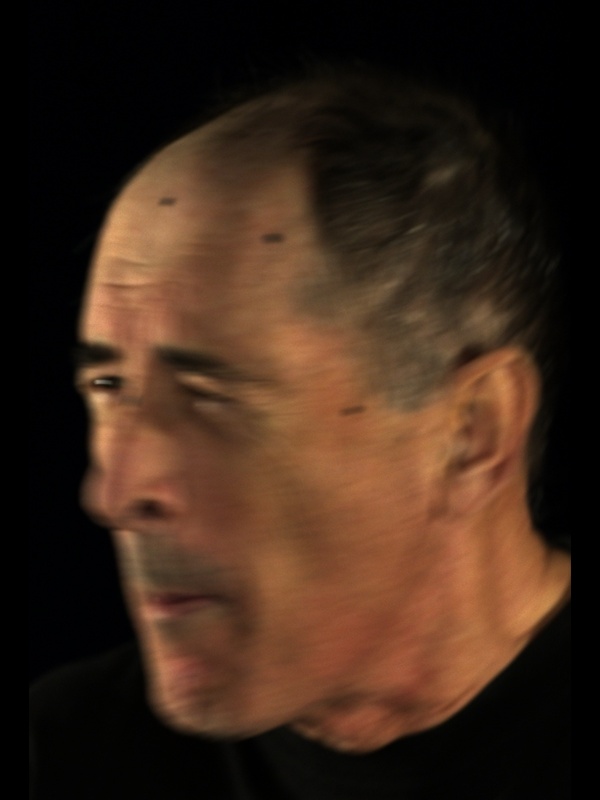}
    \caption{Plate}
\end{subfigure}
\hfill
\caption{Targeting the monocular RGB image using shape-from-shading and rotoscope curves with blendshapes and simulation from an alternate camera's perspective.}
\label{fig:plate_inversion_camA_appendix}
\end{figure}

\begin{figure*}[t]
\centering
\begin{subfigure}[b]{\dimexpr0.22\linewidth+10pt\relax}
    \makebox[10pt]{\raisebox{35pt}{\rotatebox[origin=c]{90}{Blendshapes}}}%
    \includegraphics[width=\dimexpr\linewidth-10pt\relax]{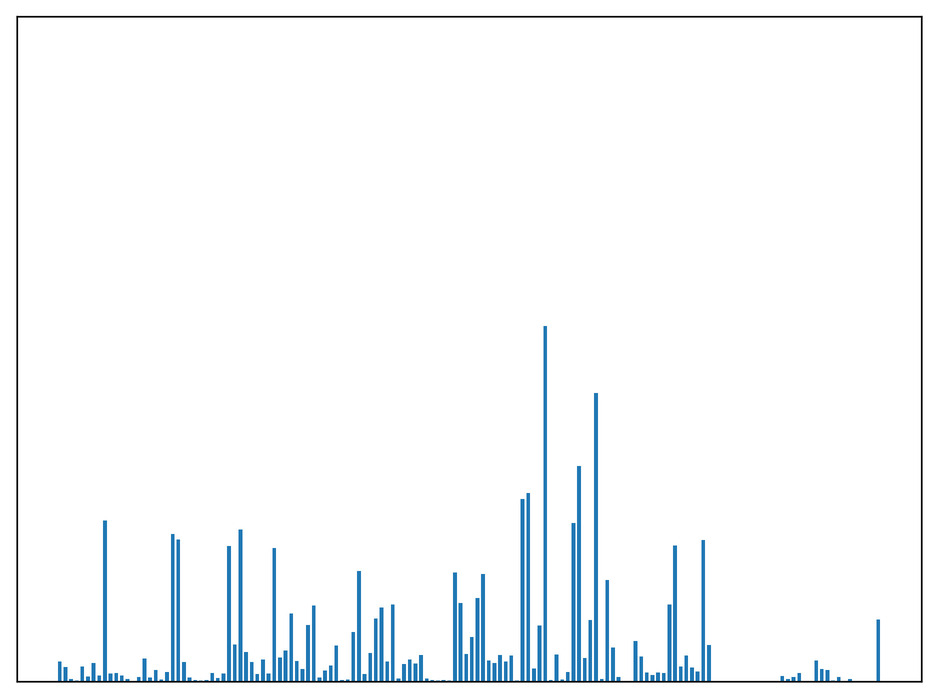}
    \makebox[10pt]{\raisebox{35pt}{\rotatebox[origin=c]{90}{Simulation}}}%
    \includegraphics[width=\dimexpr\linewidth-10pt\relax]{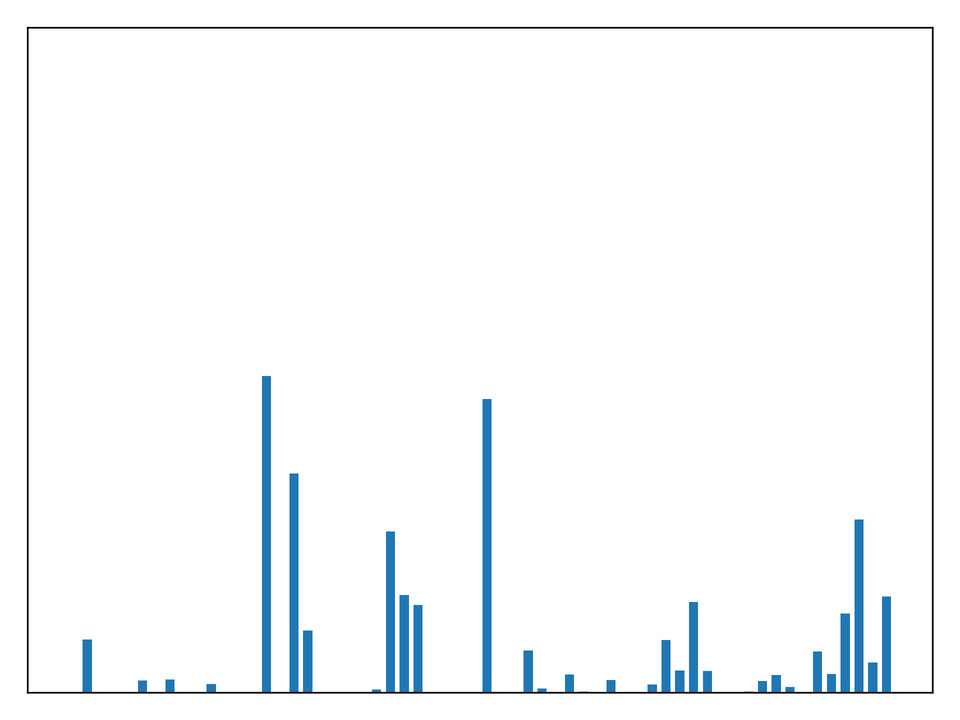}
    \caption{1112}
\end{subfigure}
\begin{subfigure}[b]{0.22\linewidth}
    \includegraphics[width=\linewidth]{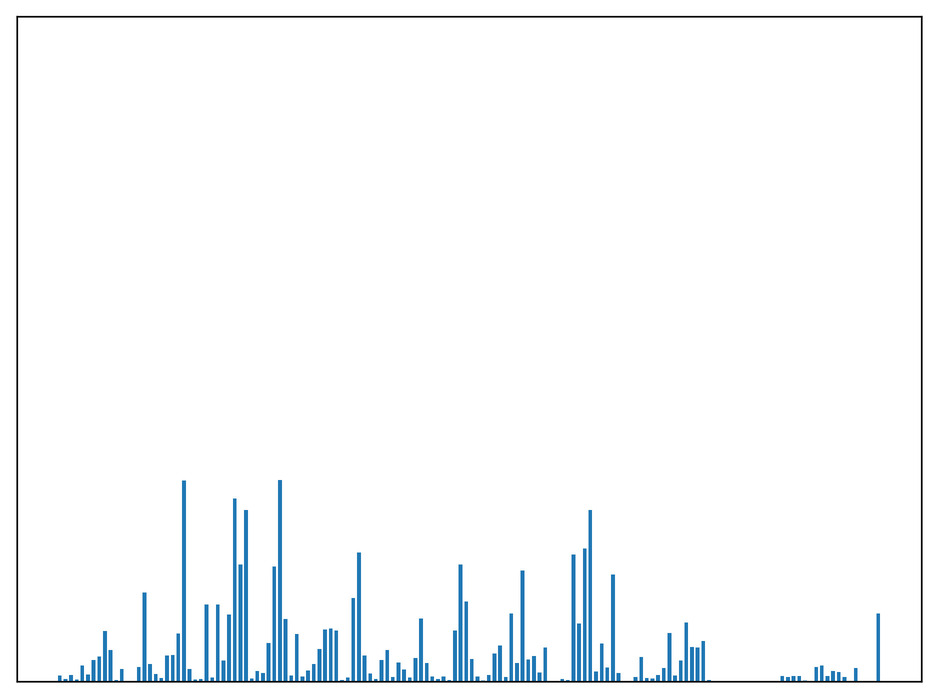}
    \includegraphics[width=\linewidth]{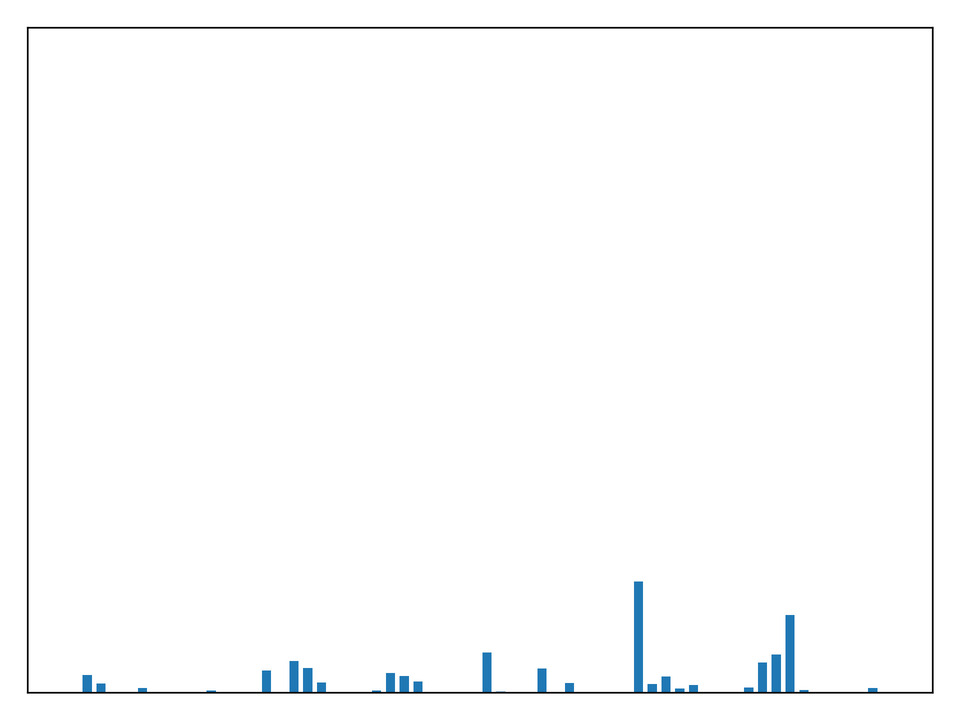}
    \caption{1115}
\end{subfigure}
\begin{subfigure}[b]{0.22\linewidth}
    \includegraphics[width=\linewidth]{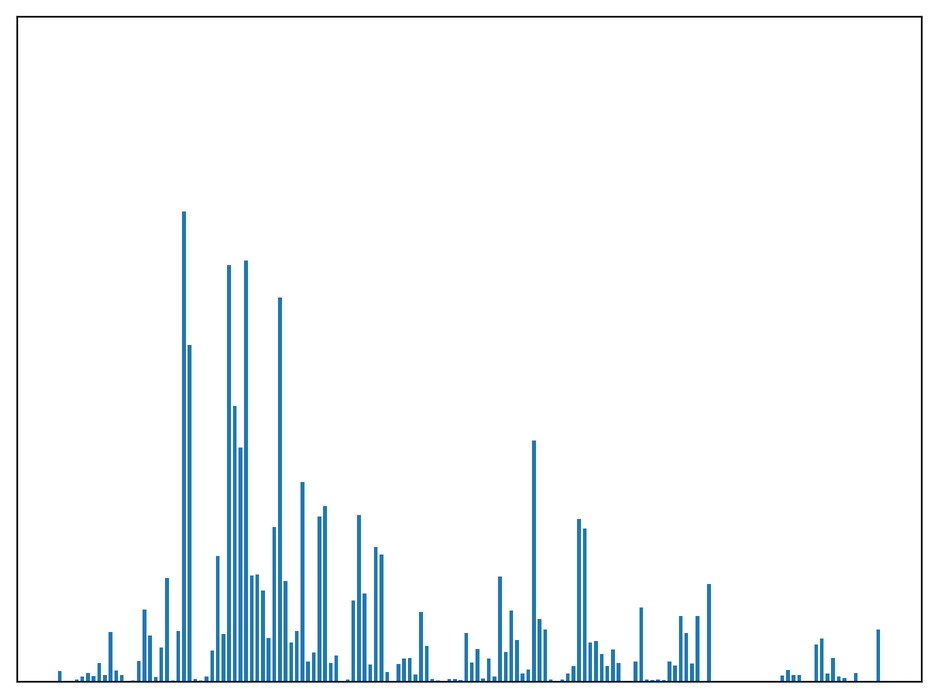}
    \includegraphics[width=\linewidth]{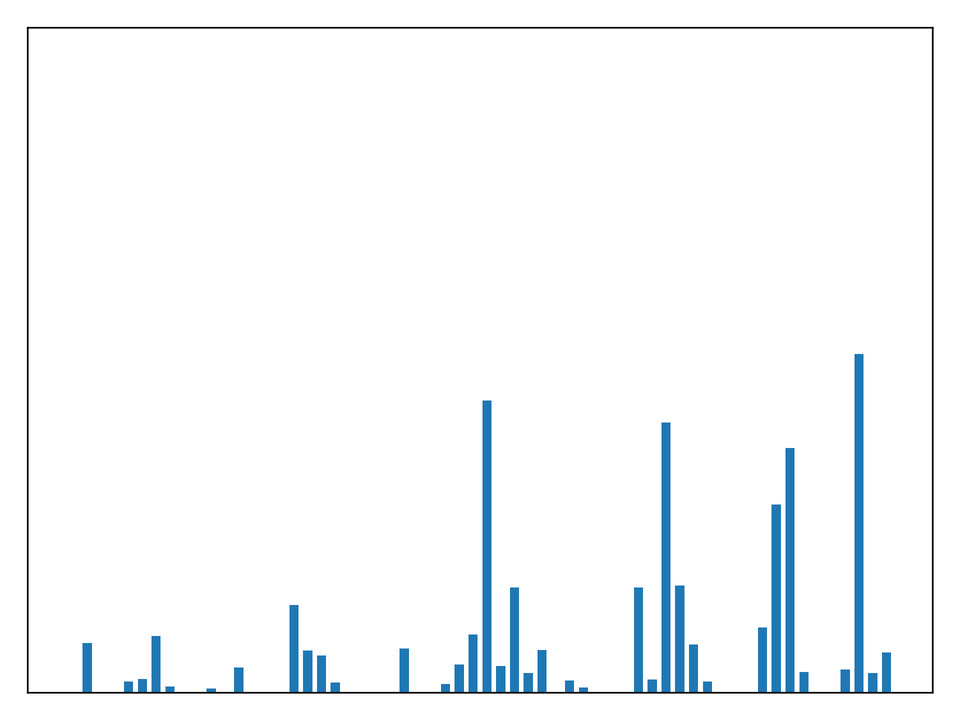}
    \caption{1120}
\end{subfigure}
\begin{subfigure}[b]{0.22\linewidth}
    \includegraphics[width=\linewidth]{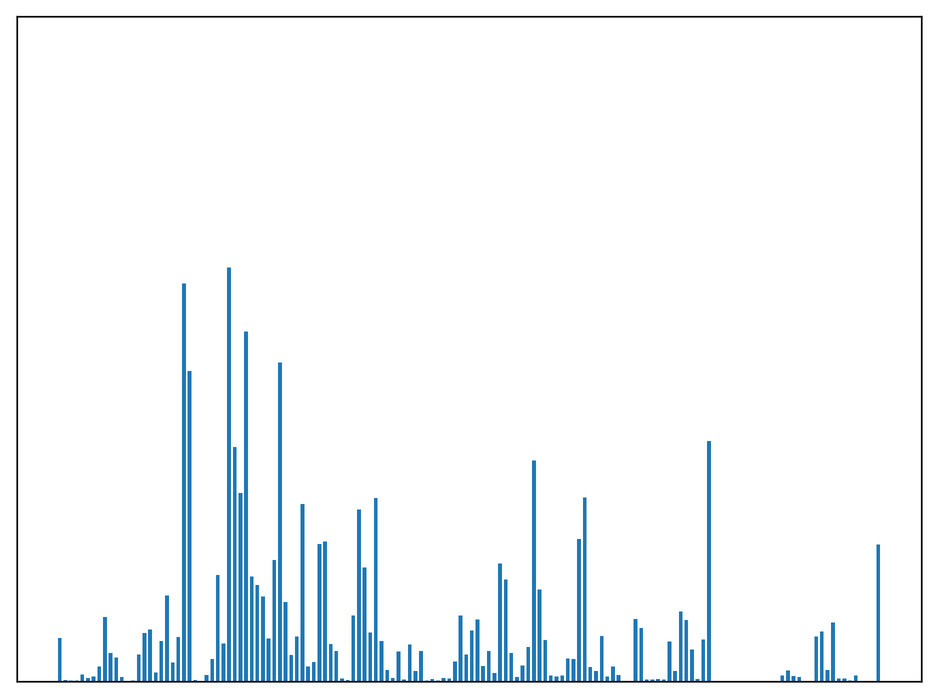}
    \includegraphics[width=\linewidth]{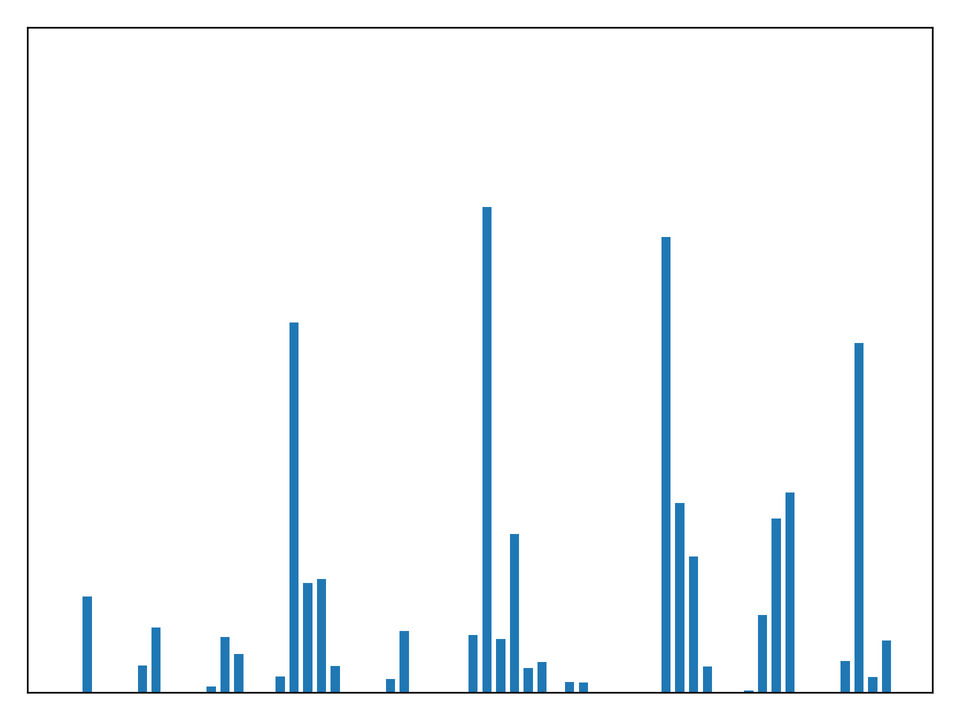}
    \caption{1130}
\end{subfigure}
\hfill
\begin{subfigure}[b]{\dimexpr0.22\linewidth+10pt\relax}
    \makebox[10pt]{\raisebox{35pt}{\rotatebox[origin=c]{90}{Blendshapes}}}%
    \includegraphics[width=\dimexpr\linewidth-10pt\relax]{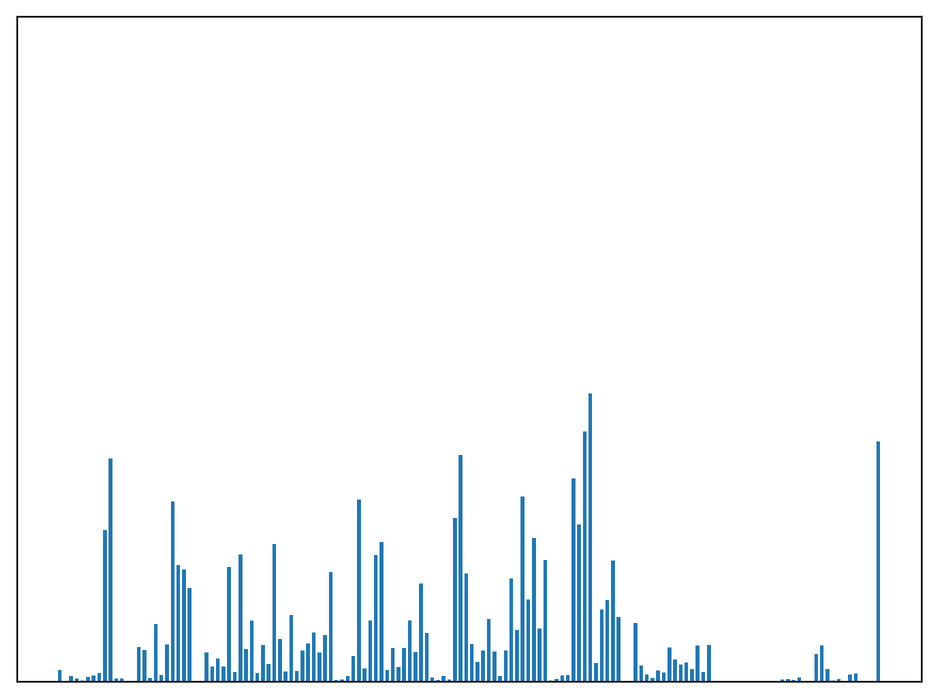}
    \makebox[10pt]{\raisebox{35pt}{\rotatebox[origin=c]{90}{Simulation}}}%
    \includegraphics[width=\dimexpr\linewidth-10pt\relax]{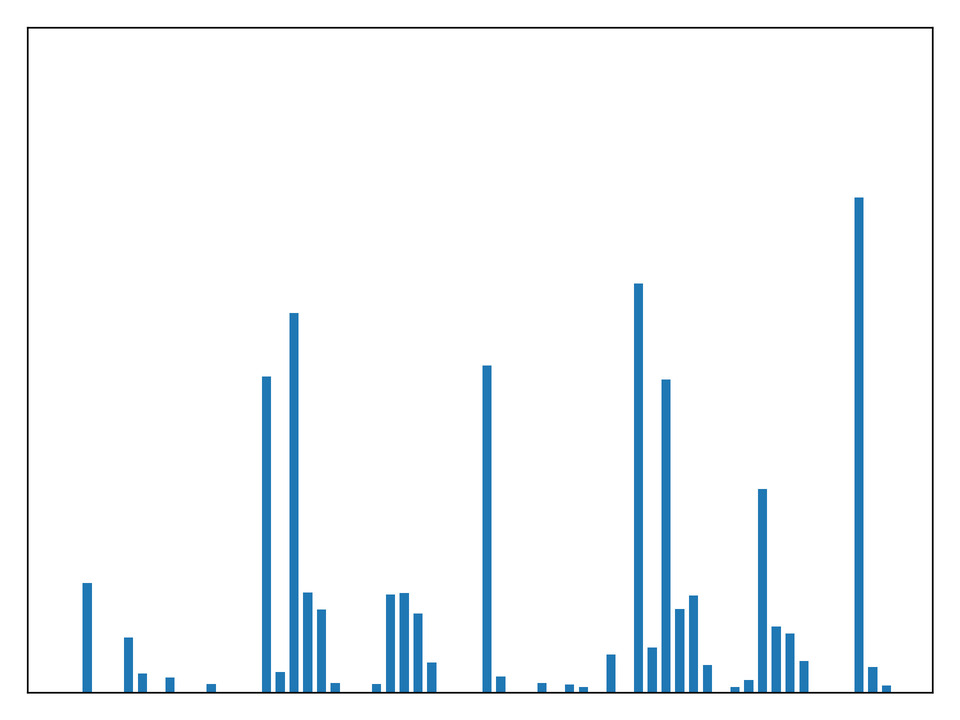}
    \caption{1134}
\end{subfigure}
\begin{subfigure}[b]{0.22\linewidth}
    \includegraphics[width=\linewidth]{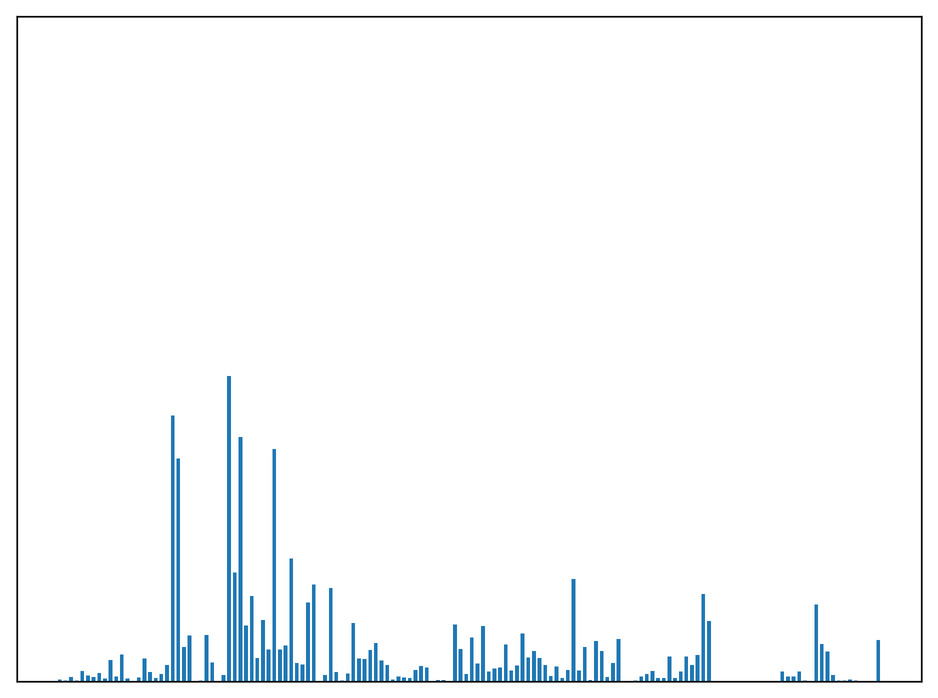}
    \includegraphics[width=\linewidth]{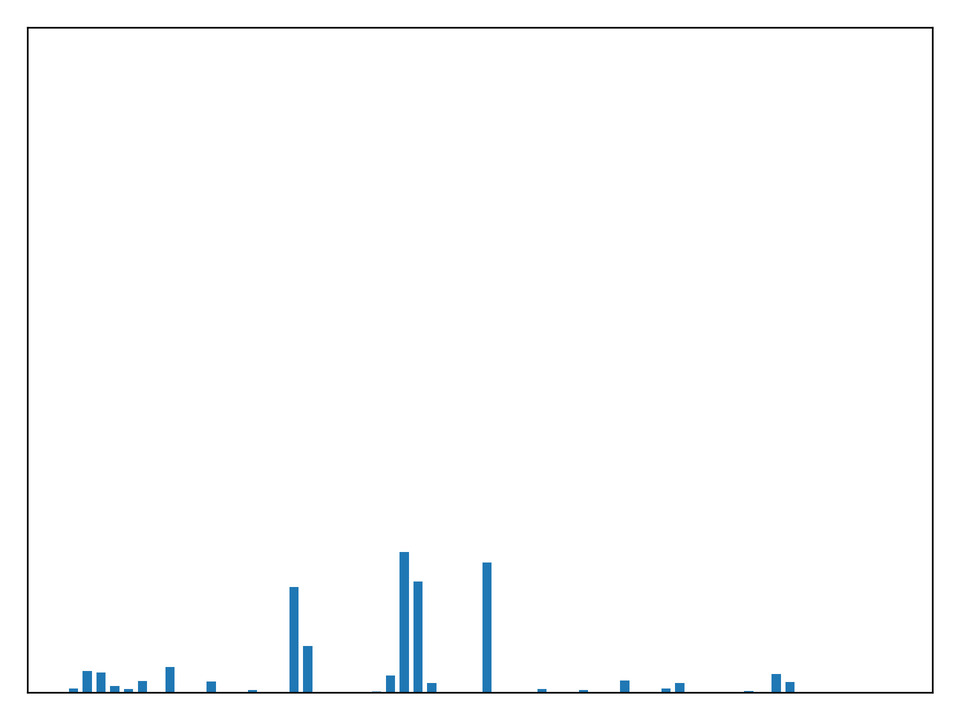}
    \caption{1155}
\end{subfigure}
\begin{subfigure}[b]{0.22\linewidth}
    \includegraphics[width=\linewidth]{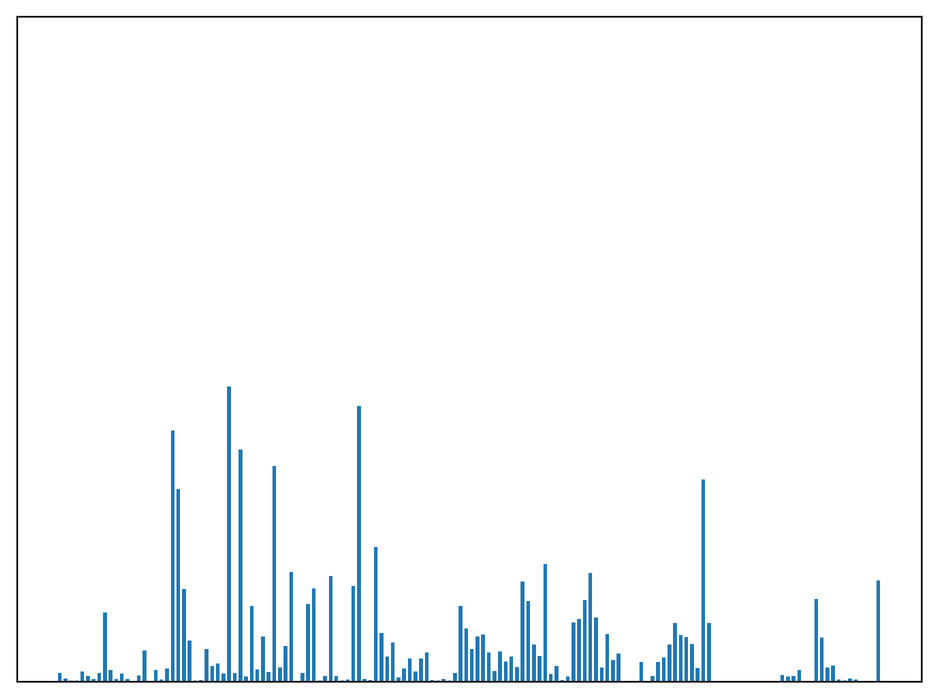}
    \includegraphics[width=\linewidth]{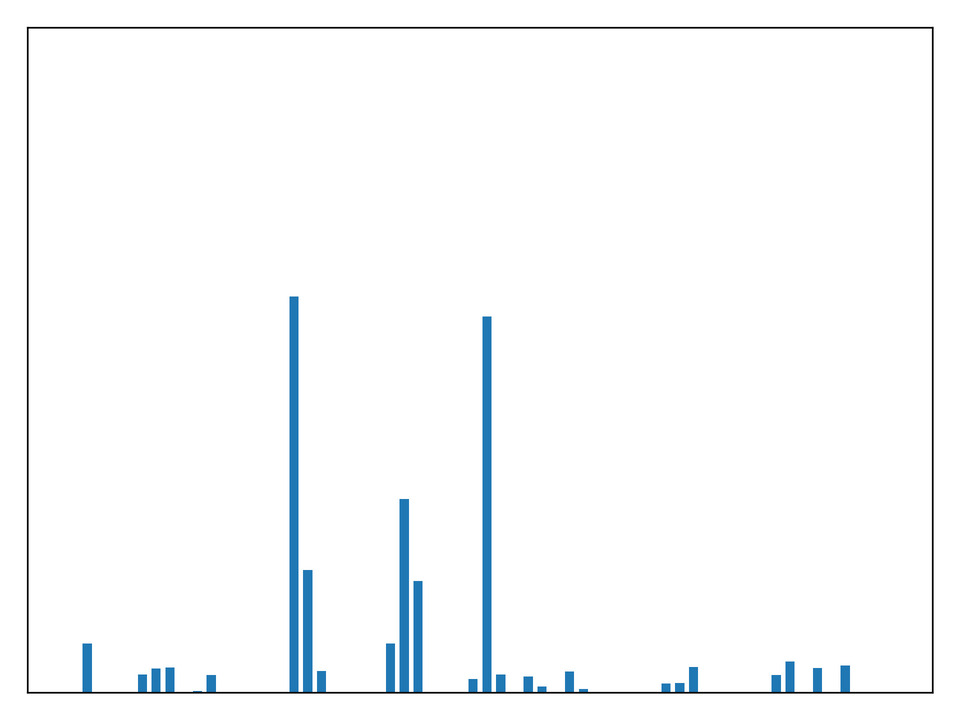}
    \caption{1160}
\end{subfigure}
\begin{subfigure}[b]{0.22\linewidth}
    \includegraphics[width=\linewidth]{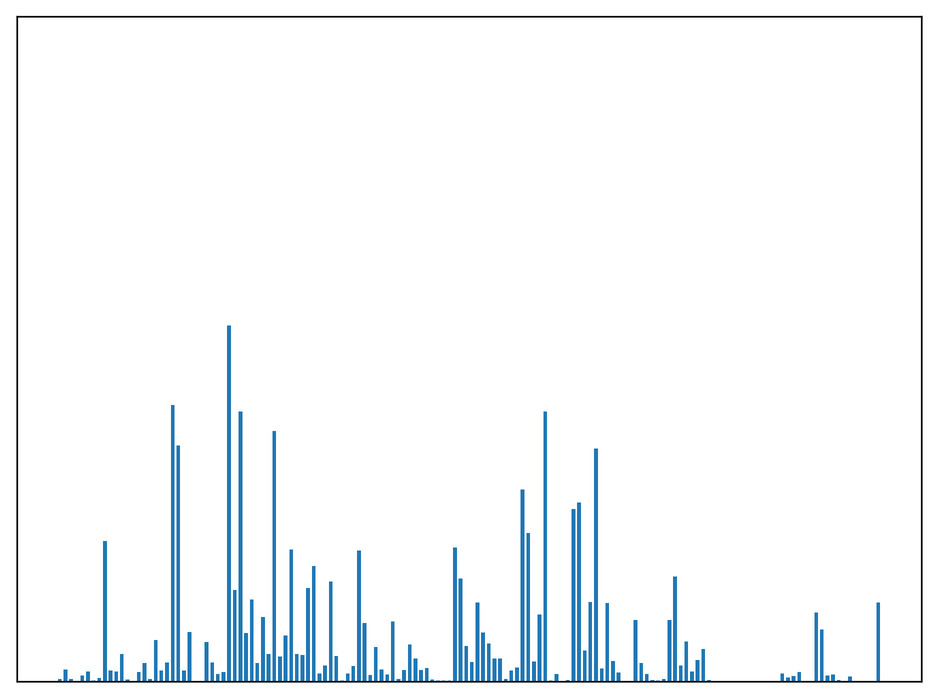}
    \includegraphics[width=\linewidth]{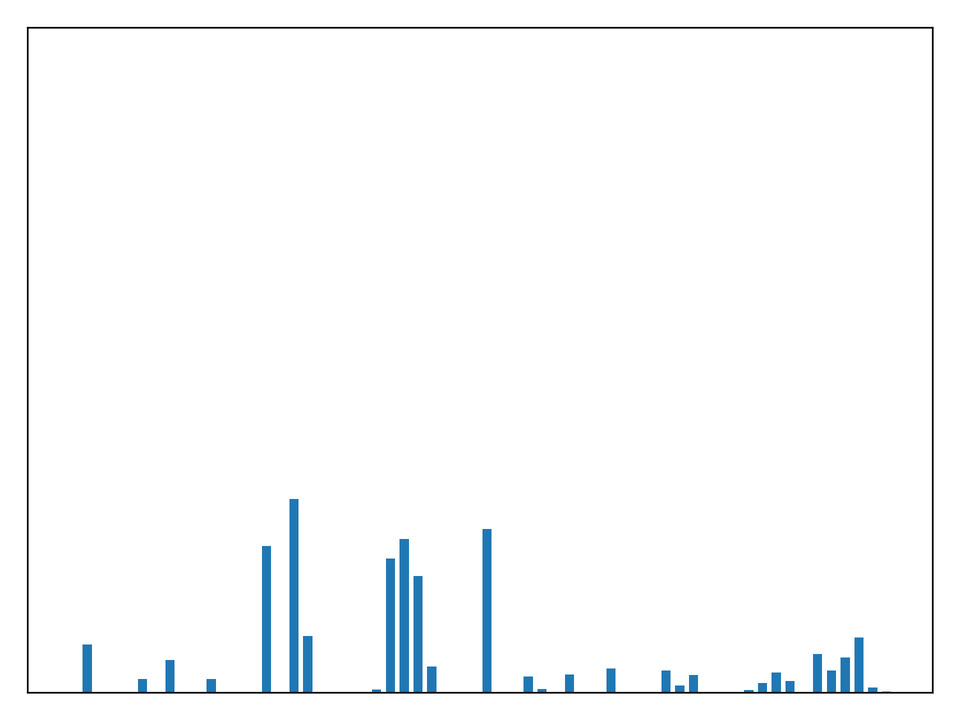}
    \caption{1170}
\end{subfigure}
\hfill
\caption{Comparisons between the blendshape weights and muscle activations for all the monocular shape-from-shading results.
The corresponding geometry for frames \num{1115}, \num{1120}, \num{1130}, and \num{1155} are shown in Figures \ref{fig:plate_inversion_camB_appendix} and \ref{fig:plate_inversion_camA_appendix}.
The corresponding geometry for frames \num{1112}, \num{1134}, \num{1160}, and \num{1170} are shown in the main paper.
}
\label{fig:plate_weights_appendix}
\end{figure*}

\begin{figure*}[t]
\centering
\begin{subfigure}[b]{0.22\linewidth}
    \includegraphics[width=\linewidth]{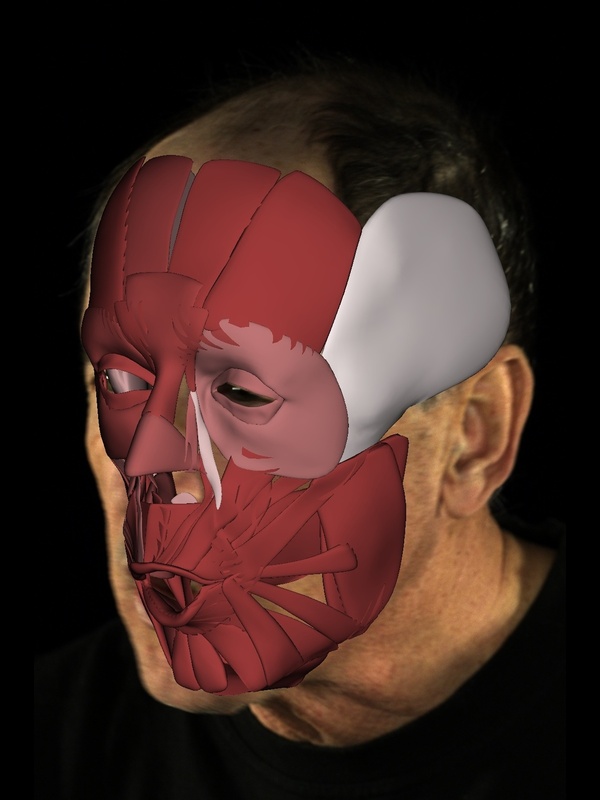}
    \caption{1112}
\end{subfigure}
\begin{subfigure}[b]{0.22\linewidth}
    \includegraphics[width=\linewidth]{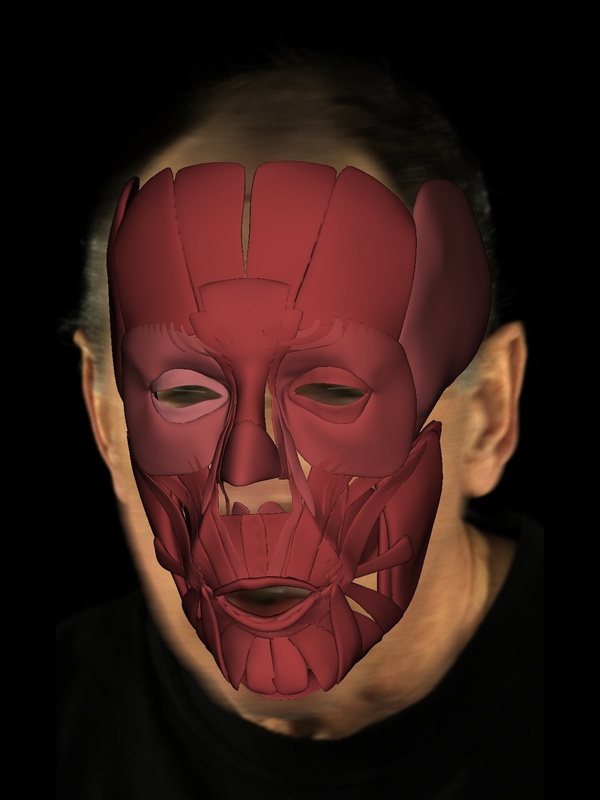}
    \caption{1115}
\end{subfigure}
\begin{subfigure}[b]{0.22\linewidth}
    \includegraphics[width=\linewidth]{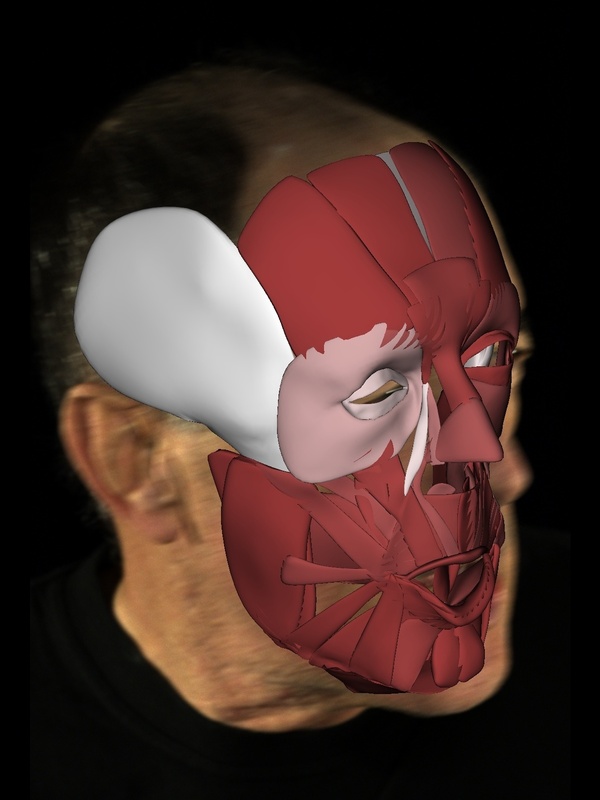}
    \caption{1120}
\end{subfigure}
\begin{subfigure}[b]{0.22\linewidth}
    \includegraphics[width=\linewidth]{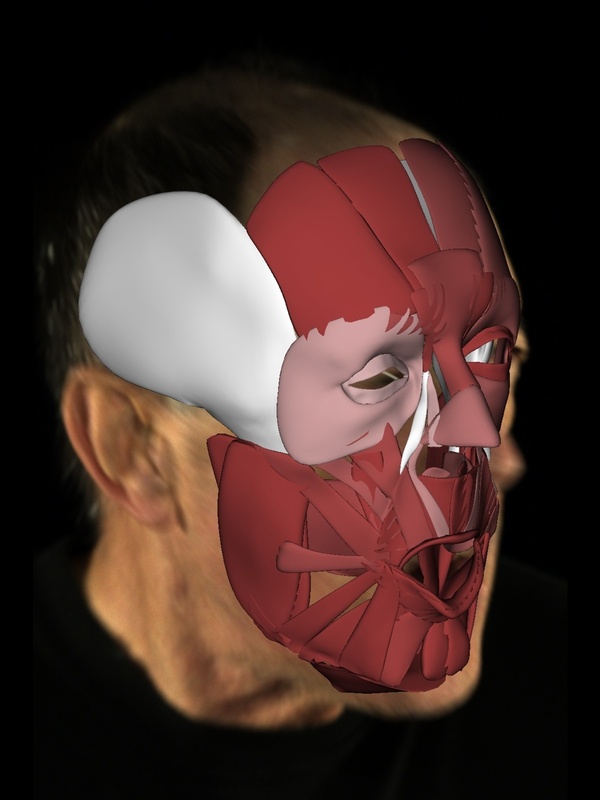}
    \caption{1130}
\end{subfigure}
\hfill
\begin{subfigure}[b]{0.22\linewidth}
    \includegraphics[width=\linewidth]{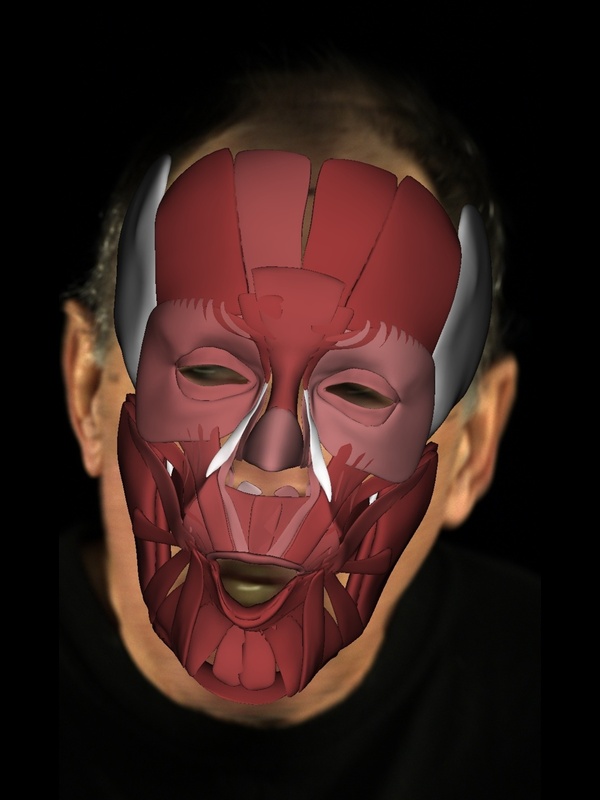}
    \caption{1134}
\end{subfigure}
\begin{subfigure}[b]{0.22\linewidth}
    \includegraphics[width=\linewidth]{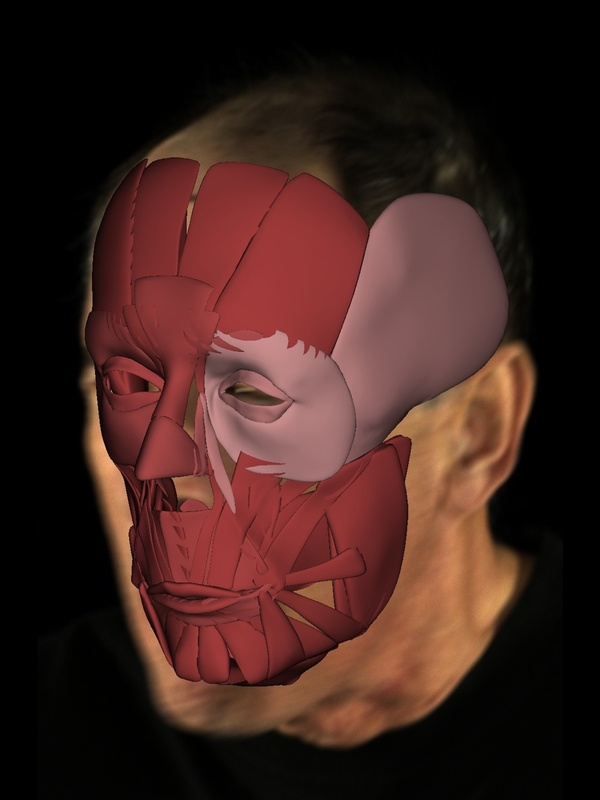}
    \caption{1155}
\end{subfigure}
\begin{subfigure}[b]{0.22\linewidth}
    \includegraphics[width=\linewidth]{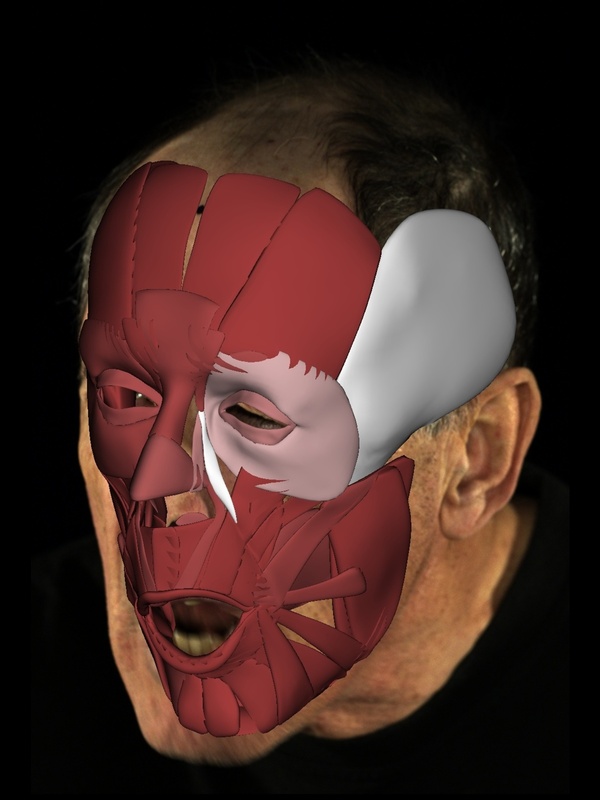}
    \caption{1160}
\end{subfigure}
\begin{subfigure}[b]{0.22\linewidth}
    \includegraphics[width=\linewidth]{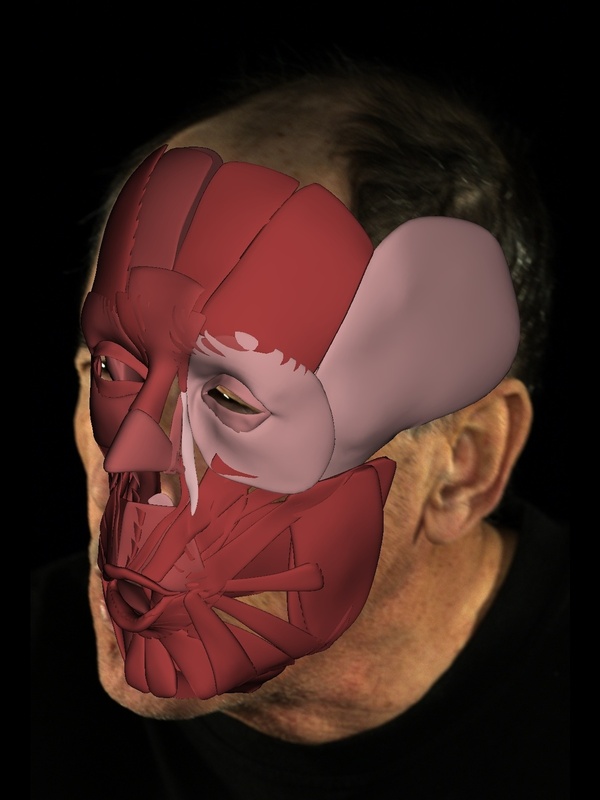}
    \caption{1170}
\end{subfigure}
\hfill
\caption{Muscle activations from Figure \ref{fig:plate_weights_appendix} visualized where activations greater than \num{0.5} are colored white and activations at \num{0} are colored red.}
\label{fig:plate_muscles_appendix}
\end{figure*}

We show additional results for targeting monocular RGB images in Figure \ref{fig:plate_inversion_camB_appendix}.
Furthermore, we show the resulting geometry and plates for the same frames from another camera perspective in Figure \ref{fig:plate_inversion_camA_appendix}.
The corresponding blendshape weights and muscle activations are shown in Figure \ref{fig:plate_weights_appendix}.
A visualization of the muscles' activations is shown in Figure \ref{fig:plate_muscles_appendix}.
Currently, the muscle activations resulting from targeting RGB images do not permit as clean of an interpretation as those obtained when targeting geometry, although the incisivus labii superioris muscles tend to become activated in conjunction with expressions involving the mouth.
However, we note that the general magnitude of the activations tends to match the magnitude of the expression.
Future work calibrating the muscle model will improve semantic intepretability.

\clearpage

\clearpage

{\small
\bibliographystyle{ieee}
\bibliography{biblio}
}

\end{document}